\documentclass[12pt]{article}
\pdfoutput=1

\usepackage{graphicx} % Required for inserting images

\usepackage[margin=1in]{geometry}
\usepackage[normalem]{ulem}

\usepackage{times}
\usepackage{float}
\usepackage{placeins}
\usepackage{latexsym}
\usepackage{microtype}
\usepackage{hyperref}
\usepackage[english]{datetime2}
\usepackage[most]{tcolorbox}
\usepackage[linesnumbered,ruled,vlined]{algorithm2e}
\usepackage{amsmath,amssymb}
\usepackage{subcaption}
\usepackage{caption}
\usepackage[T1]{fontenc}
\usepackage[absolute,overlay]{textpos}
\usepackage{booktabs}
\usepackage{tikz}
\usetikzlibrary{matrix,decorations.pathreplacing,calc}
\usetikzlibrary{arrows.meta}

\usepackage[utf8]{inputenc}
\usepackage{authblk}
\usepackage{longtable}
\usepackage{hyphenat}
\usepackage{tabularx}
\usepackage{booktabs}
\usepackage{subcaption}
\usepackage{wasysym}
\usepackage[table]{xcolor}
\usepackage{siunitx}

\usepackage{listings}
\usepackage{titlesec}

\definecolor{HighlightA}{RGB}{200,225,255} % darker blue
\definecolor{HighlightB}{RGB}{200,245,200} % darker green
\definecolor{HighlightC}{RGB}{255,225,190} % darker orange

\DTMnewdatestyle{monthyeardate}{%
}

\usepackage[backend=biber,style=authoryear]{biblatex}
\usepackage{inconsolata}

\usepackage{graphicx}

\title{AI Learning and Conceptual Transfer in the Game of Hidden Rules}

\author[1]{Christo Mathew}
\author[2]{Wentian Wang}
\author[1]{Jacob Feldman}
\author[1]{Lazaros K. Gallos}
\author[1,3]{Paul B. Kantor}
\author[1]{Vladimir Menkov}
\author[1]{Hao Wang}

\affil[1]{Rutgers University, New Brunswick}
\affil[2]{University of Southern California}
\affil[3]{Paul B. Kantor, Consultant}
\begin{document}

\begin{textblock*}{7cm}(12cm,1cm)
\raggedleft
\small
\textbf{TECHNICAL REPORT}\\
\DTMsetdatestyle{monthyeardate}\today
\end{textblock*}

\maketitle
\begin{abstract}
This report summarizes the work conducted on the Game of Hidden Rules (GOHR), focusing on reinforcement learning agents trained to infer hidden rules from trial-and-error feedback, representation design, rule difficulty analysis, transfer learning, generalization, and pseudo-bot-assisted human learning analysis.  The report 
% \pk{I would take out: excludes the meta-model experiments and instead} 
focuses on the Transformer-based A2C framework, Feature-Centric and Object-Centric representations, experimental findings, and 
classification
% behavioral modeling 
of human learning data. 
% \jf{I would say: classification of human learning data}. 
\end{abstract}
\tableofcontents
% \newpage
% \clearpage
% \section{Introduction}
% \subsection{Motivation}
% % Discuss why hidden-rule learning is important.
% % Explain GOHR as a controlled environment for studying rule discovery, generalization, and transfer.

% \subsection{Problem Statement}
% % Define the central problem: learning latent rules from sparse accept/reject feedback.

% \subsection{Summary of Contributions}
% % Suggested bullets:
% % 1. GOHR environment refactor and Gymnasium-compatible design.
% % 2. FC and OC state representations.
% % 3. Transformer-based A2C model.
% % 4. Rule difficulty and property-level analysis.
% % 5. Generalization and transfer experiments.
% % 6. Pseudo-bot-assisted human learning analysis.

% \subsection{Report Organization}

\section{Introduction}

\subsection{Motivation}

Humans and intelligent agents frequently encounter situations in which the underlying rules governing the environment are not explicitly provided. Instead, these rules must be inferred through interaction, observation, and feedback. Learning such hidden structures is fundamental to many real-world problems, including scientific reasoning, strategic decision making, robotics, and adaptive control.

The \emph{Game of Hidden Rules} (GOHR) provides a controlled environment for studying this type of learning. In GOHR, an agent observes a board containing multiple objects and must infer an unknown latent rule that determines whether a particular action is accepted or rejected. The agent does not receive direct supervision regarding the rule itself; instead, learning occurs only through sparse feedback indicating whether an attempted move satisfies the hidden constraints.
% \pk{s}.

This setting presents several important challenges for machine learning systems. First, the agent must discover an
% \pk{an} 
abstract relational structure from limited feedback. Second, it must generalize across different rule families that vary in complexity and representation. Third, it has the opportunity to
% \pk{has the opportunity to}
% must 
transfer previously learned knowledge to new but structurally related tasks. These challenges are closely related to broader questions in reinforcement learning and representation learning regarding abstraction, compositionality, and transfer.

GOHR is particularly well suited for studying these questions because the environment contains diverse rule families with interpretable structure. Some rules depend on object features such as shape or color, while others depend on spatial relationships, bucket ordering, or combinations of multiple rule components. This diversity enables systematic analysis of rule difficulty, generalization, and transfer behavior under controlled conditions.

In addition to reinforcement learning analysis, GOHR also provides an opportunity to compare artificial learning behavior with human decision patterns.\footnote{In fact, it was designed and built with such comparison in mind~\cite{bier_can_2019}.} Understanding whether machine-learning representations capture structures similar to those used by humans may provide insight into interpretable and transferable learning strategies.

\subsection{Problem Statement}

The central problem addressed in this work is the learning of latent rules from sparse accept/reject feedback within the GOHR environment (Section ~\ref{sec: Game of Hidden Rules}); see also~\cite{pulick_comparing_2024}.

At each step, the agent observes the current board state and selects an action corresponding to placing an object into one of several buckets. The environment evaluates the action according to a hidden rule and returns only limited feedback indicating whether the move was accepted or rejected. The agent is therefore required to infer the latent rule indirectly through interaction with the environment.

Several research questions arise from this setting:

\begin{itemize}

    \item How does the internal state representation affect learning performance and generalization?

    \item Which rule properties contribute most strongly to learning difficulty?

    \item Can previously learned rules facilitate learning of structurally related rules (concept transfer)?

    \item Do different rule families induce consistent transfer-learning structures?

    \item To what extent do reinforcement-learning agents exhibit behavior patterns similar
    % \pk{similar}
    %comparable 
    to human learners?

\end{itemize}

To investigate these questions, this work compares two state representations:

\begin{enumerate}

    \item a \emph{Feature-Centric (FC)} representation that encodes the board as feature maps over spatial positions, and

    \item an \emph{Object-Centric (OC)} representation that explicitly represents individual objects and their attributes.

\end{enumerate}

Both representations are trained using a transformer-based Actor-Critic reinforcement-learning framework. The resulting agents are examined
% \pk{examined }%evaluated
across independent rule learning, generalization, transfer learning, and transfer-geometry analyses.

\subsection{Summary of Contributions}

The primary contributions of this work are summarized below:

\begin{enumerate}

    \item We develop an Object-Centric (OC) state representation based on explicit object attributes and systematically compare its learning, generalization, and transfer behavior against the previously proposed Feature-Centric (FC) representation (\cite{pulick_game_2022})
    % \pk{previously used \cite{pulick_game_2022}}
    % existing 
    .

    \item We develop a transformer-based Actor-Critic (\cite{mathew2025toward})
    % \pk{CITATION?}
    reinforcement-learning architecture for learning hidden rules from sparse feedback.

    \item We analyze rule difficulty across multiple rule families and investigate how structural properties of the rules
    % \pk{of the rules} 
    influence 
    %learning 
    performance of the learning systems.
    % \pk{ of the learning systems} .

    \item We study generalization behavior across related shape and color rules under both FC and OC representations.

    \item We conduct extensive transfer-learning experiments, including both simple-to-simple transfer and compound-rule transfer settings.

    \item We analyze a
    % \pk{a }
    % the 
    geometry of transfer learning using hierarchical clustering, multidimensional scaling (MDS), and tanglegram comparisons to study relationships between rule families and transfer behavior.

    \item We analyze
    % \pk{analyze }
    a pseudo-bot-assisted human learning seeking
    % \pk{seeking}
    behavioral differences between assisted and non-assisted 
    human
    % \pk{human} 
    gameplay.

    % \item We provide a Gymnasium-compatible implementation of GOHR that supports both legacy and standard observation pipelines and facilitates integration with modern reinforcement-learning frameworks.
    We develop a Gymnasium-compatible~\cite{towers2026gymnasium} implementation of GOHR that preserves the legacy observation pipeline through compatibility wrappers while enabling integration with modern reinforcement-learning frameworks.

    % \item We refactor the GOHR environment into a modular Gymnasium-compatible 
    % \pk{SHOULDM'T THERE BE A CITATION ABOUT GYMNASIUM}  
    % framework with improved extensibility, wrapper support, and reinforcement-learning integration.

\end{enumerate}

% \subsection{Report Organization}

% The remainder of this report is organized as follows.

% Chapter~2 introduces the Game of Hidden Rules environment and describes both the legacy implementation and the new Gymnasium-compatible framework. Chapter~3 presents the reinforcement-learning methodology, including the FC and OC state representations, transformer-based Actor-Critic architecture, and evaluation metrics. Chapter~4 describes the experimental setup and training configuration.

% Chapter~5 presents the experimental results, including independent rule learning, generalization analysis, transfer-learning experiments, multidimensional scaling analysis, and transfer-geometry clustering analysis. Chapter~6 presents the human learning analysis using pseudo-bot-assisted modeling. Finally, Chapters~7 and~8 discuss the broader implications of the findings and summarize the conclusions and future directions of this work.

% =========================================================
\section{The Game of Hidden Rules}\label{sec: Game of Hidden Rules}

\subsection{Overview of GOHR}

The \textbf{Game of Hidden Rules (GOHR)} is the primary environment used in this work. It is a controlled experimental framework designed to study rule discovery, learning, and generalization in both humans and artificial agents. GOHR was developed by researchers at the University of Wisconsin, Madison, and has been 
% \pk{I WOULD TAKE OUT: widely} 
used as a testbed for studying hidden-structure learning (\cite{pulick_game_2022,bier_can_2019,bier_gohr_nodate,pulick_comparing_2024}).  
% \pk{I THINK WHEN WE JUST LIST REFERENCES WE SHOULD PUT (, ) around them }

The environment consists of a $6 \times 6$ grid-based board and four buckets positioned at the corners. At the start of each episode, $n=9$ objects are randomly placed on the board. Each object is characterized by two attributes: \textit{shape} and \textit{color}. A hidden rule governs how objects must be assigned to buckets.

The objective of the player (human or agent) is to correctly infer this hidden rule through trial-and-error interaction and successfully place all objects into valid buckets. Since the rule is not explicitly provided, the task involves two coupled challenges: identifying the latent rule that governs object-bucket assignments and learning a policy that selects actions consistent with that rule.
% Since the rule is not explicitly provided, the task requires simultaneous \textbf{rule inference} and \textbf{policy learning}.
% \pk{I DON'T UNDERSTAND WHICH IS WHICH HERE}

Two key entities define the environment:
\begin{itemize}
    \item \textbf{Rules:} Define constraints mapping object properties (e.g., color, shape, position) to valid bucket assignments.
    \item \textbf{Pieces (Objects):} Game elements with attributes such as shape and color, randomly initialized on the board at the start of each episode.
\end{itemize}

\subsection{The Game Mechanics}

\subsubsection{Board Layout}

The GOHR board is a $6 \times 6$ grid indexed from 1 to 36, with four buckets located at the corners. Each grid cell corresponds to a unique position index, while spatial coordinates $(x,y)$ define the true layout.

\begin{table}[htb]
\centering
\setlength{\tabcolsep}{9pt}
\renewcommand{\arraystretch}{1.15}
\begin{tabular}{|c|*{6}{c}|c|}
\hline
\textbf{Bucket 0} & \textbf{Col1} & \textbf{Col2} & \textbf{Col3} & \textbf{Col4} & \textbf{Col5} & \textbf{Col6} & \textbf{Bucket 1} \\
(7,0) & \multicolumn{6}{c|}{} & (7,7) \\
\hline
\textbf{Row6} & 31 & 32 & 33 & 34 & 35 & 36 & \\
\textbf{Row5} & 25 & 26 & 27 & 28 & 29 & 30 & \\
\textbf{Row4} & 19 & 20 & 21 & 22 & 23 & 24 & \\
\textbf{Row3} & 13 & 14 & 15 & 16 & 17 & 18 & \\
\textbf{Row2} & 7  & 8  & 9  & 10 & 11 & 12 & \\
\textbf{Row1} & 1  & 2  & 3  & 4  & 5  & 6  & \\
\hline
\textbf{Bucket 3} & \multicolumn{6}{c|}{} & \textbf{Bucket 2} \\
(0,0) & \multicolumn{6}{c|}{} & (0,7) \\
\hline
\end{tabular}
\caption{Representation of the GOHR board.}
\label{tab:board_representation}
\end{table}

\subsubsection{Pieces, Buckets, and Rules}

Each object on the board is defined by: 
% \pk{WHY USE FINITE IN ONE AND PREDEFINED IN THE OTHER. THE GOHR SUPPORTS ADDED COLOR OR SHAPE DEFINITION ALTHOUGH WE DID NOT USE IT IN THESE EXPERIMENTS.}
\begin{itemize}
    \item \textbf{Shape:} One of several shape categories (e.g., square, circle, triangle, star)
    % One of a finite set (e.g., square, circle, triangle, star)
    \item \textbf{Color:}  One of several color categories (e.g., red, blue, yellow, black)
    % One of a predefined set (e.g., red, blue, yellow, black)
    \item \textbf{Position:} Grid location $(x,y)$
\end{itemize}

At each timestep, the agent selects an action corresponding to selecting an object and 
% \pk{selecting and objects and }
assigning it
% \pk{it}
to one of the four buckets. The validity of this assignment depends on the hidden rule.
Rules can vary in complexity and may depend on: %\pk{DOES THIS INDICATE THAT THE ORDER OF THE MOVES MAY BE IMPORTANT? }\christo{MODIFIED 3RD BULLET}
\begin{itemize}
    \item Single features (e.g., color $\rightarrow$ bucket)
    \item Spatial properties (e.g., quadrant-based placement)
    % \christo{
    \item  Sequential constraints, where the correctness of a move depends on previously accepted moves (e.g., bucket-ordering or reading-order rules)
    % }
    % \item Feature ordering or permutations
    \item Conditional dependencies between attributes
\end{itemize}

\subsubsection{Feedback and Episode Termination}
After each action, the environment provides feedback through status codes:
\begin{itemize}
    \item \textbf{response\_code:}
    \begin{itemize}
        \item 0 (ACCEPT): Valid move.
        \item 4 (DENY): Invalid move.
        \item 7 (IMMOVABLE): Selected object cannot be moved
    \end{itemize}
    \item \textbf{finish\_code:} Indicates the status of the episode : 
    %whether the episode is :
    
    \begin{itemize}
        \item 0 (CONTINUE): The episode continues because movable objects remain on the board.
        \item 1 (FINISH): All objects have been removed from the board.
        \item 2 (STALEMATE): Objects remain on the board, but none can be moved.
    \end{itemize}
    
    % \pk{SOMETHING MISSING}
    \item \textbf{move\_count:} Total number of actions taken so far
\end{itemize}

The reward function is defined as:
\[ r =  \begin{cases}
            0 & \text{if the move is valid (ACCEPT)} \\
            -1 & \text{if the move is invalid (DENY or IMMOVABLE)}
        \end{cases}
\]

An episode terminates when all objects are successfully placed into valid buckets according to the hidden rule.

\subsection{GOHR Environment}
\label{sec:gohr_environment}

All experiments in this work were conducted using the GOHR environment introduced in~\cite{pulick_game_2022,bier_can_2019,bier_gohr_nodate,pulick_comparing_2024}.

%\christo{
The environment consists of a $6 \times 6$ board containing movable objects and four corner buckets. At each timestep, the agent selects an object and assigns it to a bucket. The environment returns the updated board state together with status information indicating whether the move was accepted, denied, or immovable.
%}

%\christo{
The legacy GOHR implementation provides a reinforcement-learning style interface through \texttt{reset()} and \texttt{step()} functions. A distinguishing characteristic of this implementation is that observations include temporal history. In particular, the observation contains the current board state together with several previous successful board states and the corresponding actions that produced them. As a result, the environment provides a history-augmented state representation rather than only the current state.
%}

%\christo{
This differs from standard reinforcement learning environments, where history modeling is typically handled by the agent architecture rather than being embedded directly in the environment. Nevertheless, this legacy observation pipeline was used throughout all experiments reported in this work to maintain consistency with previous GOHR studies.
%}

%\christo{
A Gymnasium-compatible refactoring of GOHR was also developed to improve modularity, support modern reinforcement-learning libraries, and provide a unified interface for both Feature-Centric and Object-Centric representations. Preliminary experiments using the standard Gymnasium observation design, in which observations contain only the current state, did not reproduce the learning behavior obtained with the legacy history-augmented observation pipeline. Consequently, all experimental results reported in this work use the legacy implementation. The Gymnasium refactoring and associated experiments are described in Appendix~\ref{appendix:gymnasium_env}.
%}

% =========================================================
\section{Methodology}

\subsection{Reinforcement Learning Formulation}

We formulate the Game of Hidden Rules (GOHR) as a reinforcement learning problem in which an agent must infer an unobserved rule through interaction with the environment. At each timestep $t$, the agent observes an encoded board state $s_t$, selects an action $a_t$, and receives a reward $r_t$ based on whether the attempted move is accepted by the environment. The hidden rule is not directly observed, making the task partially observable.

The agent is trained to learn a policy $\pi_\theta(a_t \mid s_t)$ that maps encoded states to a distribution over actions. A critic network estimates the value function $V_\phi(s_t)$, which is used to compute advantages for policy optimization. The objective is to learn a policy that minimizes incorrect moves and successfully places all pieces into their correct buckets according to the hidden rule.

\begin{figure}[t]
    \centering
    \includegraphics[width=0.75\linewidth]{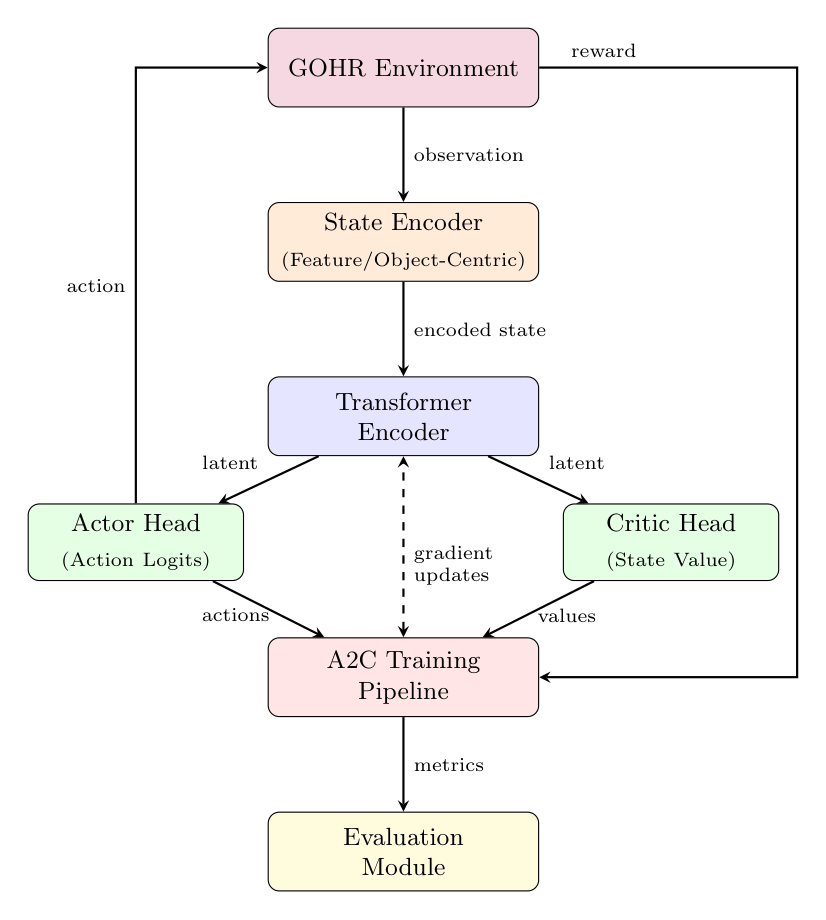}
    \caption{
    Overview of the GOHR reinforcement learning pipeline. The environment produces observations that are encoded into FC or OC state representations, processed by a Transformer encoder, and passed to actor and critic heads for A2C training.
    }
    \label{fig:rl_pipeline}
\end{figure}

Fig.~\ref{fig:rl_pipeline} summarizes the overall training pipeline used in this work, from environment interaction to actor-critic optimization.

\subsection{State and Feature Encoding}

The raw game state returned by the GOHR environment is transformed into one of two representations: a Feature-Centric (FC) representation or an Object-Centric (OC) representation. These encodings differ in how they represent board structure, object identity, and feature relationships.

\subsubsection{Feature-Centric Representation}

In the Feature-Centric representation, the board is encoded as a collection of spatial feature maps over the $6 \times 6$ grid. Each categorical attribute is represented as a separate binary feature map. Specifically, the representation contains eight channels corresponding to four shapes and four colors:

\[
S_{\mathrm{FC}} \in \{0,1\}^{6 \times 6 \times 8}.
\]

\begin{figure}[t]
    \centering
    \includegraphics[width=0.95\linewidth]{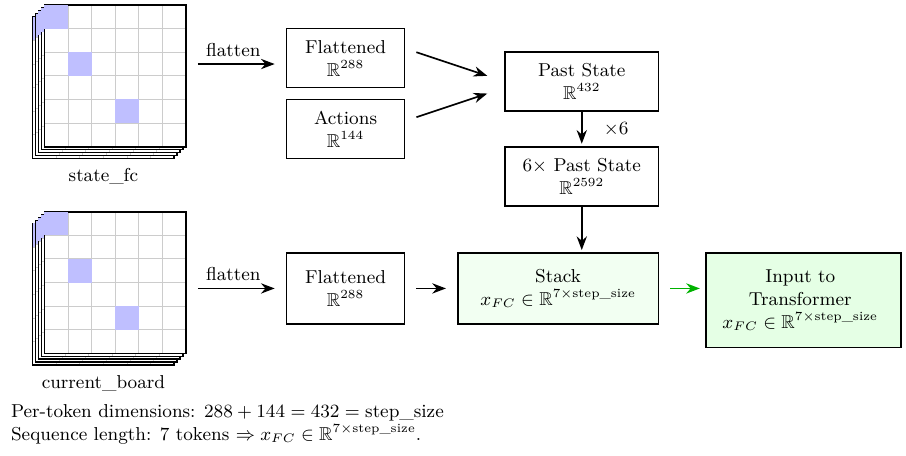}
    \caption{
    Feature-Centric state construction. The current board is encoded into spatial feature maps, where color and shape attributes are represented as separate channels over the $6 \times 6$ grid.
    }
    \label{fig:fc_encoding}
\end{figure}

Fig.~\ref{fig:fc_encoding} illustrates how the current GOHR board state is converted into the Feature-Centric input representation used by the model.

A piece activates one shape channel and one color channel at its board position. For example, a red square at a given grid cell activates both the ``red'' channel and the ``square'' channel at that same position. Thus, color and shape are not explicitly fused into a single object representation; instead, the model must infer their association through spatial alignment across channels.

For a single board state, the FC representation has dimensionality

\[
6 \times 6 \times 8 = 288.
\]

\subsubsection{Object-Centric Representation}

In the Object-Centric representation, each game piece is represented as an individual object vector. Each object is encoded using one-hot representations for color, shape, and position. The object feature vector consists of:

\begin{itemize}
    \item 4 dimensions for color,
    \item 4 dimensions for shape,
    \item 6 dimensions for the $x$-coordinate,
    \item 6 dimensions for the $y$-coordinate,
    % \item 4 dimensions for the bucket/action encoding.
\end{itemize}

Thus, each object is represented as a 20-dimensional vector. 
% \pk{i DON'T UNDERSTAND THIS. THE BUCKET IS A PROPERTY OF THE ACTION, NOT OF THE OBJECT; THE OBJECT HAS ONLY 20 DIMENSIONS? AND I THOUGHT IT HAD AN OBJECT ID [0,...8]}\christo{MODIFIED THE UPCOMING PARAGRAPH REGARDING THE BUCKET. THE INDICES OF OBJECTS ARE NOT PASSED EXPLICITLY BUT BY THE POSITION OF OBJECT IN THE MATRIX. FOR EXAMPLE ROW 0 CORRESPONDS TO OBJECT 1, WHEN THE OBJECT IS REMOVED THIS ROW IS ASSIGNED 0'S} For $n=9$ objects, a single board state is encoded as

% \[
% S_{\mathrm{OC}} \in \mathbb{R}^{9 \times 24}.
% \]
\[
S_{\mathrm{OC}} \in \mathbb{R}^{9 \times 20}.
\]

To incorporate move history, the implementation appends a 4-dimensional bucket encoding to historical object representations. Consequently, the tensors provided to the learning model have dimension $9 \times 24$, where the final four dimensions store the bucket associated with a previous move. For the current board state, these four dimensions are initialized to zero because no move has yet been selected.

\begin{figure}[t]
    \centering
    \includegraphics[width=0.95\linewidth]{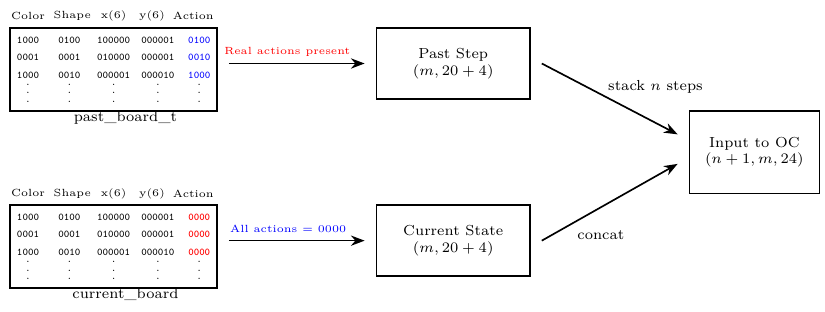}
    \caption{
    Object-Centric state construction. Each game piece is encoded as an individual object vector containing its color, shape, position, and bucket/action-related information.
    }
    \label{fig:oc_encoding}
\end{figure}

Fig.~\ref{fig:oc_encoding} illustrates how the same board state is represented using the Object-Centric encoding.

Unlike the FC representation, the OC representation explicitly binds each object's color, shape, and position into a single vector. Objects are stored in fixed slots corresponding to their object indices, allowing the model to preserve associations among an object’s attributes throughout processing. This reduces the need for the model to infer which color, shape, and position features belong to the same game piece.
% This makes object identity explicit and reduces the burden on the model to infer which attributes belong to the same piece.

\subsubsection{Temporal History Encoding}

A single board state is insufficient to fully infer the hidden rule, since the agent must reason from previous successful and unsuccessful interactions. Therefore, the input includes the current board state along with a fixed-length history of the most recent successful board states and their corresponding actions.

In this work, we use the current state and the six most recent successful states, giving a temporal context length of seven: 
%\pk{CAN WE THINK OF CASES WHERE THIS WILL KILL ANY CHANCE OF LEARNING THE RULE?}

\[
x = [S_t, S_{t-1}, S_{t-2}, \ldots, S_{t-6}].
\]

For the FC representation, previous actions are encoded as one-hot vectors over the position-bucket action space:

\[
|\mathcal{A}_{\mathrm{FC}}| = 36 \times 4 = 144.
\]

Each past FC timestep therefore consists of the flattened board representation and the previous action encoding:

\[
288 + 144 = 432.
\]

The resulting FC input has dimensionality

\[
x_{\mathrm{FC}} \in \mathbb{R}^{7 \times 432}.
\]

For the OC representation, the temporal input is represented as a stack of object-wise feature matrices:

\[
x_{\mathrm{OC}} \in \mathbb{R}^{7 \times 9 \times 24}.
\]

This temporal history allows the model to compare recent accepted moves and infer consistent feature-bucket or object-bucket relationships.

\subsection{Action Space}

At each timestep, the agent selects a piece and assigns it to one of four buckets. The action space differs between the FC and OC representations.

For the FC model, actions are defined over board positions and buckets. Since there are 36 grid positions and 4 buckets, the action space is

\[
|\mathcal{A}_{\mathrm{FC}}| = 36 \times 4 = 144.
\]

For the OC model, actions are defined over objects and buckets. Since there are 9 objects and 4 buckets, the action space is

\[
|\mathcal{A}_{\mathrm{OC}}| = 9 \times 4 = 36.
\]

An action mask is used to prevent the agent from selecting invalid actions, such as actions corresponding to empty board positions or unavailable objects.

\subsection{Reward Function}

The reward function is sparse and negative-only. A successful move receives zero reward, while an incorrect or invalid move receives a penalty:

\[
r_t =
\begin{cases}
0, & \text{if the move is accepted}, \\[4pt]
-1, & \text{if the move is denied or immovable}.
\end{cases}
\]

This reward structure encourages the agent to minimize errors and discover the hidden rule with as few incorrect moves as possible.

\subsection{Transformer-Based Actor-Critic Architecture}
\label{sec:transformer_architecture}

We use a Transformer-based actor-critic architecture for both FC and OC inputs. The Transformer encoder processes the encoded state history and produces a latent representation used by both the policy network and the value network.

\begin{figure}[t]
    \centering
    \includegraphics[width=\linewidth]{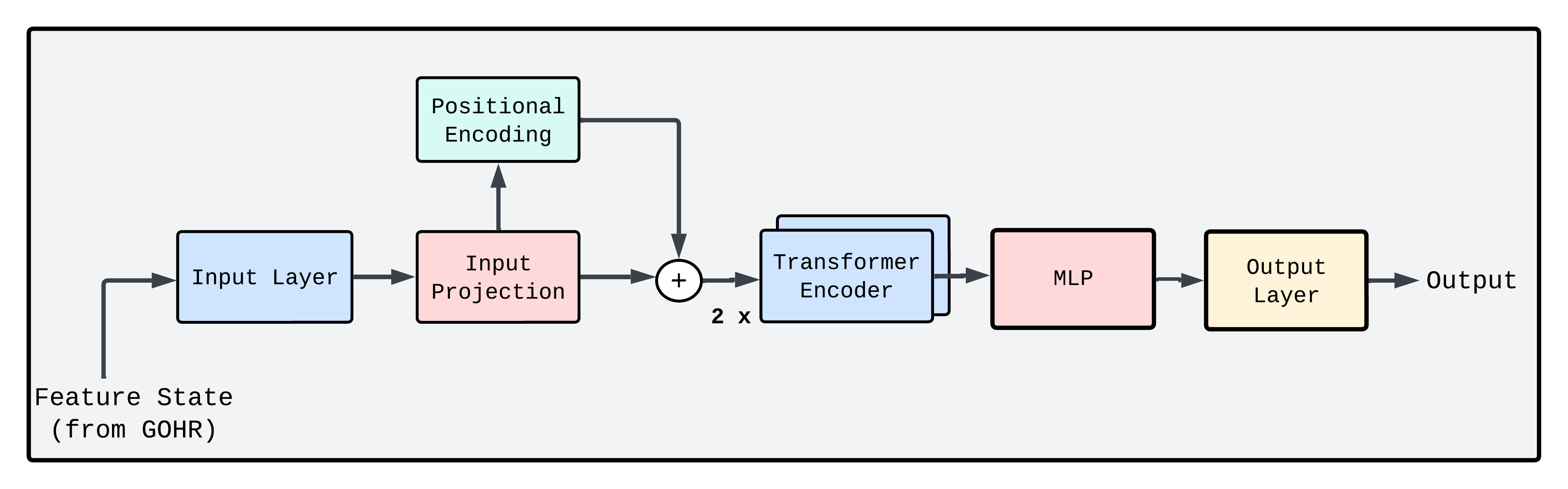}
    \caption{
    Transformer-based architecture used for GOHR policy and value learning. Encoded feature states are projected into an embedding space, combined with positional encodings, passed through Transformer encoder layers, and mapped to output predictions through an MLP and output layer.
    }
    \label{fig:transformer_architecture}
\end{figure}

Fig.~\ref{fig:transformer_architecture} shows the Transformer-based model architecture used to process encoded GOHR state histories.

\subsubsection{Input Projection and Positional Encoding}

The input representation is first projected into a shared embedding dimension $d_{\mathrm{model}}$ using a learned linear projection. Positional and temporal embeddings are added to preserve ordering information.

For the FC model, each timestep in the temporal history is treated as a token. For the OC model, each object within each timestep is treated as an object-level token. Objects occupy fixed positions in the input tensor according to their object indices, while temporal embeddings encode the ordering of states in the history.
% , with embeddings used to encode both object identity and temporal position.

\subsubsection{Transformer Encoder}

The projected tokens are passed through a Transformer encoder consisting of multi-head self-attention and feedforward layers (\cite{vaswani2017attention}). Self-attention allows the model to capture dependencies among features, objects, positions, and previous actions.

For an input sequence $X$, multi-head self-attention is computed as

\[
\mathrm{MHSA}(X) = \operatorname{Concat}(h_1,\ldots,h_H)W^O,
\]

where each attention head is given by

\[
h_i = \operatorname{softmax}
\left(
\frac{Q_iK_i^\top}{\sqrt{d_k}}
\right)V_i.
\]
%\pk{ARE Q,K,V DEFINED? OR A REFERENCE FOR THE NOTATION?}
Here  \(Q_i\), \(K_i\), and \(V_i\) denoting the query, key, and value projections for head \(i\), and \(d_k\) the dimensionality of the key vectors.
This mechanism enables the model to learn relationships that are important for hidden-rule inference, such as color-bucket mappings, shape-bucket mappings, spatial ordering, and object-level dependencies.

\subsubsection{Actor Head}

The actor head maps the Transformer output representation to action logits. These logits are masked using the valid action set and converted into a categorical policy distribution:

\[
\pi_\theta(a_t \mid s_t) = \operatorname{softmax}(W_\pi z + b_\pi),
\]

where $z$ is the aggregated Transformer representation. During training, actions are sampled from this distribution, while during evaluation the action with the highest probability may be selected.

\subsubsection{Critic Head}

The critic head maps the same Transformer representation to a scalar value estimate:

\[
V_\phi(s_t) = W_v z + b_v.
\]

This value estimate is used to compute the advantage function for actor-critic training.

\subsection{A2C Learning Algorithm}
\label{sec:a2c_algorithm}

The model is trained using Advantage Actor-Critic (A2C)~\cite{mnih2016asynchronous}. After each episode, discounted returns are computed as

\[
G_t = \sum_{k=0}^{T-t-1} \gamma^k r_{t+k}.
\]

where $\gamma$ is the discount factor.

The advantage estimate is then computed as

\[
A_t = G_t - V_\phi(s_t).
\]

The critic is trained by minimizing the mean-squared error between predicted values and empirical returns:

\[
L_{\mathrm{critic}}(\phi)
=
\frac{1}{N}
\sum_t
\left(G_t - V_\phi(s_t)\right)^2.
\]

The actor is trained using an entropy-regularized policy gradient loss:

\[
L_{\mathrm{policy}}(\theta)
=
-\frac{1}{N}
\sum_t
\left(
\log \pi_\theta(a_t \mid s_t) A_t
+
\beta H_t
\right),
\]

where $H_t$ is the entropy of the policy distribution and $\beta$ is the entropy regularization coefficient. The entropy term encourages exploration and prevents the policy from becoming prematurely deterministic.

%\christo{
In addition to policy sampling, exploration is further encouraged through an $\epsilon$-greedy strategy. At training step $t$, the exploration probability is

\[ \epsilon_t
=
\epsilon_{\mathrm{end}}
+
(\epsilon_{\mathrm{start}}-\epsilon_{\mathrm{end}})
\exp\left(
-\frac{t}{\epsilon_{\mathrm{decay}}}
\right),
\]

%\christo{
where $\epsilon_{\mathrm{start}}$, $\epsilon_{\mathrm{end}}$, and $\epsilon_{\mathrm{decay}}$ control the initial exploration rate, final exploration rate, and exploration decay schedule, respectively. With probability $\epsilon_t$, the agent selects a random valid action; otherwise, it follows the learned policy distribution.

\subsection{Evaluation Metrics}

We evaluate learning performance using three convergence metrics: $M^\star$, $E^\star_{\mathrm{mean}}$, and $E^\star_{\mathrm{max}}$. These metrics quantify how quickly and reliably the agent learns a hidden rule.

\begin{enumerate}
    \item \textbf{\texorpdfstring{$M^\star$}{M*}}:
    
    The move-based convergence metric $m^\star$ identifies the first move after which the agent makes only successful moves for a fixed window of length $W_{m^\star}$:
        
    \[
    m^\star
    =
    \min
    \left\{
    m \in \mathbb{N}
    :
    \forall i \in [m, m + W_{m^\star} - 1],
    \;
    \mathtt{response\_code}(i)=\mathtt{A}
    \right\}.
    \]
        
    Here, $\mathtt{A}$ denotes an accepted move. The aggregated metric $M^\star$ is computed as the median of $m^\star$ across multiple training runs.
    
    \item \textbf{\texorpdfstring{$E^\star_{\mathrm{mean}}$}{E*mean}}:
    
    The mean-window convergence metric $e^\star_{\mathrm{mean}}$ identifies the earliest episode at which the average error rate within a sliding window of size $W_{\mathrm{mean}}$ falls below a threshold $T_{\mathrm{mean}}$:
        
    \[
    e^\star_{\mathrm{mean}}
    =
    \min
    \left\{
    t
    :
    \frac{1}{W_{\mathrm{mean}}}
    \sum_{k=0}^{W_{\mathrm{mean}}-1}
    E_{t+k}
    \leq
    T_{\mathrm{mean}}
    \right\}.
    \]
        
    The aggregated metric $E^\star_{\mathrm{mean}}$ is computed as the median of $e^\star_{\mathrm{mean}}$ across runs.
    
    \item \textbf{\texorpdfstring{$E^\star_{\mathrm{max}}$}{E*max}}:
    
    The max-window convergence metric $e^\star_{\mathrm{max}}$ is stricter than the mean-window metric. It identifies the earliest episode at which the maximum error rate within a sliding window remains below a threshold $T_{\mathrm{max}}$:
        
    \[
    e^\star_{\mathrm{max}}
    =
    \min
    \left\{
    t
    :
    \max_{0 \leq k < W_{\mathrm{max}}}
    E_{t+k}
    \leq
    T_{\mathrm{max}}
    \right\}.
    \]
    
    The aggregated metric $E^\star_{\mathrm{max}}$ is computed as the median of $e^\star_{\mathrm{max}}$ across runs.

\end{enumerate}

Together, these three metrics capture complementary aspects of learning: move-level stability, average episode-level convergence, and worst-case episode-level robustness.

\begin{figure}[htbp]
    \centering
    \includegraphics[
        width=0.9\linewidth,
        alt={Heatmap matrix of Spearman correlations between evaluation metrics. Rows and columns correspond to different evaluation metrics. Darker shading indicates stronger positive correlation.}
    ]{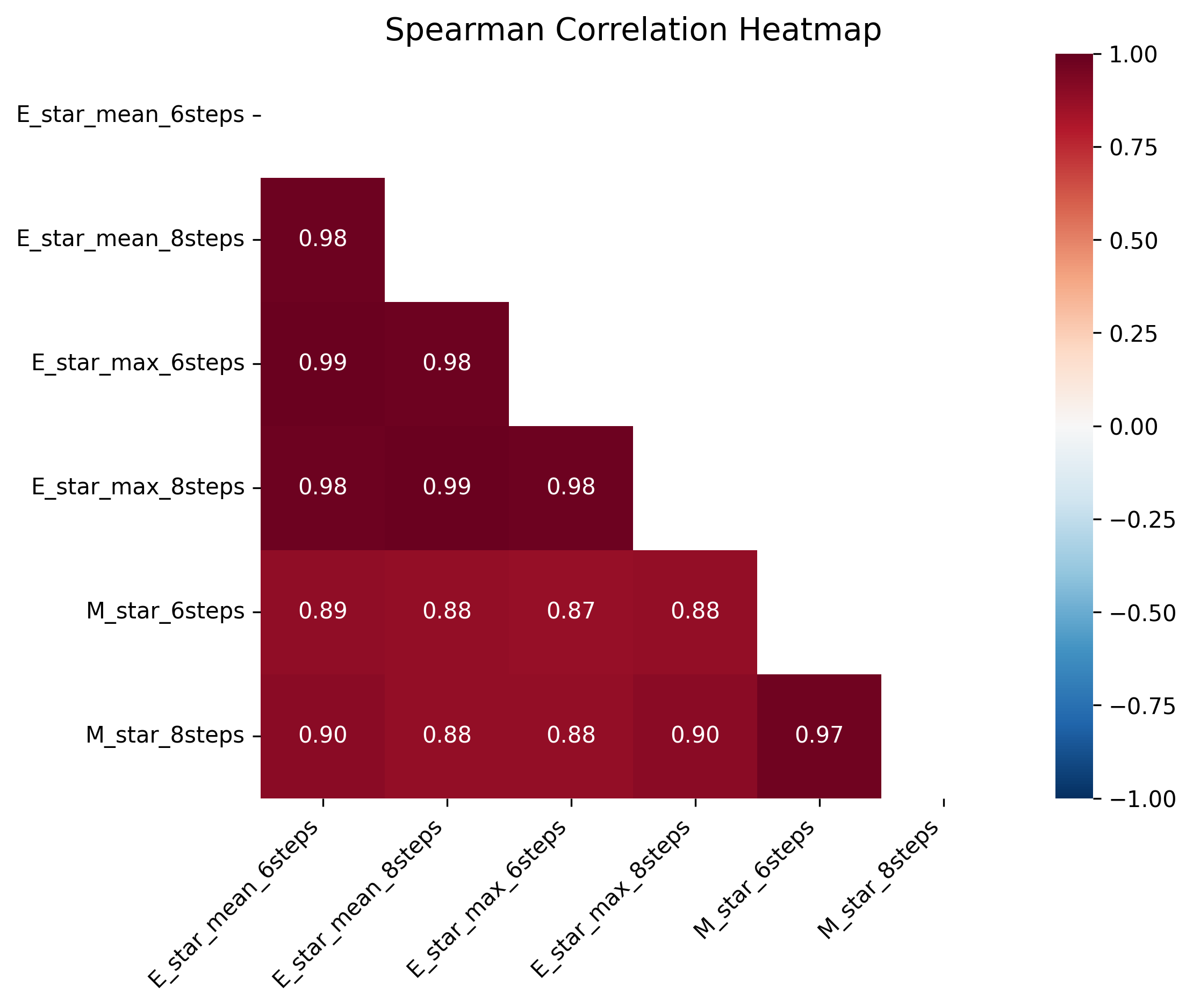}
    \caption{
    Heatmap of Spearman correlations between evaluation metrics evaluated using 6-step and 8-step memory windows.
    }
    \label{fig:spearman_heatmap}
\end{figure}

Fig.~\ref{fig:spearman_heatmap} presents the Spearman correlation heatmap between the evaluation metrics across 18 rules using both 6-step and 8-step memory windows. 
% The corresponding numerical correlation values are provided in Table~\ref{tab:spearman-table}.

% =========================================================

% =========================================================
\section{Environmental and Training Setup}

\subsection{Environment and Rule Suite}

All experiments were conducted in the GOHR $6 \times 6$ environment with nine objects and four buckets. At the beginning of each episode, a hidden rule determines the correct bucket assignment for each object. The agent does not observe this rule directly and must infer it through interaction with the environment.

The experiments evaluate the agent across a suite of hidden rules that vary in both difficulty and structural complexity, including feature-based, spatial, ordering-based, and relational rules. The complete list of rules used in the experiments, together with their descriptions, is provided in Appendix~\ref{appendix:rules}.

\subsection{Model Configuration and Hyperparameters}

Both the Feature-Centric (FC) and Object-Centric (OC) agents use the Transformer-based A2C architecture described in Sections~\ref{sec:transformer_architecture}--~\ref{sec:a2c_algorithm}. The two models differ only in their state representation and action-space formulation.

The hyperparameters used across the reinforcement learning experiments are summarized in Table~\ref{tab:hyperparameters}.

\begin{table}[htb]
\centering
\caption{Training hyperparameters used across the reinforcement learning experiments.}
\label{tab:hyperparameters}
\begin{tabular}{@{} l l @{}}
\toprule
\textbf{Parameter} & \textbf{Value} \\
\midrule
Algorithm                        & A2C \\
Model                            & Transformer-based actor-critic \\
Learning rate $\alpha$           & $1 \times 10^{-5}$ \\
Discount factor $\gamma$         & $0.001$ \\
$\epsilon_{\mathrm{start}}$      & $0.5$ \\
$\epsilon_{\mathrm{end}}$        & $0.0001$ \\
$\epsilon_{\mathrm{decay}}$      & $200$ \\
Batch size                       & $1$ \\
Initial number of objects        & $9$ \\
Maximum episodes per run         & $10{,}000$ \\
Number of independent runs       & $5$ \\
Temporal history length          & Current state + 6 previous successful states \\
\bottomrule
\end{tabular}
\end{table}
%\pk{HAVE WE SAID WHAT THE EPSILON PARAMETERS ARE? I DID NOT SEE AN $\alpha$ AND THERE IS A $\beta$ NOT DEFINED.}\christo{ADDED EXPLANATION OF THE EPSILON, ALPHA AND BETA}
Here, $\alpha$ denotes the learning rate used by the optimizer, $\gamma$ is the discount factor, and $\beta$ is the entropy regularization coefficient used in the policy loss.
We use a small discount factor, $\gamma = 0.001$, which places most of the learning signal on immediate rewards. This choice reflects the structure of GOHR, where rewards are sparse and negative-only, and the primary objective is to minimize incorrect moves rather than optimize long-horizon cumulative reward. %\pk{SHOULD WE SAY SOMETHING ABOUT THE FACT THAT THIS IS NOT QUITE CONSISTENT WITH OUR USE OF 10 GOOD MOVES, OR K PERFECT BOARDS AS A METRIC. BECAUSE THE AI IS NOT BEING TRAINED TO DO WHAT WE ARE EVALUATING?}\christo{ADDED NEXT PARAGRAPH}

The evaluation metrics described in Section~\ref{sec:a2c_algorithm} measure the extent to which the agent exhibits stable rule-following behavior. Metrics such as $M^\star$, $E^\star_{\mathrm{mean}}$, and $E^\star_{\mathrm{max}}$ quantify the consistency of correct actions across extended sequences of moves and episodes. Although these metrics are not optimized directly by the A2C objective, successful performance on them indicates that the agent has acquired the underlying hidden rule and can apply it reliably over time.

\subsection{Training and Evaluation Protocol}

Unless otherwise stated, each experiment was repeated for five independent runs with different random seeds. 
Each run was trained until convergence according to the evaluation metrics introduced in Section~\ref{sec:a2c_algorithm}:

\begin{enumerate}
    \item the agent achieves at least $W_{m^\star}=10$ consecutive successful moves, corresponding to the move-level convergence criterion used to define $m^\star$;

    \item the mean-window error rate satisfies
    \[
    e^\star_{\mathrm{mean}} < 0.25
    \]
    over a sliding window of $W_{\mathrm{mean}}=10$ episodes;

    \item the max-window error rate satisfies
    \[
    e^\star_{\mathrm{max}} < 0.25
    \]
    over a sliding window of $W_{\mathrm{max}}=5$ episodes.
\end{enumerate}

These criteria were used jointly to avoid premature convergence. The episode-level metrics evaluate whether the agent has reached stable low-error behavior over time, while the move-level metric verifies the ability to sustain consecutive sequences of correct actions.

\section{Experiments and Results}
The experiments using the GOHR environment encompass two primary configurations:

\begin{enumerate}
    \item \textbf{Rule-based configuration:} In this configuration, a rule file specifying the rule definition is provided to the game server.
    \item \textbf{Trial-list-based configuration:} In this configuration, a trial list is supplied, specifying both the rule set and the initial board generation process. This method also supports transfer experiments. 
\end{enumerate}

\subsection{Independent Rule Experiments}

Eighteen different rule files were considered for these experiments. The rules used are listed in Table~\ref{tab:rule_based_exp} . Each rule was used to train a model independently, in order to assess model performance on isolated rule learning.

For the analysis of rule difficulty, we initially attempted to analyze the difficulty of
% \pk{the difficulty of } 
each rule separately. However, slight variations in metrics and the substantial overlap between metrics made it difficult to rank the rules clearly. 

% Therefore, we decided to characterize
% % \pk{characterize }
% %decompose 
% the rules according to
% % \pk{according to}
% % into 
% their individual conceptual
% % \pk{conceptual }
% properties (see Table~\ref{tab:rule-aspect-mapping}), which enabled us to better understand how the model performs for each {property} and identify their respective strengths and weaknesses. 
% \pk{IS ``THEIR'' STRENGTHS OF THE PROPERTIES, OR OF THE WAYS FOR LEARNING?}\christo{MODIFIED PARAGRAPH BELOW}

Therefore, we characterized the rules according to their individual conceptual properties (see Table~\ref{tab:rule-aspect-mapping}). This allowed us to evaluate performance across different classes of rule properties and determine which properties were easier or more difficult for the learning system to acquire.

We have identified the following {properties} for analysis:
\begin{enumerate}
\item \textbf{Quadrant\_to\_bucket\_mapping}: Mapping of pieces in a specified quadrant to a particular bucket.
\item \textbf{Proximity}: Removing pieces based on proximity to any bucket. ``Farthest'' indicates pieces at the center of the board, while ``nearest'' indicates pieces in any of the corners.
\item \textbf{Reading\_order}: Removing pieces in reading order (or reverse reading order), that is, from left to right and top to bottom (or right to left and bottom to top).
\item \textbf{Feature\_to\_bucket\_mapping}: Mapping each color/shape to any bucket.
\item \textbf{Feature\_ordering}: Removing one piece of each color/shape in order (e.g., blue after red, black after blue).
\item \textbf{All\_pieces\_of\_feature}: Removing all pieces of one color/shape, then proceeding to the next.
\item \textbf{Bucket\_ordering}: Assigning pieces to a specified order of buckets.
\item \textbf{Conditional}: Skipping pieces based on specified conditions.
\end{enumerate}

All rules can be characterized as using
% \pk{using }
either one of these properties or a combination of multiple properties.

\begin{table}[htb]
\centering
\caption{Mapping of Rules to Their Corresponding {Properties}}
\label{tab:rule-aspect-mapping}
\begin{tabularx}{\textwidth}{lX}
\toprule
\textbf{Rule Name} & \textbf{Properties} \\
\midrule
cm\_RBKY & Feature\_to\_bucket\_mapping \\
sm\_csqt & Feature\_to\_bucket\_mapping \\
allOfColOrd\_BRKY & All\_pieces\_of\_feature  \\
allOfShaOrd\_qcts & All\_pieces\_of\_feature \\
col1Ord\_BRKY & Feature\_ordering \\
col1Ord\_KRBY & Feature\_ordering \\
colOrdL1\_BRKY & Feature\_ordering + Conditional \\
sha1Ord\_qcts & Feature\_ordering \\
shaOrdL1\_qcts & Feature\_ordering + Conditional \\
col1OrdBuck\_BRKY0213 & Feature\_to\_bucket\_mapping + Feature\_ordering \\
sha1OrdBuck\_qcts0213 & Feature\_to\_bucket\_mapping + Feature\_ordering \\
ordL1 & Reading\_order \\
ordRevOfL1 & Reading\_order \\
ordL1\_Nearby & Reading\_order + Proximity \\
ordRevOfL1\_Remotest & Reading\_order + Proximity \\
cw & Bucket\_ordering \\
ccw & Bucket\_ordering \\
cw\_0123 & Bucket\_ordering \\
cw\_qn2 & Quadrant\_to\_bucket\_mapping + Bucket\_ordering \\
quadNearby & Quadrant\_to\_bucket\_mapping \\
quadMixed1 & Quadrant\_to\_bucket\_mapping \\
\bottomrule
\end{tabularx}
\end{table}

\subsubsection{Difficulty Analysis of Rule {Properties}}
% \pk{Our analysis makes use of the notion of ``degree of abstraction.'' We define these degrees both intuitively and by observation of the difficulty of various rules. }
Our analysis makes use of the notion of ``degree of abstraction.'' We define these degrees both intuitively and by observation of the difficulty of various rules.
We observed a clear trend in the difficulty of learning each rule, according to its conceptual  properties. The difficulty {is largely determined by the degree of abstraction in the feature underlying each property. } 
Even when considering features at the same level of abstraction, the
% \pk{the } 
mapping to buckets is generally easier to learn than an allowed
% \pk{an allowed } 
ordering based on features.

Based on our experiments and observations, {we establish the relative ordering of difficulty for rule properties in the FC and OC models, providing insight into how each representation influences learning performance.} The experiment results can be found in  Tables~\ref{tab:aspects_FC} and ~\ref{tab:aspects_OC}. 

% When comparing {properties}, we primarily relied on $M^\star$ values. $m^\star$ is a step-level measure, defined directly in terms of whether individual moves obey the rule properties and requiring that there be a string of correct moves beginning at move $m^\star$.
% % \pk{and requiring that there be a string of correct moves begining at move $m^\star$.}
% This makes it straightforward to compute, even for properties that are not well-defined on every move (for example, bucket\_ordering rules, where correctness/performance can only be evaluated after the first bucket placement has established a reference point, or conditional properties, which only apply when specific preconditions are met). 
%\pk{I CAN;T QUITE UNDERSTAND WHAT THIS MEANS.}

% By contrast, $e^\star_{mean}$ and $e^\star_{max}$ are episode-level measures that require computing per-episode correctness rates. For properties that only become meaningful in certain moves, many episodes provide few opportunities to evaluate them, making these statistics noisier and harder to interpret. 
%\pk{OK, MAYBE I SEE WAHT YOU ARE GETTING AT. IT IS SOMETHING ABOUT HOW OFTEN ONE CAN SEE WHETHER THE PROPERTY MATTERS? } \christo{MODIFIED THE PARAGRAPHS. MODIFIED PARAGRAPHS IS WRITTEN BELOW.}

When comparing rule properties, we primarily relied on $M^\star$ values. Unlike $E^\star_{\mathrm{mean}}$ and $E^\star_{\mathrm{max}}$, which aggregate performance over entire episodes, $M^\star$ is defined directly in terms of sequences of correct moves. Consequently, it provides a more direct measure of when the agent begins to apply a rule property consistently.

This distinction is particularly important because different rule properties influence different subsets of moves. For example, conditional properties are relevant only when their associated preconditions are satisfied, while some ordering-based properties become informative only after earlier actions have established the appropriate context. As a result, many moves within an episode may provide little information about whether a particular property has been learned. Episode-level error rates therefore tend to be noisier and less directly tied to the acquisition of a specific property. By focusing on sustained sequences of correct actions, $M^\star$ provides a more interpretable measure of property-level learning and facilitates comparisons across different classes of rule properties.

%\christo{
By contrast, $E^\star_{\mathrm{mean}}$ and $E^\star_{\mathrm{max}}$ remain useful complementary measures of convergence because they quantify average and worst-case episode-level performance. However, for the purpose of comparing the relative difficulty of individual rule properties, $M^\star$ provides the most direct indication of when a property has been acquired and applied consistently.

\paragraph{FC Model:}
In the FC model, positional properties are the easiest to learn (lowest level of abstraction), followed by properties that depend on piece features (higher level of abstraction).
\begin{enumerate}
    \item \textbf{Easiest properties}: Quadrant\_mapping, proximity, and reading\_order were the most accessible properties for the FC model to learn. This can be attributed to their primary dependence on piece positions, which the FC model captures most effectively. Among these, quadrant\_mapping was the easiest, followed by proximity with nearly identical learning curves, while reading\_order proved more challenging.

    \item \textbf{Moderate Difficulty}: Feature\_to\_bucket mapping follows next, depending on individual piece features such as color or shape being mapped to output buckets. The all\_pieces\_of\_feature property, which also depends on piece features, exhibits similar difficulty.

    \item \textbf{Higher Difficulty}: Bucket\_ordering proved more challenging, as it depends on an abstract ordering of buckets that is difficult for the FC model representation to capture.

    \item \textbf{Most Challenging}: Feature\_ordering and conditional properties represent the most challenging categories for the FC model, exhibiting very slow learning curves.

\end{enumerate}

\paragraph{OC Model:}

While the FC model showed significant differences in difficulty between properties, the OC model exhibited relatively smaller differences in difficulty and learning curves between properties.

\begin{enumerate}
    \item \textbf{Easiest properties}: Feature\_to\_bucket mapping, all\_pieces\_of\_feature, and quadrant\_mapping were most accessible for the model to learn. These properties depend on the lowest level of feature abstraction, directly utilizing color, shape, or x,y coordinates. These properties demonstrated nearly identical learning curves with minimal differences. Notably, quadrant\_mapping was the most challenging among these three, which can be attributed to its dependence on two features (x and y coordinates) rather than a single feature.

    \item \textbf{Moderate Difficulty}: Bucket\_ordering proved slightly more challenging than the easiest properties.

    \item \textbf{Higher Difficulty}: Reading\_order and proximity exhibited slower learning curves compared to the above properties. This can be attributed to the higher level of feature abstraction, as the model must derive positional information from the provided x and y coordinates.

    \item\textbf{Most Challenging}: Consistent with the FC model, feature\_ordering and conditional properties remained the most challenging for the OC model to learn, exhibiting very slow learning curves.

\end{enumerate}

\subsubsection{Rule Difficulty Analysis}

Having established the relative difficulty ordering for individual rule properties, we now examine the difficulty ordering of complete rules. 

{Rules ordered by increasing difficulty are presented in Tables~\ref{tab:FC_difficulty} and~\ref{tab:OC_difficulty}. The rules are grouped into categories according to the ranges of their metric values. While the ordering of rules within each category is somewhat ambiguous and may vary depending on the chosen metric, we consistently observe a clear relative ordering across categories. This relative structure remains stable across all three metrics.}

\paragraph{FC Model}

\begin{enumerate}
    \item \textbf{Highly Learnable}: The most accessible rules are the quadrant\_mapping rules: quadNearby and quadMixed1.

    \item \textbf{Moderately Learnable}: Rules based on proximity and reading\_order follow: ordL1\_Nearby, ordL1, ordRevOfL1, and ordRevOfL1\_Remotest.

    \item \textbf{Challenging}: Rules depending on feature\_to\_bucket mapping and all\_pieces\_of\_feature properties include cm\_RBKY, sm\_csqt, allOfShaOrd\_qcts, and allOfColOrd\_BRKY.

    \item \textbf{More Challenging}: The rules col1OrdBuck\_BRKY0213 and sha1OrdBuck\_qcts0213, which are primarily feature\_to\_bucket mapping rules, show slower learning due to the additional feature ordering component. The ccw and cw rules demonstrate comparable difficulty to these combined rules.

    \item \textbf{Highly Challenging}: Rules depending on feature\_ordering properties (sha1Ord\_qcts and col1Ord\_BRKY) are very difficult to learn.

    \item \textbf{Most Challenging}: When conditional properties are added to feature\_ordering, the rules become most challenging for the model. This category includes colOrdL1\_BRKY and shaOrdL1\_qcts.

\end{enumerate}

\begin{figure}[htbp]
  \centering
  \includegraphics[width=\linewidth]{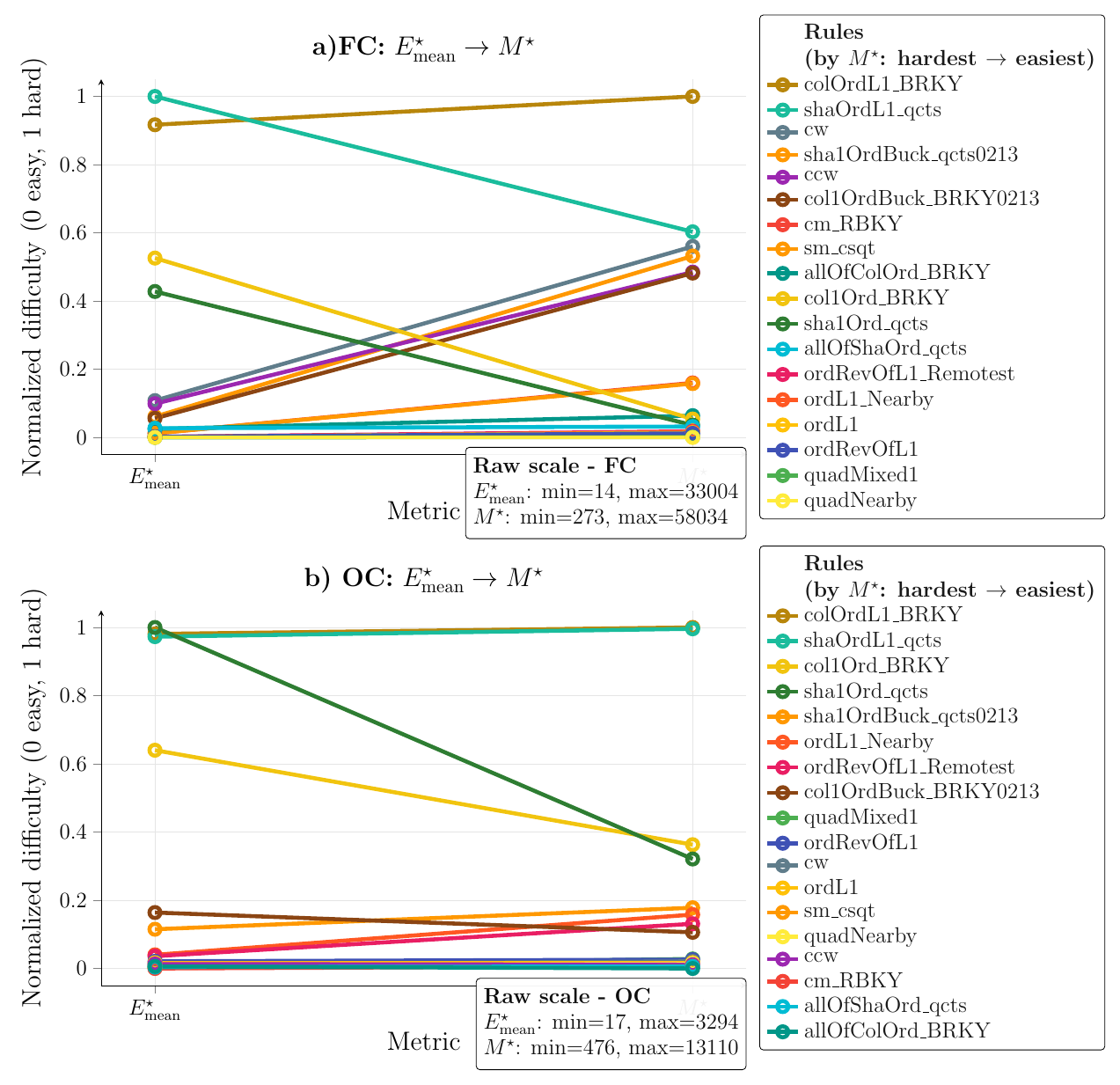}
  \caption{Crossings plots for rule difficulty between different metrics. 
    Each line connects a rule's normalized difficulty score in the  $E^\star_{mean}$ (left) to the $M^\star$ (right). 
    Labels are displayed in the order of their $M^\star$ values ;color of rule label corresponds to the color of the line it represents; raw scales (min, median, max per model) are shown inside each panel. A scale including the minimum, median and max values of each metrics is included. 
    (a) shows the comparison of metrics in FC model and (b) shows the comparison of the metrics in OC model.} 
  \label{fig:fc_oc_diff_analysis}
\end{figure}

\paragraph{OC Model}

In the OC model, rule difficulties are more closely clustered, making ordering more difficult based on final metrics, with apparent crossings between metrics.
\begin{enumerate}
    \item \textbf{Learnable}: The most accessible rules include allOfColOrd\_BRKY, allOfShaOrd\_qcts, cm\_RBKY, sm\_csqt, quadNearby, quadMixed1, cw, ccw, ordL1 and ordRevOfL1. These rules primarily rely on properties such as feature\_to\_bucket mapping, all\_pieces\_of\_feature removal, quadrant\_mapping, and bucket\_ordering. Among them, ordL1 and ordRevOfL1, which are based on reading\_order, show slightly higher difficulty but remain close to the others in this group and can still be considered part of the learnable category.

    \item \textbf{Moderately Challenging}: When proximity is added to the reading properties, rules become more challenging to learn, including ordRevOfL1\_Remotest and ordL1\_Nearby. The rules col1OrdBuck\_BRKY0213 and sha1OrdBuck\_qcts0213, which combine feature\_to\_bucket mapping with feature\_ordering, demonstrate comparable difficulty to reading+proximity rules.

    \item \textbf{Most Challenging}: Similar to the FC model, rules depending on feature\_ordering are the most difficult, including col1Ord\_BRKY, sha1Ord\_qcts, colOrdL1\_BRKY, and shaOrdL1\_qcts. The addition of conditional properties makes colOrdL1\_BRKY and shaOrdL1\_qcts the most challenging rules for the model to learn.

\end{enumerate}
\subsubsection{Key Observations}

{
In Fig.~\ref{fig:fc_oc_diff_analysis}(a) we observe numerous crossings between the %$e^\star_{\text{mean}}$
$E^\star_{\text{mean}}$ and $M^\star$ rankings. Most crossings occur within or between closely related rule properties categories (as defined in the Rules/Rule Property Difficulty Analysis), indicating that 
% in the FC representation the relative difficulty is metric-sensitive: rules can swap order depending on whether we emphasize learning speed ($e^\star_{\text{mean}}$) or final convergence ($m^\star$). 
 the relative difficulty of rules in the FC representation is sensitive to the choice of convergence metric. Rankings derived from episode-level error-rate windows ($E^\star_{\mathrm{mean}}$) do not always agree with those derived from move-level sequences of correct actions ($M^\star$).

%\pk{IS IT OBVIOUS THAT ONE IS ABOUT LEARNING SPEED AND THE OTHER ABOUT FINAL CONVERGENCE. m\_star IS NOT REALLY FINAL - IT IS 'FIRST OCCURRENCE.' WE USE IT BECAUSE IT IS USEFUL FOR HUMANS. }\christo{MODIFIED THE ABOVE SENTENCE} 
This effect is especially visible among feature-dependent properties (feature\_to\_bucket, all\_pieces\_of\_feature, feature\_ordering), whereas the easiest positional rule properties (e.g., quadrant/proximity/reading\_order) and the hardest feature\_ordering/conditional rule properties tend to preserve their extremes of the ranking.}

{By contrast, Fig.~\ref{fig:fc_oc_diff_analysis}(b) shows substantially fewer crossings, yielding a more stable rule ordering across metrics. This aligns with our earlier observation that 
% OC exhibits smaller between-difficulty gaps: 
 OC representation exhibits smaller gaps between rule properties:feature\_to\_bucket, all\_pieces\_of\_feature, and quadrant\_to\_bucket\_mapping cluster together on the
%easy
easier side, while feature\_ordering and conditional rules remain hardest, with the same relative ordering under both metrics. In short, OC provides a more coherent notion of difficulty across measures, whereas FC shows larger metric-driven variability, primarily within or at the boundaries of related categories.}  %\pk{GOOD!}

The complete results of the independent rule experiments are available at: \url{https://tinyurl.com/4vfjp6yn}.
%\pk{IS THIS THE FINAL DATA? }\christo{YES}

\subsection{Generalization Analysis}
\label{sec:generalization}

We conducted a set of experiments to evaluate how well the learned representations generalize beyond the training distribution. The Feature-Centric (FC) and Object-Centric (OC) models were evaluated under different forms of distribution shift aligned with the inductive biases of their respective representations.

% ============================================================
\subsubsection{Feature-Centric (FC) Generalization}
\label{sec:fc_generalization}
% ============================================================

\begin{figure}[htp]
    \centering
    \includegraphics[width=0.4\linewidth,
    alt={Checkerboard diagram indicating the positional split used during training. White squares represent locations available during training, while the remaining positions are excluded and used only during testing.}]
    {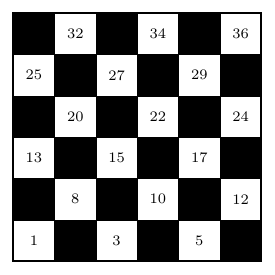}
    \caption{Checkerboard positional split used for FC generalization experiments. White squares indicate positions available during training.}
    \label{fig:checkerboard}
\end{figure}

For the FC model, generalization was evaluated using the \texttt{train} and \texttt{test} modes provided by the GOHR environment. In \texttt{train} mode, objects were restricted to a predefined subset of board positions, whereas in \texttt{test} mode at least one object was placed in a location not encountered during training. This setup evaluates whether the FC representation learns position-invariant rule structure or instead relies on memorizing spatial activation patterns.

To ensure broad spatial coverage, we adopted the checkerboard positional split shown in Fig.~\ref{fig:checkerboard}. During training, objects were placed only on the white squares, while testing allowed placements across the full board.

\begin{figure}[htb]
    \centering
    \includegraphics[width=\textwidth,
    alt={Comparison of Feature-Centric in-distribution and out-of-distribution error rates across rules. The model is trained using restricted board positions and evaluated on unrestricted placements. Out-of-distribution errors increase substantially for most rules, especially spatially dependent rules.}]
    {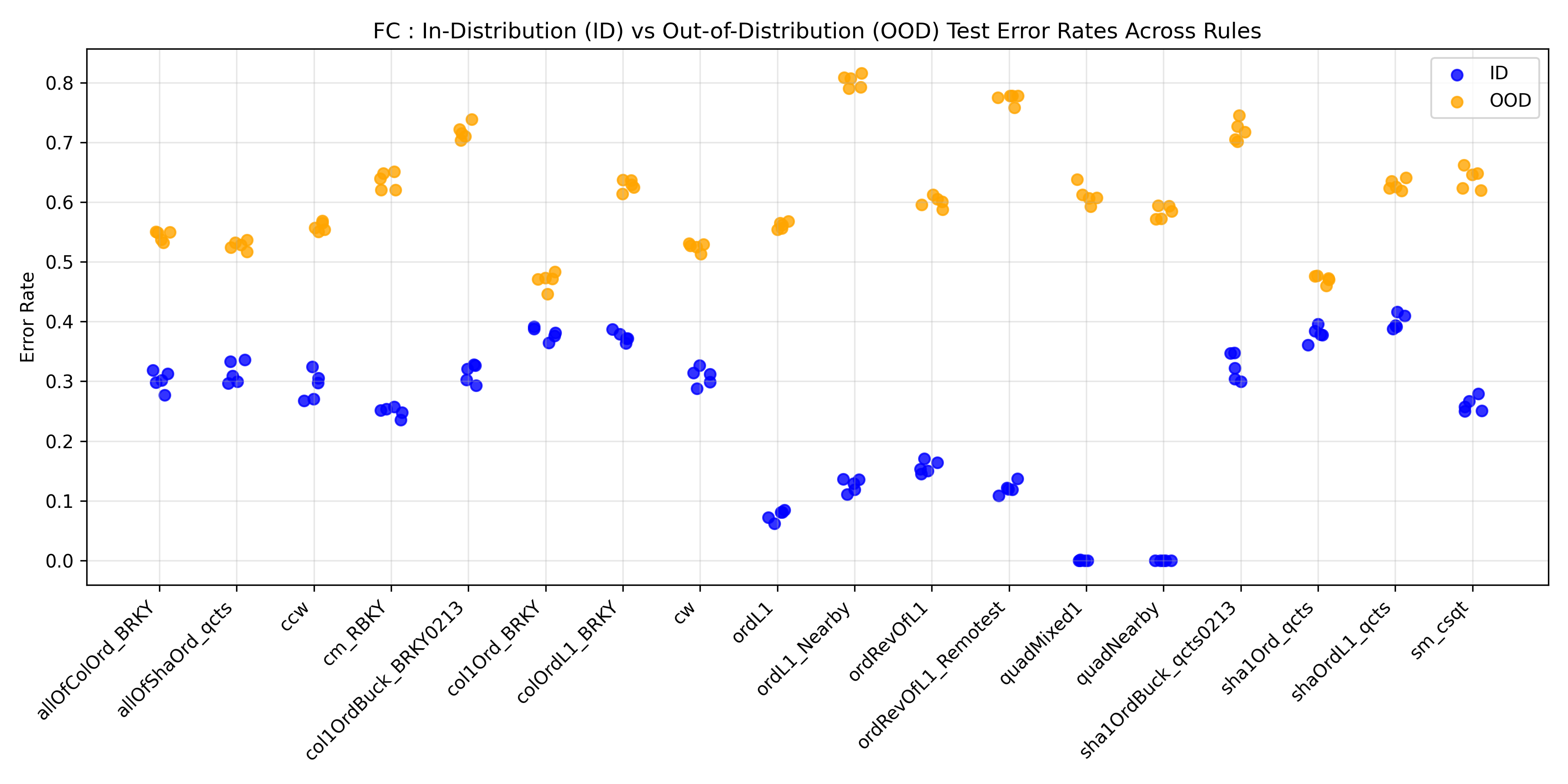}
    \caption{
    Feature-Centric (FC) generalization performance.
    In-distribution (ID) and out-of-distribution (OOD) error rates across rules. The FC model is trained using restricted checkerboard positions and evaluated on unrestricted placements. OOD errors increase substantially for most rules, indicating strong dependence on positional patterns observed during training.}
    \label{fig:fc_outofdist}
\end{figure}

Fig.~\ref{fig:fc_outofdist} presents the ID and OOD error rates across rules. The positional shift produces a substantial degradation in performance. While ID error rates typically remain within the 25--35\% range, OOD errors increase sharply for most rules, commonly reaching 55--75\%.

Several rules exhibit particularly severe failures. The rules
% \texttt{colOrderBRKY0123}, 
\texttt{quadMixed1}, \texttt{quadNearby} and reading rules like \texttt{ordL1}, \texttt{ordL1\_Nearby}
exceed 80\% OOD error. These rules depend heavily on spatial structure, %\pk{ARE YOU SURE \texttt{colOrderBRKY0123} DEPENDS ON SPATIAL STRUCTURE? }\christo{THAT WAS A MISTAKE, CORRECTED IT TO BE READING ORDER RULES LIKE ordL1} 
suggesting that the FC representation strongly couples rule learning with absolute board coordinates observed during training. When objects appear in previously unseen locations, the learned feature activations no longer align reliably with the underlying rule.

Overall, the FC experiments indicate limited positional generalization. The representation tends to encode location-specific spatial templates rather than learning more abstract rule-consistent relationships.

% ============================================================
\subsubsection{Object-Centric (OC) Generalization}
\label{sec:oc_generalization}
% ============================================================

\begin{figure}[htb]
    \centering
    \includegraphics[width=\textwidth,
    alt={Comparison of Object-Centric in-distribution and out-of-distribution error rates across rules. The model is trained with nine objects and evaluated both with nine objects and with five objects. Error increases remain comparatively modest except for strongly spatial rules.}]
    {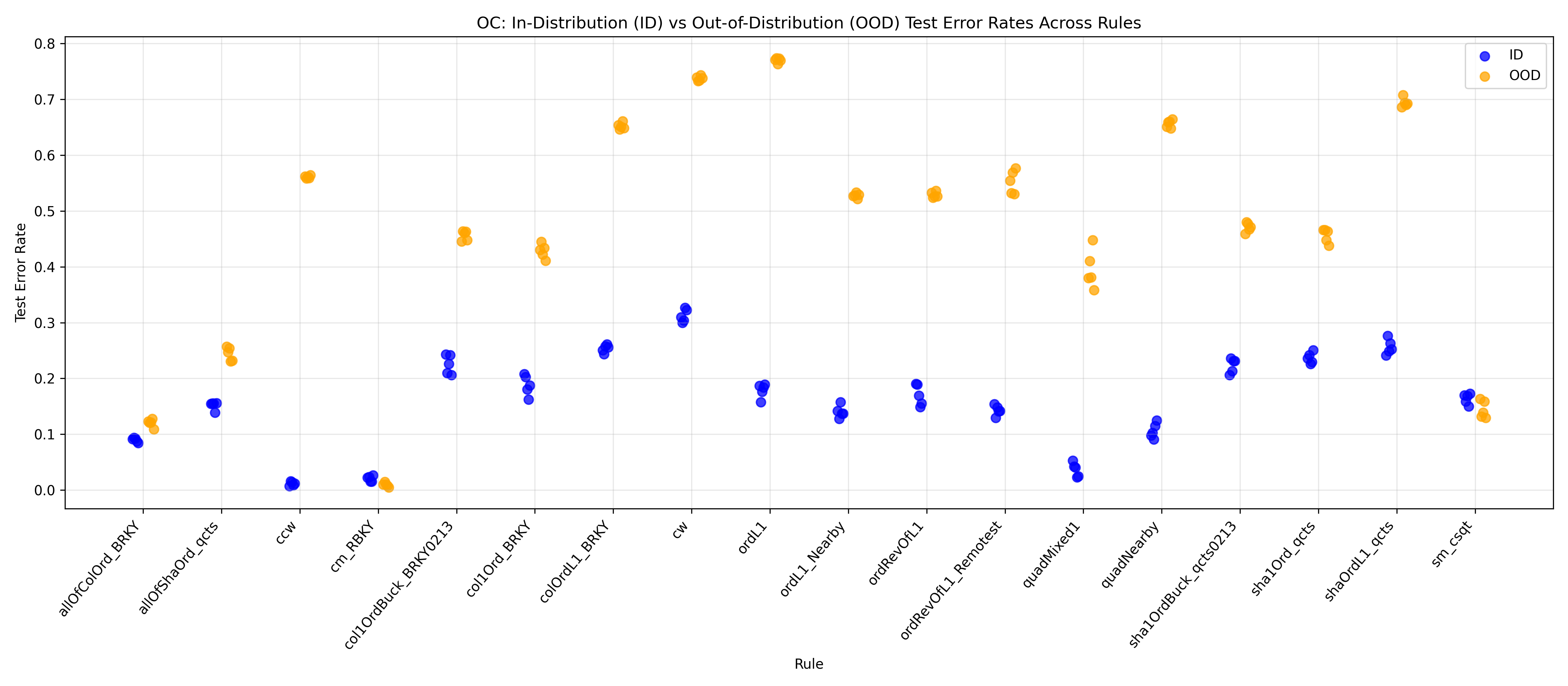}
    \caption{
    Object-Centric (OC) generalization performance.
    %\christo{
    In-distribution (ID) and out-of-distribution (OOD) error rates across rules. The OC model is trained using boards containing 9 objects and evaluated using both 9-object (ID) and 16-object (OOD) settings. The figure shows the effect of increasing scene cardinality beyond the range observed during training.
    }
    %}
    \label{fig:oc_outofdist}
\end{figure}

Unlike the FC representation, the OC model represents each object independently through its attributes and position encoding, reducing reliance on fixed board layouts. To evaluate its generalization capability, we introduced a distribution shift in the number of objects present during evaluation.

The OC model was trained exclusively using boards containing 9 objects and evaluated under two conditions:
\begin{itemize}
    \item \textbf{In-distribution (ID):} evaluation with 9 objects
    \item \textbf{Out-of-distribution (OOD):} evaluation with 
    % 5
    16 objects
\end{itemize}

% This setup tests whether the learned rule structure transfers across different scene cardinalities.
%\christo{
This experiment evaluates whether rules learned from smaller scenes transfer to larger and denser object configurations that were never encountered during training.

% Fig.~\ref{fig:oc_outofdist} shows the resulting ID and OOD error rates. In contrast to FC, the OC representation maintains comparatively stable performance under the distribution shift. ID error rates generally fall within the 10--25\% range for most rules. Under the OOD setting with 5 objects, errors increase only moderately, typically remaining between 15--30\%.

% The largest degradation again occurs for the spatially sensitive rules \texttt{quadNearby} and \texttt{quadMixed1}, where OOD errors approach 50--55\%. These rules rely strongly on precise spatial relationships, making them more sensitive to changes in scene structure. Nevertheless, even in these cases, the degradation remains substantially smaller than that observed for FC.

% These results suggest that the OC representation captures rule structure in a more transferable manner, allowing the learned policy to adapt more effectively to changes in object configuration and scene composition. \pk{THINKING ABOUT THIS, IT IS NOT CLEAR WHY HAVING FEWER OBJECTS ON THE BOARDS MAKES IT HARDER TO LEARN THE RULE. FOR OOD THERE ARE NOT AS MANY PIECES ON THE BOARD. WHY IS THE AI NOT MOVING THE ONES THAT IT SEES CORRECTLY? I THINK YOU HAVE SOME INSIGHT HERE THAT I DO NOT SEE..}

%\christo{
Fig.~\ref{fig:oc_outofdist} presents the ID and OOD error rates across rules. Increasing the number of objects from 9 to 16 produces a measurable degradation in performance for most rules. However, the extent of degradation varies substantially across rule types.

%\christo{
Rules based primarily on local feature mappings, such as color-to-bucket and shape-to-bucket mappings, exhibit comparatively modest increases in error. In contrast, ordering-based rules and rules requiring reasoning over larger sets of objects often show substantially larger OOD errors. These rules require the agent to process longer object sequences and maintain more complex relational structure than was encountered during training.

%\christo{
These results suggest that the OC representation is capable of transferring learned rule structure to scenes containing substantially more objects than those observed during training. Nevertheless, increasing scene cardinality introduces additional complexity, particularly for rules that depend on ordering, relational comparisons, or global spatial structure. Thus, while OC exhibits meaningful generalization beyond the training distribution, its performance remains sensitive to large increases in scene complexity.
% } 
% \pk{I AM NOT CONVINCED THAT THIS IS THE EXPLANATION, BUT I THINK IT IS OK TO LEAVE IT THIS WAY. }

% ============================================================
\subsubsection{Observations}
\label{sec:generalization_observation}
% ============================================================

% The generalization experiments reveal a clear contrast between the two representations. FC models degrade sharply when evaluated on previously unseen positions, indicating strong dependence on positional patterns encountered during training. In contrast, OC models remain considerably more stable under changes in the number of objects and overall scene composition. 

% Both representations struggle on highly spatial rules such as \texttt{quadNearby} and \texttt{quadMixed1}, highlighting the difficulty of learning robust spatial abstractions from raw coordinate encodings alone. However, the magnitude of degradation is consistently smaller for OC.

% Taken together, these results show that OC models experience smaller degradation than FC models under the evaluated distribution shifts. 

%\christo{
The FC and OC representations were evaluated under different forms of distribution shift chosen to challenge the aspects of the environment most closely associated with each representation.
%}

%\christo{
For the FC representation, generalization was evaluated using previously unseen board positions. Since FC encodes the board as a spatial feature map, positional changes provide a direct test of whether the learned policy captures rule structure independently of the specific locations observed during training. 
% The results show substantial degradation under this positional shift, indicating that FC often relies on spatial activation patterns associated with training locations.
%}
The results show substantial degradation under this positional shift, suggesting that FC does not fully abstract the underlying rule structure from board position. One possible explanation is that the learned policy relies in part on position-specific patterns observed during training rather than learning a more location-independent representation of the rule.
% \pk{OR MAYBE IT JUST WAS LEARNING RULES FOR EACH POSITION, RATHER THAN ABSTRACTING?}

%\christo{
For the OC representation, generalization was evaluated by increasing the number of objects from 9 to 16. Because OC represents objects individually rather than as fixed board locations, changes in scene cardinality provide a more natural test of its ability to transfer learned rule structure to larger and denser scenes. The results show that performance generally degrades as scene complexity increases, although the magnitude of this effect varies considerably across rule types.
%}

%\christo{
Several rules exhibit notable sensitivity under their respective distribution shifts. In FC, strong degradation is observed for reading-order and spatially structured rules such as \texttt{ordL1}, \texttt{ordL1\_Nearby}, \texttt{quadNearby}, and \texttt{quadMixed1}. In OC, larger performance drops are typically observed for ordering-based and relational rules, which require reasoning over larger sets of objects when scene cardinality increases.
%}

%\christo{
Together, these results characterize how each representation behaves when evaluated under distribution shifts that target its underlying inductive biases.
%}

%\pk{TO ME, THE CHANGE FROM SEEING 9 TO ONLY SEEING 5 SEEMS 'MUCH SMALLER' THAN THE CHANGE FROM BLACK SQUARES TO WHITE SQUARES. DID OC SHOW NO CHANGE WITH THE CHECKERBOARD EXPERIMENTS?}. 
% \christo{CONDUCTED EXPERIMENT BY TESTING WITH 16 OBJECTS AS THAT MIGHT BE BETTER EXPERIMENT FOR GENERALIZATION AND UPDATED THE CONTENT.}
% \pk{THANK YOU. I THINK THERE ARE MANY INTERESTING QUESTIONS HERE, BEARING ON WHETHER THE RL SYSTEM IS FORMING CONCEPTS. BUT WE HAVE RUN OUT OF MONEY. }
% \subsection{Generalization Analysis}
% \subsubsection{Feature-Centric Generalization}
% % Position-based train/test split, checkerboard setup, OOD positions.

% \subsubsection{Object-Centric Generalization}
% % Train with one number of objects, test with another.

% \subsubsection{Comparison Between FC and OC Generalization}
% % Discuss inductive bias differences.

\subsection{Similarity of Shape and Color Rules}

The independent-rule experiments revealed several pairs of rules that are structurally analogous. In particular, the color-based and shape-based matching rules differ only in whether the rule operates on object color or object shape. Since both attributes are represented using one-hot encodings of equal dimensionality, one would expect these rule pairs to exhibit similar learning behavior.

To evaluate whether the corresponding shape and color rules were statistically similar, we applied the non-parametric Kruskal--Wallis test.

\begin{tcolorbox}[colback=gray!10!white, colframe=black, title=Kruskal--Wallis Test]
\textbf{Null hypothesis ($H_0$):} The medians of the metric distributions are identical across groups.\\[0.5em]
\textbf{Alternative hypothesis ($H_1$):} At least one group differs in median from the others.
\end{tcolorbox}

The analysis was conducted using the three evaluation metrics introduced earlier:
\[
m^\star, \quad e^\star_{\mathrm{mean}}, \quad e^\star_{\mathrm{max}}.
\]

We performed two complementary analyses:
\begin{enumerate}
    \item consistency across repeated runs of the same rule;
    \item similarity between corresponding shape-based and color-based rules.
\end{enumerate}

% ============================================================
\subsubsection{Within-Rule Comparison}
% ============================================================

To evaluate the stability of training outcomes, we first examined whether repeated runs of the same rule produced statistically similar results. For each rule, the Kruskal--Wallis test was applied independently to the five runs associated with each evaluation metric.

For every rule, at least one of the three metrics failed to reject the null hypothesis, \footnote{
% \pk{
We checked an enormous number of hypotheses, and so expect to see many false flags of significance. Thus the conclusions here are informal, rather than statistical
% }
}indicating that the runs were not very different
% \pk{not very different }
%statistically indistinguishable 
along that evaluation dimension. Although variability remained across some metrics, the results suggest that the learning behavior for a given rule is reasonably stable across repeated training runs.

% ============================================================
\subsubsection{Comparison Between Shape and Color Rules}
% ============================================================

We next compared corresponding shape-based and color-based rules. These rule pairs differ only in the attribute used by the rule definition while preserving the same underlying structural pattern. Examples include:
\[
\texttt{allOfColOrd\_BRKY} \leftrightarrow \texttt{allOfShaOrd\_qcts}
\]
and
\[
\texttt{col1Ord\_BRKY} \leftrightarrow \texttt{sha1Ord\_qcts}.
\]

For each rule pair, the Kruskal--Wallis test was applied separately to all three evaluation metrics.

To reduce run-to-run variance and obtain more stable estimates, we aggregated seven independent sets of experiments, resulting in a total of 35 runs per rule.

Out of the 30 total comparisons
\[
(10 \text{ rule pairs}) \times (3 \text{ metrics}),
\]
29 produced $p$-values greater than 0.05, indicating no statistically significant difference between the corresponding shape and color rules.

The only exception occurred for the pair
\[
\texttt{(col1Ord\_BRKY,\ sha1Ord\_qcts)}
\]
under the FC model using the $m^\star$ metric, where the test yielded
\[
p = 0.036.
\]

However, the remaining two metrics for this pair produced non-significant results ($p > 0.05$). Overall, the analysis indicates that the corresponding shape-based and color-based rules behave similarly under the learned representations.

% ============================================================
\subsubsection{Effect of Increasing the Number of Runs}
% ============================================================

Because non-parametric statistical tests are sensitive to sample size, we additionally investigated the effect of increasing the number of runs on the stability of the conclusions.

For the within-rule analysis, we pooled three independent sets of experiments, increasing the number of runs per rule from 5 to 15.

Under the original 5-run setup, 40 out of 60 comparisons ($66\%$) were classified as statistically indistinguishable. After increasing the sample size to 15 runs, this proportion increased to 50 out of 60 comparisons ($83\%$).

These results suggest that increasing the number of runs reduces variance and improves the stability of the statistical conclusions. At the same time, the original 5-run analysis still captures the overall similarity trends, while the larger pooled analysis acts as an additional robustness check.

% ============================================================
\subsubsection{Summary}
% ============================================================

The statistical analysis supports three main conclusions:

\begin{enumerate}
    \item Repeated runs of the same rule generally produce consistent learning behavior, with each rule exhibiting at least one metric for which the runs are statistically indistinguishable.
    
    \item Corresponding shape-based and color-based rules exhibit highly similar performance characteristics. Across 30 pairwise comparisons, 29 showed no statistically significant difference.
    
    \item Increasing the number of runs improves the stability of the statistical conclusions, indicating that some of the variability observed under small sample sizes is attributable to stochastic variation across training runs.
\end{enumerate}

The detailed statistical results are available at:
\url{https://tinyurl.com/55e4bxbz}.
%\pk{ARE THEY UP TO DATE HERE AT THIS SITE}\christo{YES, THESE DATA WASN'T CHANGED AFTER FIRST TECH REPORT}
% \subsubsection{Within-Rule Comparison}
% \subsubsection{Between-Rule Comparison}
% \subsubsection{Statistical Testing}
% Mann-Whitney, Kruskal-Wallis, or other tests used.

\subsection{Transfer Experiments}

% ============================================================
\subsubsection{Simple-to-Simple Transfer}
\label{sec:simple_to_simple_transfer}
% ============================================================
% \pk{Pausing here 6/8/2026 6:36:41 PM}
To study transfer behavior between structurally simple rules, we conducted a set of simple-to-simple transfer experiments using both the Feature-Centric (FC) and Object-Centric (OC) representations.

The experiments were performed using four major groups of rules:

\begin{enumerate}
    \item \textbf{Bucket-ordering rules:}
    \texttt{cw}, \texttt{ccw}, \texttt{buckets\_0213}, \texttt{buckOrd02}, \texttt{buckOrd13}
    
    \item \textbf{Quadrant-based rules:}
    \texttt{quadNearby}, \texttt{quadMixed1}
    
    \item \textbf{Feature-based rules:}
    \texttt{sm\_csqt}, \texttt{sm\_qcts}, \texttt{cm\_RBKY}, \texttt{cm\_BRYK}
    
    \item \textbf{Reading-order rules:}
    \texttt{ordL1}, \texttt{ordRevOfL1}
\end{enumerate}

% \pk{Note that each group of rules seems to contain only a single ``abstract concept.''}  
% For each predecessor rule, the model was first trained independently for five runs. 
%\pk{I THINK THIS MEANS THAT IT WAS TRAINED 5 TIMES STARTING FROM SOME RANDOM POSITION, AND THE RESULTING WEIGHTS WERE SAVED. IS THAT CORRECT?} \christo{YES}
% The resulting checkpoints were saved separately and later reused as initialization for downstream target rules. Each transferred checkpoint was then trained on the target rule for three additional runs. 
%\pk{CAN WE SAY THAT EACH PAIR WAS TESTED 15 TIMES,BUT THEY ARE NOT STATISTICALLY INDEPENDENT. }\christo{REWRITING THIS INTO NEXT THREE PARGRAPH FOR CLARITY}
% \pk{CHRISTO: THANK YOU FOR THE CLARIFICATION}

%\christo{
Note that each group of rules seems to contain only a single ``abstract concept.''
For each predecessor rule, five independent training runs were performed starting from different random initializations. The resulting checkpoints were saved and subsequently used as initialization points for transfer learning experiments on downstream target rules.
%}

%\christo{
For each predecessor-target pair, every saved predecessor checkpoint was used to initialize three independent training runs on the target rule. Consequently, each transfer pair was evaluated using a total of fifteen transfer runs (five predecessor checkpoints multiplied by three target-rule training runs per checkpoint).
%}

%\christo{
Although fifteen transfer runs were obtained for each pair, the runs are not fully statistically independent because groups of three runs share the same predecessor checkpoint. The results should therefore be interpreted as repeated transfer evaluations from a common set of pretrained initializations rather than as fifteen completely independent training runs. 
% \pk{GREAT! THANK YOU}
%}

To compare transfer effects across FC and OC fairly, transfer performance was normalized relative to the baseline convergence cost of the target rule. Specifically, for predecessor rule $r_p$ and target rule $r_t$, we define the normalized transfer ratio:
\[
T(r_p \rightarrow r_t)
=
\frac{
M^\star_{\mathrm{transfer}}
}{
M^\star_{\mathrm{baseline}}
},
\]
where $M^\star_{\mathrm{baseline}}$ denotes the median convergence value when the target rule is trained from scratch.

Under this formulation:
\begin{itemize}
    \item $T < 1$ indicates positive transfer,
    \item $T > 1$ indicates negative transfer,
    \item $T \approx 1$ indicates little or no transfer effect.
\end{itemize}

\begin{figure}[htbp]
    \centering
    \includegraphics[width=\textwidth]
    {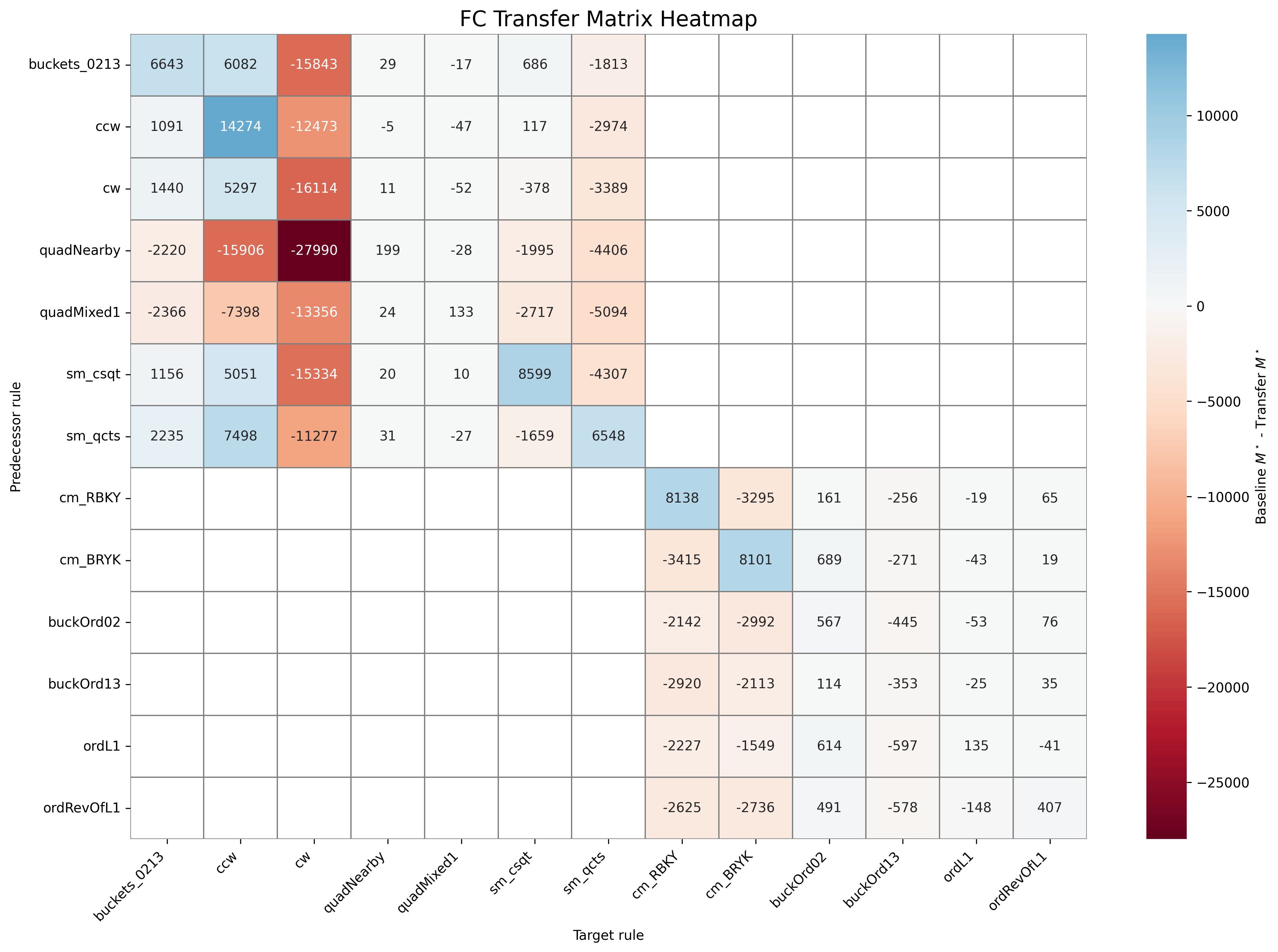}
    \caption{
    Normalized simple-to-simple transfer matrix for the FC representation. Rows denote predecessor rules and columns denote target rules. Values below 1 indicate positive transfer, while values above 1 indicate negative transfer relative to baseline training. % \pk{DO WE REALLY HAVE 2 DENSE BLOCKS? IF SO, WE CAN DO ANOTHER KIND OF METRIC HERE, LIKE THE ANALYSIS FOR THE GEMINI DATA, AND THAT WOULD BE GREAT. }
    } %\christo{IT APPEARS AS 2 DENSE BLOCKS BECAUSE EXPERIMENTS WERE ONLY CONDUCTED FOR THOSE COMBINATIONS. THE CELLS WITHOUT ANY VALUES INDICATES NO EXPERIMENTS WERE CONDUCTED FOR THAT PREDECESSOR-TARGET PAIR. }
    \label{fig:fc_simple_transfer_heatmap}
\end{figure}

\begin{figure}[htb]
    \centering
    \includegraphics[width=\textwidth]
    {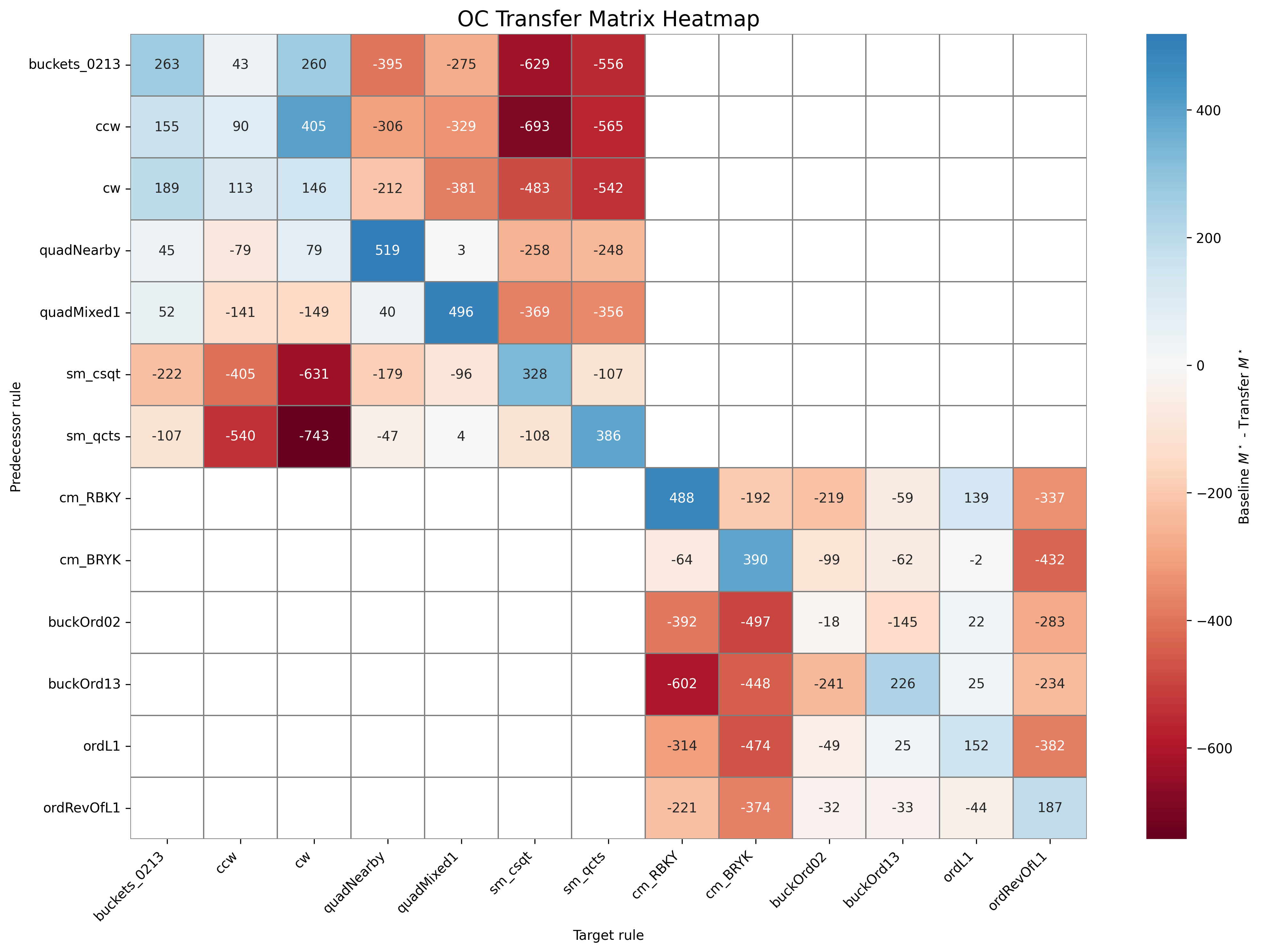}
    \caption{
    Normalized simple-to-simple transfer matrix for the OC representation. Positive transfer is concentrated primarily within structurally related rule families, while negative transfer appears systematically between incompatible relational structures.
    }
    \label{fig:oc_simple_transfer_heatmap}
\end{figure}

The resulting transfer matrices for FC and OC are shown in Figs.~\ref{fig:fc_simple_transfer_heatmap} and~\ref{fig:oc_simple_transfer_heatmap}.

% ============================================================
\paragraph{Feature-Centric (FC) Transfer Behavior}
% ============================================================

The FC representation exhibits highly variable and weakly structured transfer behavior across simple rules. Although some positive transfer occurs within related rule families, the overall transfer patterns remain inconsistent and highly sensitive to the specific predecessor-target combination.

Among bucket-ordering rules, moderate positive transfer is observed between several related pairs. In particular, \texttt{buckets\_0213}, \texttt{ccw}, and \texttt{buckOrd02} often improve one another. Similarly, the feature-based rules \texttt{sm\_csqt} and \texttt{sm\_qcts} exhibit strong mutual transfer, indicating that FC can partially reuse low-level feature activations associated with object categories.

However, transfer behavior within the FC representation remains notably unstable. The rule \texttt{cw} exhibits especially irregular behavior: depending on the predecessor checkpoint used for initialization, transfer may either improve or severely degrade convergence. In several cases, even self-transfer from pretrained \texttt{cw} checkpoints produces worse convergence than training from scratch. 

This instability suggests that FC does not consistently learn a stable reusable abstraction of the underlying rule. Instead, different runs appear to converge toward distinct positional heuristics or spatial activation patterns, some of which transfer constructively while others interfere strongly with later optimization.

Negative transfer is particularly strong between structurally incompatible rule families. For example, quadrant-based rules such as \texttt{quadNearby} and \texttt{quadMixed1} produce severe degradation when transferred to ordering-based targets such as \texttt{cw}. Likewise, ordering-based curricula often negatively affect feature-based targets.

At the same time, several predecessor-target pairs with little apparent semantic overlap still produce weak positive transfer. These effects likely arise from incidental optimization similarities or shared positional statistics rather than from true relational reuse.

Overall, the FC representation exhibits noisy and weakly organized transfer structure. Positive transfer is possible when predecessor and target share coarse structural similarities, but the learned representations remain highly checkpoint-dependent and sensitive to stochastic optimization trajectories. Oddly, the negative effects, between different rules, provide stronger support for the notion that some abstraction has been learned, while the transfer between similar rules does not provide much support.
 % \pk{Oddly, the negative effects, between different rules, provide stronger support for the notion that some abstraction has been learned, while the transfer between similar rules does not provide much support.}

% ============================================================
\paragraph{Object-Centric (OC) Transfer Behavior}
% ============================================================

In contrast to FC, the OC representation exhibits substantially more coherent and semantically structured transfer behavior. Positive transfer is concentrated primarily within related rule families, while negative transfer emerges systematically between incompatible relational structures.

The clearest transfer cluster appears among the bucket-ordering rules:
\[
\texttt{cw},
\texttt{ccw},
\texttt{buckets\_0213},
\texttt{buckOrd02},
\texttt{buckOrd13}.
\]
These rules consistently transfer positively to one another, indicating that OC successfully learns reusable relational abstractions associated with bucket-order structure.

Similarly, the feature-based rules
\[
\texttt{sm\_csqt},
\texttt{sm\_qcts},
\texttt{cm\_RBKY},
\texttt{cm\_BRYK}
\]
also form a coherent transfer group, with strong positive transfer occurring within the family. This suggests that OC representations effectively reuse object-level feature relationships across related tasks.

The quadrant-based rules \texttt{quadNearby} and \texttt{quadMixed1} likewise exhibit moderate positive mutual transfer, although the effect is weaker than in the ordering and feature families.

Unlike FC, the negative transfer observed in OC is highly systematic rather than noisy. Ordering-based predecessors frequently interfere with feature-based targets, while feature-based curricula negatively affect ordering-based tasks. This behavior suggests that OC learns relatively strong relational abstractions which can either facilitate or conflict with downstream rule structure depending on structural alignment.

Reading-order rules (\texttt{ordL1}, \texttt{ordRevOfL1}) occupy an intermediate position within the transfer matrix. They share partial compatibility with ordering-based tasks due to their sequential structure, but also exhibit substantial negative transfer when combined with unrelated feature-based rules.

Another important difference from FC is the overall stability of the OC transfer patterns. The OC representation produces substantially more consistent transfer behavior across checkpoints, indicating that the learned relational representations are more stable and reproducible across independent training runs. 
% \pk{I THINK THIS MEANS THAT THE EFFECTS ARE VERY SIMILAR FOR EACH OF THE 5 PREPARATIONS. ARE THE DATA IN AN APPENDIX OR SOMEWWHERE ON THE INTERNET. IF SO, IT MAKES SENSE TO ADD A REFERENCE. MAYBE I WILL COME TO YOUR DISCUSSION LATER ONE.}
% #ref

Overall, the simple-to-simple transfer experiments reveal a strong contrast between the two representations. FC transfer behavior is highly unstable and dominated by checkpoint-specific positional heuristics, whereas OC produces coherent family-level transfer structure consistent with reusable object-centric relational abstractions.

% \subsubsection{Transfer to Compound Rules}
% Compound target rules and component-based curricula.
% ============================================================
\subsubsection{Compound Transfer Experiments}
\label{sec:compound_transfer}
% ============================================================

To study 
% compositional
%\christo{
compound
%}
% \pk{WHY USE BOTH WORDS" COMPOUND, COMPOSITIONAL?} 
transfer behavior, we conducted a set of transfer-learning experiments involving compound rules constructed from simpler component rules. The experiments were performed using the trial-list-based configuration of the GOHR environment.

Each trial-list defines a sequence of training phases, where every row specifies the rule encountered during a particular stage of training. The experiments are designed to analyze whether prior exposure to the component rules of a compound target facilitates faster learning, and how transfer changes when structurally unrelated rules are introduced into the curriculum.

Consider four independent rules $A$, $B$, $C$, and $D$, and a compound rule $A+B$ formed by combining the relational constraints of rules $A$ and $B$. Two types of transfer-learning protocols were studied:

\begin{enumerate}
    \item \textbf{Sequential Transfer.}
    In this setting, the agent is first trained on the two component rules separately, followed by training on their compound rule:
    \[
    A \rightarrow B \rightarrow (A+B).
    \]

    \item \textbf{Partial / Mixed Transfer.}
    In this setting, the agent is exposed to one relevant component together with one unrelated rule before training on the compound target:
    \[
    A \rightarrow C \rightarrow (A+B).
    \]
    \item \textbf{Unrelated Transfer.}
    In this setting, the agent is exposed to two unrelated rule before training on the compound target:
    \[
    A \rightarrow B \rightarrow (C+D).
    \] 
    % \pk{THANKS!}
\end{enumerate}
%\pk{DID WE ALSO DO THE A,B -> C+D. I KNOW WE DID THEM FOR HUMANS. }\christo{WE HAVE DONE SOME EXPERIMENTS ON THAT ALSO. I HAVE ADDED IT. }

For all transfer phases, $\epsilon$ was reinitialized to $\epsilon_{\text{start}}$ at the beginning of each stage to encourage exploration and avoid premature convergence to policies inherited from earlier rules.

Each experiment was repeated across multiple random seeds. For every curriculum configuration, the convergence metric $M^\star$ was recorded. To compare transfer performance across different targets and representations, transfer performance was normalized by the baseline convergence of the target rule trained from scratch:
\[
\frac{M^\star_{\text{transfer}}}
     {M^\star_{\text{baseline}}}.
\]

Values below $1$ indicate positive transfer (faster convergence than baseline), whereas values above $1$ indicate negative transfer.

% ============================================================
\subsubsection{Transfer to \texttt{ordL1\_Nearby}}
% ============================================================

\begin{figure}[htbp]
    \centering
    \includegraphics[width=0.92\linewidth]{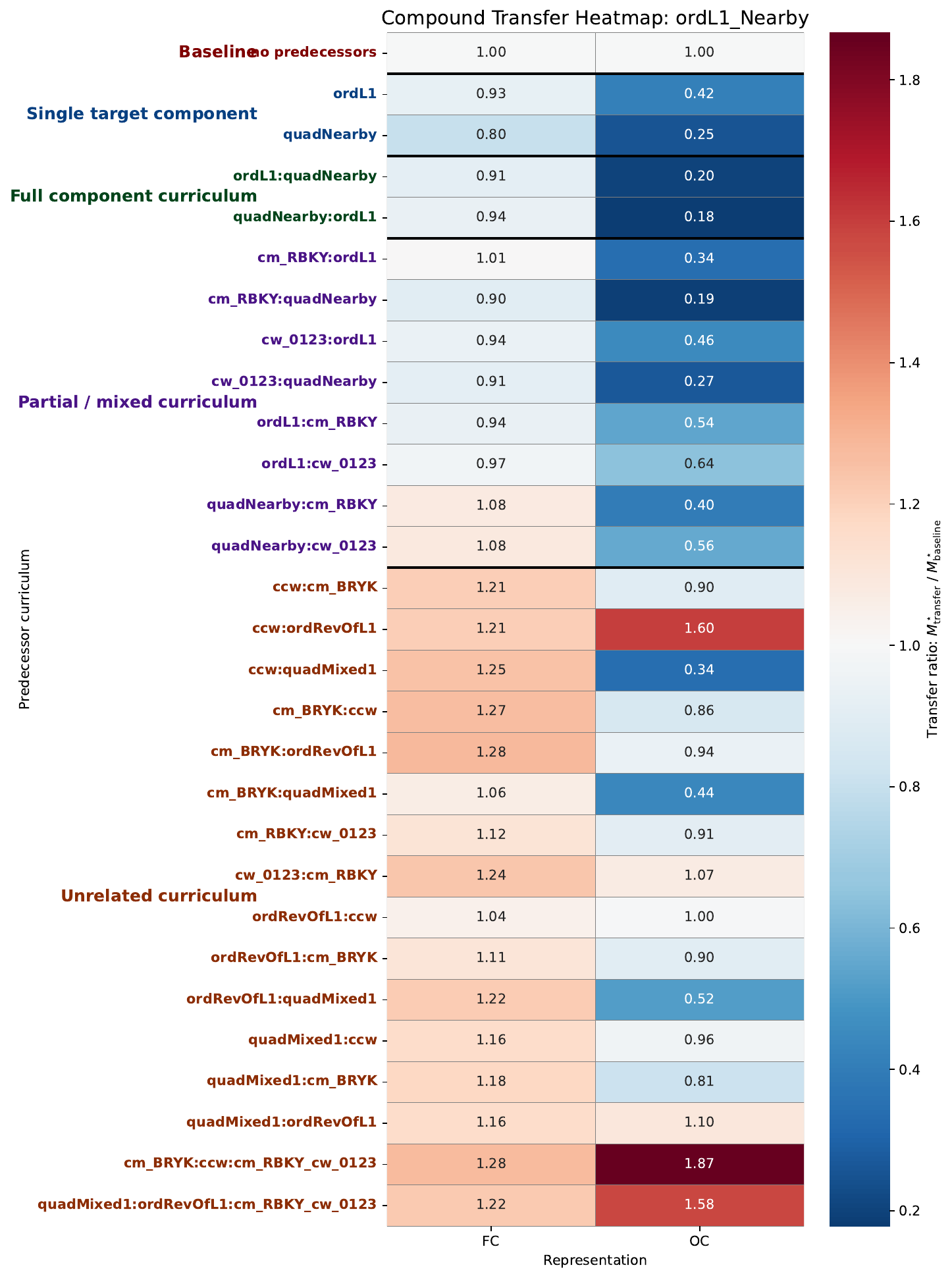}
    \caption{
    Transfer heatmap for the compound rule \texttt{ordL1\_Nearby}. %\christo{
    The left column corresponds to the Feature-Centric (FC) representation and the right column to the Object-Centric (OC) representation.  %}
    Each cell reports the normalized transfer ratio
    $
    M^\star_{\mathrm{transfer}} / M^\star_{\mathrm{baseline}}
    $.
    Values below $1$ indicate positive transfer, while values above $1$ indicate negative transfer. %\pk{IS THIS ONLY FOR OC. IF SO WE SHOULD SAY SO.}
    }
    \label{fig:compound_ordL1_heatmap}
\end{figure}

Fig.~\ref{fig:compound_ordL1_heatmap} shows
% presents
the normalized transfer behavior %\christo{
of both the FC and OC representations for the compound rule
\[
\texttt{ordL1\_Nearby}.
\]
% which
This target rule combines two underlying concepts: reading-order structure (\texttt{ordL1}) and
% with
proximity-based spatial 
% constraints
reasoning (\texttt{quadNearby}). The predecessor curricula span reading-order, spatial, bucket-ordering, and feature-mapping rule families, allowing us to examine how transfer depends on the structural relationship between the curriculum and the target rule. 
% The pretraining curricula include reading-order rules (\texttt{ordL1}, \texttt{ordRevOfL1}), spatial rules (\texttt{quadNearby}, \texttt{quadMixed1}), bucket-ordering rules (\texttt{cw\_0123}, \texttt{ccw}), and feature-mapping rules (\texttt{cm\_RBKY}, \texttt{cm\_BRYK}).

The OC representation exhibits extremely strong positive transfer when pretrained on the true component rules. In particular, the curricula
\[
\texttt{ordL1} \rightarrow \texttt{quadNearby}
\]
and
\[
\texttt{quadNearby} \rightarrow \texttt{ordL1}
\]
reduce the normalized convergence ratio to approximately $0.2$, corresponding to an almost five-fold improvement relative to baseline training.

In contrast, the FC representation
% \pk{, not shown here,} 
%\christo{FC AND OC ARE THE LEFT AND RIGHT COLUMNS IN THE FIGURE}
exhibits only modest gains from the same curricula. Most FC transfer ratios remain close to $1$, indicating that FC reuses comparatively weaker relational structure from predecessor rules.

Partial and mixed curricula again reveal strong variation
% \pk{ variation }
%asymmetry 
in OC. Some curricula containing one relevant component continue to provide moderate transfer benefits, while others introduce substantial degradation depending on the ordering and structural compatibility of the unrelated rule.

% An interesting observation is that 
%\christo{
As shown in Fig.~\ref{fig:compound_ordL1_heatmap}, several unrelated curricula produce near-baseline behavior in FC %\pk{WE HAVE NOT SHOWN THE HEAT MAP FOR THESE. IS IT COMING LATER SOMEWHERE? IF SO, CROSS REFERENCE TO IT WITH SLASH-REF}\christo{Added the reference again.} 
% but 
while 
still yielding strong positive transfer in OC. 
% Certain
%\christo{
In particular, 
curricula involving \texttt{ccw}, \texttt{quadMixed1}, \texttt{cm\_BRYK}, and \texttt{ordRevOfL1} 
% continue to 
improve convergence in OC despite not being direct components of the target rule. This suggests that OC can sometimes reuse partially compatible spatial or ordering abstractions even across structurally distinct rules.

At the same time, the OC representation remains highly sensitive to incompatible curricula. Certain unrelated curricula still produce negative transfer exceeding the FC degradation. Thus, OC demonstrates stronger compositional reuse but also greater sensitivity to the structure and ordering of predecessor rules.

% ============================================================
\subsubsection{Transfer to \texttt{cm\_RBKY\_cw\_0123}}
% ============================================================

The second compound rule considered is
\[
\texttt{cm\_RBKY\_cw\_0123},
\]
which combines feature-to-bucket mapping with bucket-ordering structure. The pretraining curricula include feature-mapping rules (\texttt{cm\_RBKY}, \texttt{cm\_BRYK}), bucket-ordering rules (\texttt{cw\_0123}, \texttt{ccw}), reading-order rules (\texttt{ordL1}, \texttt{ordRevOfL1}), and spatial rules (\texttt{quadNearby}, \texttt{quadMixed1}). The complete transfer heatmap for this compound rule is provided in Appendix~\ref{appendix:transfer_plots}.

Both FC and OC representations exhibit strong positive transfer when the curriculum contains the true components of the target rule. In particular, the curricula
\[
\texttt{cm\_RBKY} \rightarrow \texttt{cw\_0123}
\]
and
\[
\texttt{cw\_0123} \rightarrow \texttt{cm\_RBKY}
\]
produce the strongest improvements. The effect is especially pronounced in the OC representation, where the normalized convergence ratio decreases to approximately $0.5$, indicating that the target rule can be learned in nearly half the number of episodes required from scratch.

Single-component pretraining also improves convergence, although the effect is weaker than full-component pretraining. Interestingly, pretraining only on \texttt{cm\_RBKY} consistently provides stronger transfer than pretraining only on \texttt{cw\_0123}. In FC, pretraining on \texttt{cm\_RBKY} reduces the normalized convergence ratio to approximately $0.13$, whereas pretraining only on \texttt{cw\_0123} yields a much weaker improvement. In OC, the asymmetry is even stronger: \texttt{cm\_RBKY} alone still provides substantial positive transfer, while \texttt{cw\_0123} alone produces clear negative transfer with ratios exceeding $1.7$.

Partial and mixed curricula reveal strong order sensitivity, particularly in the OC representation. Curricula containing one correct component together with an unrelated rule often produce %\christo{
widely varying
% unstable 
%\pk{MAY IT IS BETTER TO USE 'widely varying' RATHER THAN 'unstable'?} 
outcomes. Some combinations still yield moderate improvements, whereas others produce severe negative transfer. For example,
\[
\texttt{ordL1} \rightarrow \texttt{cw\_0123}
\]
and
\[
\texttt{quadNearby} \rightarrow \texttt{cw\_0123}
\]
lead to large degradation in OC, with normalized transfer ratios exceeding $2$ or $3$.

The strongest degradation occurs for unrelated curricula, where the model is pretrained exclusively on structurally incompatible rules. In these settings, both FC and OC require substantially more episodes to converge than the baseline model trained from scratch. However, the degradation is consistently larger for OC, where several curricula produce transfer ratios above $3$. This 
% indicates 
%\christo{
supports our tentative conclusion
%\pk{MAYBE SAY 'suggests' OR 'supports our tentative conclusion' } 
that OC forms strong relational abstractions during pretraining that can either strongly facilitate or strongly interfere with later learning depending on structural alignment.

%Interestingly,

To sum up, FC exhibits comparatively conservative transfer behavior. Although unrelated curricula degrade performance, the transfer ratios generally remain within the $1.2$--$1.8$ range. OC, by contrast, exhibits both stronger positive transfer and substantially stronger negative transfer, demonstrating significantly higher transfer sensitivity, 
and suggesting that in some sense the OC representation comes closer to capturing what we think of as conceptual features.

% ============================================================
\subsubsection{Summary}
% ============================================================

Across both compound-rule experiments, several consistent transfer behaviors emerge.

\begin{enumerate}
    \item \textbf{Full-component pretraining produces the strongest positive transfer.}
    Compound rules converge substantially faster when the curriculum contains the actual components of the target rule.

    \item \textbf{Transfer is strongly order-dependent.}
    Even when the same component rules are used, reversing their order can significantly change transfer behavior, especially for the OC representation.

    \item \textbf{Single-component pretraining can still provide meaningful transfer.}
    However, the magnitude of improvement depends strongly on the structural role of the component within the target rule.

    \item \textbf{Unrelated pretraining generally produces negative transfer.}
    Curricula consisting solely of structurally incompatible rules tend to slow convergence relative to baseline training.

    \item \textbf{OC exhibits stronger compositionality and stronger transfer sensitivity.}
    When structural alignment exists, OC achieves dramatic reductions in convergence time. However, OC is also substantially more vulnerable to structurally incompatible curricula.

    \item \textbf{FC exhibits weaker but more stable transfer behavior.}
    % FC generally produces smaller improvements and smaller degradations, suggesting that its representation reuses relational structure more conservatively.
    % \christo{
    The 
    FC 
    representation
    generally produces smaller improvements and smaller degradations, suggesting that its learned representations encodes relational structure more weakly than OC. As a result, transfer effects tend to be weaker in both the positive and negative directions.
    %\pk{THIS IS A KIND OF ODD CONCLUSION, SINCE IT SUGGEST THAT TEHE FC IS A CONSCIOUS AGENT, MAKING SOME DECISION ABOUT HOW TO BEHAVE. I THINK IT MIGHT BE BETTER TO SAY THAT IT 'represents concepts more weakly'}
\end{enumerate}

Overall, the compound-transfer experiments demonstrate that transfer learning in GOHR depends primarily on structural alignment between predecessor and target rules. Object-centric representations enable stronger compositional reuse of relational abstractions, but this same property also makes them substantially more sensitive to curriculum mismatch and ordering effects.
% \paragraph{Sequential Transfer}
% % Sequentially training on rule components before target.

% \paragraph{Partial Transfer}
% % Training on only some components of the target rule.

% \subsubsection{Positive and Negative Transfer}
% % Discuss aligned vs misaligned curricula.

% \subsubsection{FC vs OC Transfer Behavior}
% % Contrast conservative FC transfer and stronger OC transfer sensitivity.

\subsection{Multidimensional Scaling Analysis}
The data on difficulty provide
% \pk{\sout{s}} 
one dimension for understanding relations between rules. On the difficulty scale, rules that are ``near each other'' have something in common, which we have speculated about in the discussion above. But there are other ways of organizing the rules, based on the data we have collected. 

\begin{figure}[t]
\centering
  \includegraphics[width=0.96\linewidth]{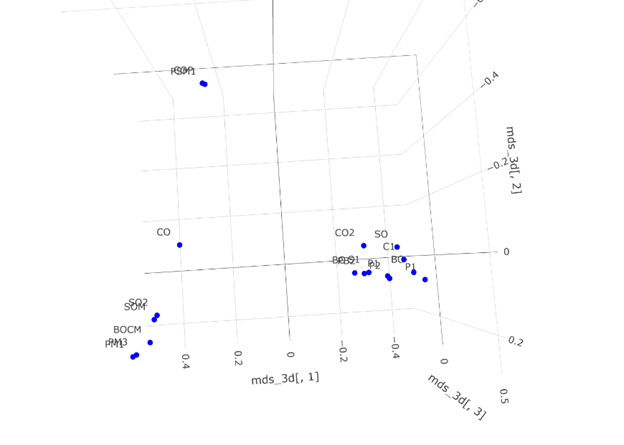} \hfill
  \caption {View of the first two principal dimensions of the MDS embedding of $p$-values as dissimilarity. Similarity is defined by being approximately as difficult for the AI system to learn. We might call this ``first order similarity.'' Note the heavy cluster containing many different rules.}
  \label{fig:topview_of_MDS}
 \end{figure}

%\ PK: I TOOK THIS OUT FOR PARAGRAPH SPACING. 

\begin{figure}[ht!]
\centering
  \includegraphics[width=0.96\linewidth]{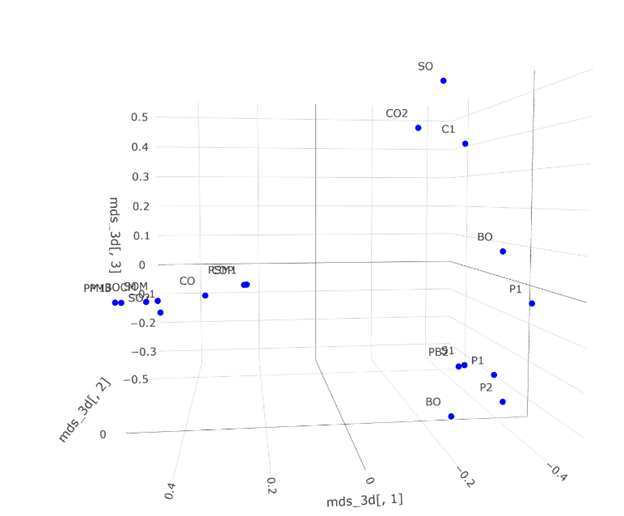} \hfill
  \caption {Perspective View of the first three  principal dimensions of the MDS embedding of $p$-values as 
  dissimilarity: ''first-order similarity''. The figure has been rotated to resolve two clusters that each contain many pointS in the two dimensional view with the first two components. }
  \label{fig:frontview_of_MDS}
 \end{figure}

For example, each data point, characterized by a $(rule,algorithm)$ pair actually represents the median of five distinct observations, as the algorithm worked on the rule from a cold start five times, That set of five values contains information beyond its median. For example, the range is approximately a 94\% confidence interval for the median of the (unknown) distribution of that specific metric for all possible encounters of that rule with that algorithm 

A more rigorous way to explore the relation is to apply the Mann-Whitney test for equality of medians. When applied to two samples, it gives us the exact probability that the two observed sets of numbers could have come from the same distribution. The exact probability is calculated by computing the value of the specific Mann-Whitney U statistic, and asking how many of the values could be more extreme than the one that is observed.  We have calculated these {$p$-values} {for the similarity between every possible pair of rule sets among the 18 total rule sets (see Table~\ref{tab:rule_based_exp}). The resulting $p$-values were recorded in a matrix.}

Specifically, we recognize that the $p$-value is a measure of similarity. Values close to $1$ are an indication that the sets of values could indeed come from the same distribution, while values close to zero are very unlikely, if both situations have the same distribution of difficulties.  For an initial look at the structure of the data, we compute a dissimilarity matrix $D_{i,j}=1-p_{MannWhitney}(i,j)$. We then used the R statistIcal environment to transform this into a lower diagonal distance array. We then used the basic R tool for multidimensional scaling, to  look for  a $3-dimenional$ embedding of the data into a Euclidean space. The results are a rotatable plot, The first two principal dimensions of the analysis are shown in Fig.~\ref{fig:topview_of_MDS}. It is easy to see that there are a few clusters, but each of them  seems to contain ``too many'' different rules. 

We continued the analysis, looking at a view using the third dimension as the vertical axis. In that view, shown in Fig.~\ref{fig:frontview_of_MDS} it is clear that the large cluster is resolved by differing values along the (now vertical) third axis.  %\pk{CHRISTO - CAN YOU FIGURE OUT HOW TO PLACE THE WHOLE ROTATABLE DIAGRAM ONTO ACTION IN THE SAME DIRECTORY THAT THIS ONE, FROM THE ROW AND COLUMN ANALYSIS, SITS. 
%\url{http://action.rutgers.edu/REPORTS/clean_FC_tmp_rowAndColumnAnalysis.html}
%THE DIRECTORY IS action:/opt/apache-tomcat-11.0.18/webapps/REPORTS IF YOU CANNOT WRITE IN IT, PLACE THE FILE IN YOUR OWN HOME DIRECTORY, WORLD READABLE, AND SEND ME A NOTE. .}\christo{I HAVE ADDED THE ROTATABLE DIAGRAM IN THE DIRECTORY(action:/opt/apache-tomcat-11.0.18/webapps/REPORTS ) , THOUGH I HAVEN'T UPDATED THE INDEX.HTML TO PROVIDE THE LINK TO THE FILE. } 
% \pk{CHRISTO - I HAVE UPDATED THE INDEX PAGE. SO I GUESS WE SHOULD JUST ADD A POINTER TO IT. RIGHT NOW IT IS NOT PERFECT BECAUSE (A) I JUST DUMPED THE WHOLE OUTPUT FILE FOR THE FC CASE AND (B) WHILE I SHOWED THE COLOR PLOT (AT THE VERY END), YOU ONLY SHOWED THE BLACK AND WHITE PLOT. HOW HARD WOULD IT BE TO GET THEM TO BE bOTH THE SAME FORMAT? }
Future work will address the question of whether all of these clusters make sense both in terms of the specifics of the algorithm and in terms of the underlying concepts within the rules.

% ============================================================
\subsection{Transfer Geometry and Clustering Structure}
\label{sec:transfer_geometry}
% ============================================================

To further analyze the structure of transfer learning relationships, we constructed a transfer-difference matrix from the simple-to-simple transfer experiments. This analysis builds upon the predecessor-conditioned transfer matrix presented earlier. Whereas the previous matrix summarized absolute convergence performance after transfer, the present matrix measures the change in convergence relative to baseline learning, thereby isolating the effect of transfer itself.
% Each matrix entry was defined as

% \[
% \Delta(i,j)
% =
% M^\star(r_i \mid \text{baseline})
% -
% M^\star(r_i \mid r_j),
% \]

% where \(r_i\) is the target (successor) rule and \(r_j\) is the predecessor rule used for pretraining. Thus, each row characterizes how a target rule responds to different predecessor curricula, while each column characterizes how a predecessor rule influences learning across multiple successor rules.

% Negative values indicate positive transfer (reduced sample complexity after pretraining), while positive values indicate interference or negative transfer.

% \christo{Corrected the equation and corresponding explanation}

\[
\Delta(i,j)
=
M^\star_{\mathrm{base}}(r_i)
-
M^\star_{\mathrm{transfer}}(r_i \leftarrow r_j),
\]

where $r_i$ is the target (successor) rule and $r_j$ is the predecessor rule used for pretraining. $M^\star_{base}(r_i)$ is the median convergence metric obtained when learning $r_i$ without a predecessor rule, while $M^\star_{transfer}(r_i \leftarrow r_j)$ is the median convergence metric obtained when learning $r_i$ after training on $r_j$. 

Thus, each row characterizes how a target rule responds to different predecessor curricula, while each column characterizes how a predecessor rule influences learning across multiple successor rules. 
%\pk{GOOD! CAN WE SAY THAT THIS IS A MORE REFINED EXTENSION OF THE MATRIX SHOWN BEFORE, FOR M* CONDITIONED ON PREDECESSOR?}\christo{MODIFIED THE FIRST PARAGRAPH IN SUBSECTION}

Positive values of $\Delta(i,j)$ indicate positive transfer, meaning that pretraining on $r_j$ reduced the median convergence time for learning $r_i$. Negative values indicate interference or negative transfer, meaning that pretraining on $r_j$ increased the median convergence time.

Because not every predecessor-successor pair was experimentally evaluated, the resulting matrix contained missing entries. 
%\christo{
For the clustering and multidimensional-scaling analyses, each rule was represented by its pattern of transfer relationships with the other rules. Missing entries were therefore replaced using mean-based imputation so that these transfer patterns could be compared consistently while preserving either predecessor-rule or successor-rule statistics.
%}
%\pk{IF I UNDERSTAND CORRECTLY, THE FULL DATA SET HAS TWO DENSE BLOCKS IN. WOULD IT BE POSSIBLE TO DO THE ANALYSIS FOR EACH BLOCK SEPARATELY. OF COURSE THEY WOULD HAVE NO OVERLAP AT ALL, BUT WE COULD SEE WHETHER THEY FALL INTO THE SAME GROUPS AS THEY DO WHEN WE USE IMPUTATION.} 
Two complementary imputations were therefore used:
%\pk{SOMEHWERE IN HERE WE SHOULD SAY THAT EACH RULE IS CONVERTED TO A VECTOR OF ITS TRANSFER FROM ALL OF THE OTHER VECTORS FOR WHICH DATA IS AVAILABLE. AND SOME IMPORTATON WAS DONE (I THINK IT IS THE MEAN OF THE VALUES FOR THAT CASE)}\christo{ADDED THE PARAGRAPH ABOVE}
\begin{enumerate}
    \item \textbf{Column-mean imputation} preserved predecessor-rule statistics by replacing missing values in each column with the column mean. The resulting matrix was used for row clustering.
    
    \item \textbf{Row-mean imputation} preserved successor-rule statistics by replacing missing values in each row with the row mean. The transpose of this matrix was used for column clustering.
\end{enumerate}

%\christo{
The resulting analyses provide a second-order notion of similarity among rules. Earlier analyses compared rules directly in terms of learning difficulty or convergence performance. Here, rules are compared through their patterns of transfer interactions with other rules. Two rules are therefore considered similar when they exhibit similar transfer relationships, even if their standalone learning difficulty differs substantially.

Hierarchical clustering and multidimensional scaling (MDS) were then applied separately to the FC and OC representations.

\subsubsection{Row clustering: similarity of successor-rule behavior.}

Row clustering groups rules according to how they respond to predecessor experience. In this view, two rules cluster together when they are similarly helped or hindered by the same predecessor rules.
\begin{figure}[htbp]
    \centering
    \includegraphics[width=0.8\linewidth]{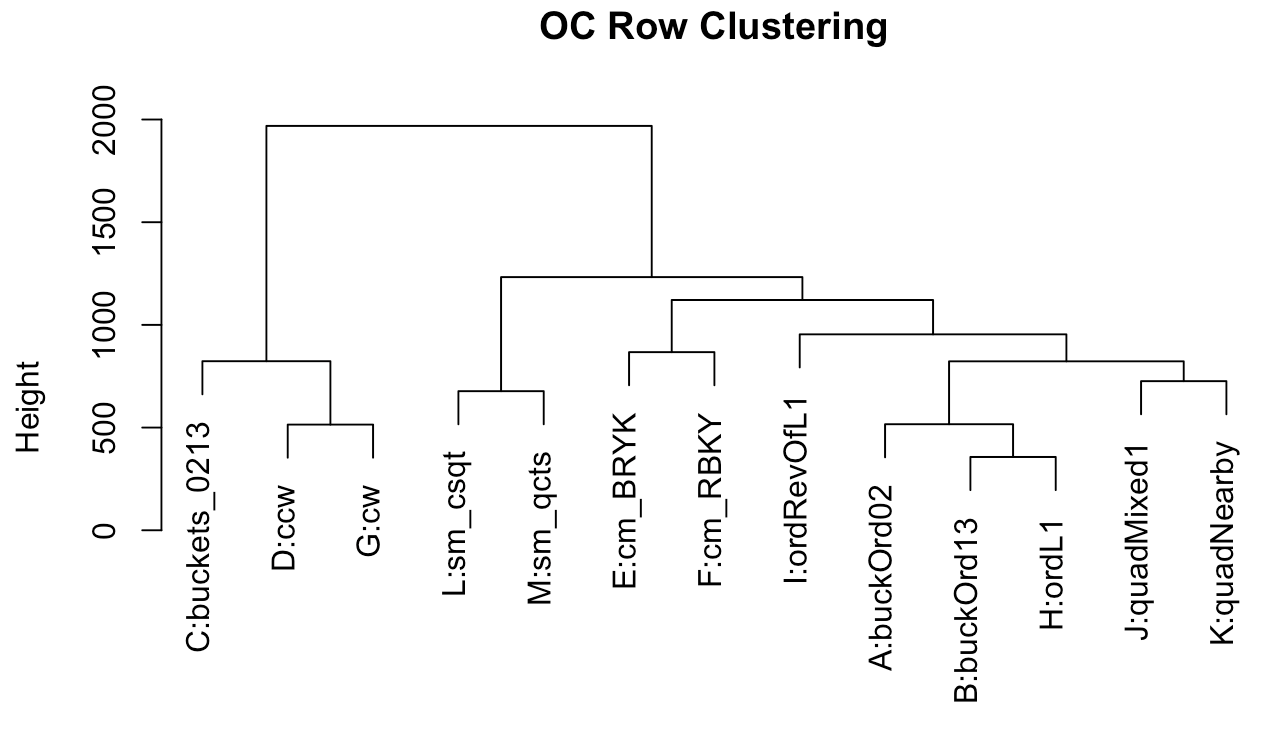}
    \caption{Hierarchical clustering of OC successor-rule transfer behavior. Rules that cluster together are affected similarly by predecessor curricula.}
    \label{fig:oc_transfer_row_cluster}
\end{figure}

The OC row clustering (Fig.~\ref{fig:oc_transfer_row_cluster}) reveals several coherent structural groups. The spatial rules
\[
\texttt{quadNearby}, \quad \texttt{quadMixed1}
\]
form a 
stable 
%\pk{MAYBE SAY 'an early cluster'? } 
%\christo{
early cluster, indicating that they are influenced similarly by predecessor experience. Likewise, the feature-based rules
\[
\texttt{cm\_RBKY}, \quad \texttt{cm\_BRYK}
\]
appear close together, suggesting that OC appears to learn transferable feature-to-bucket abstractions that generalize across related mappings.

The bucket-ordering rules
\[
\texttt{cw}, \quad \texttt{ccw}, \quad \texttt{buckets\_0213}
\]
also form a relatively coherent group in OC. This is consistent with the earlier transfer-learning results showing strong positive transfer within ordering-based rule families.

The FC row clustering exhibits a noticeably different structure. While some local groupings remain visible, the clusters are less stable and more weakly separated. In particular, \texttt{cw} behaves as an outlier in several FC analyses, appearing far from the remaining ordering rules despite belonging conceptually to the same family. This observation is consistent with the earlier finding that FC transfer involving \texttt{cw} was unusually unstable and highly checkpoint-dependent.

Overall, the FC row structure appears less compact and less consistently organized, suggesting that FC does not organize successor-rule sensitivity into a strongly coherent transfer geometry. 
% \pk{IS THERE A WAY, SOMEWHERE IN HERE, TO INTRODUCE THE NOTION THAT THESE ROW AND COLUMN CALCULATOINS ARE KINDS OF 'second order similarity' measure - based on the transfer from one rule to another, rather than their degrees of difficulty.'?}

\subsubsection{Column clustering: similarity of predecessor-rule influence.}

Column clustering groups predecessor rules according to how they affect future learning across many target rules.

\begin{figure}[htbp]
    \centering
    \includegraphics[width=0.82\linewidth]{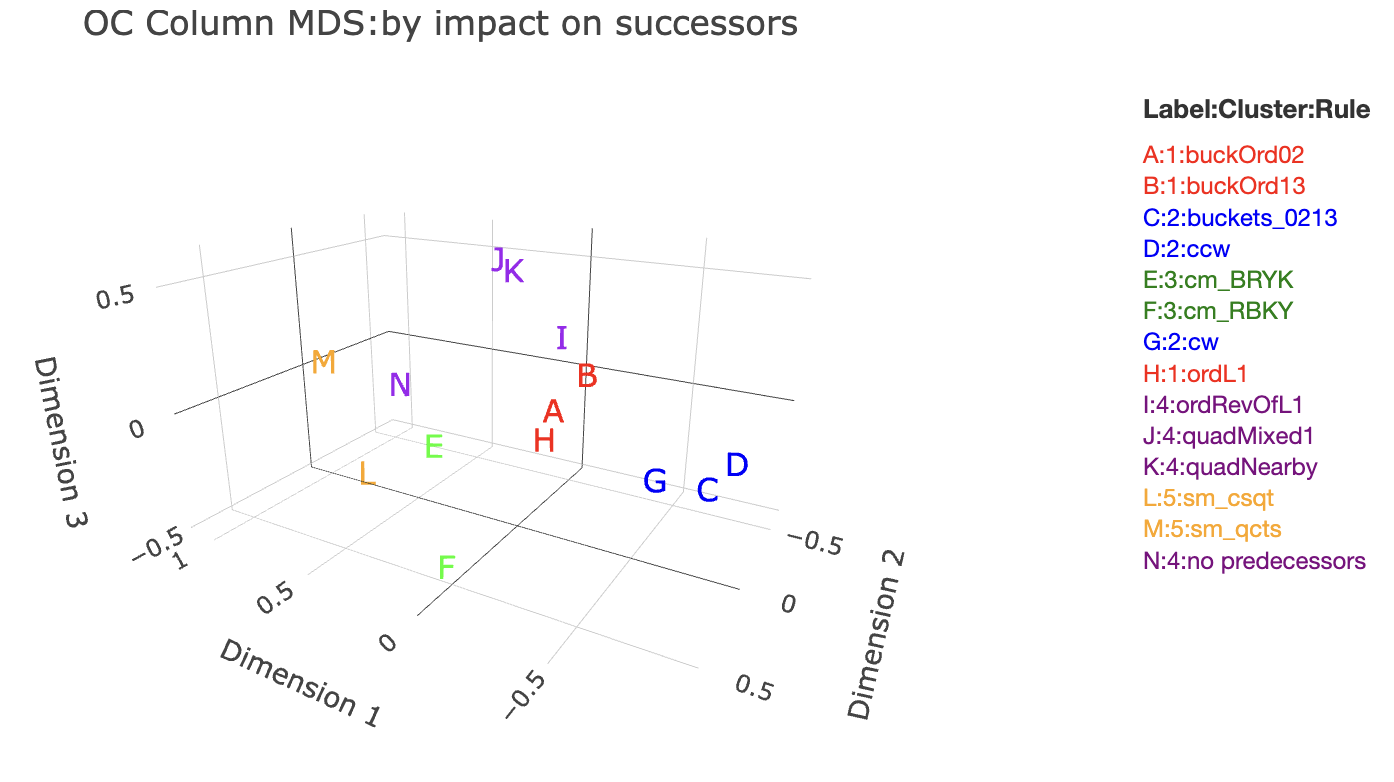}
    \caption{3-dimensional MDS embedding of OC successor-rule transfer behavior. Distances reflect similarity in how rules respond to predecessor experience.}
    \label{fig:oc_transfer_column_mds}
\end{figure}

In OC, the predecessor influence structure is relatively well organized. Ordering rules, spatial rules, and feature-based rules each tend to form recognizable clusters in the MDS embeddings and dendrograms. The corresponding OC column MDS visualization (Fig.~\ref{fig:oc_transfer_column_mds}) shows clear spatial separation between several rule families, particularly between ordering-based and spatial rules.

The FC column structure exhibits weaker geometric separation between rule families. Although some related rules still appear near one another, the embeddings show greater overlap between rule families, indicating that predecessor-rule influence in FC is less systematically organized.

An especially notable observation is that the FC column clustering places several structurally unrelated rules near one another. 
% suggesting 
This observation, that transfer effects in FC are less consistently aligned with the underlying conceptual structure of the rules reinforces our earlier data suggesting that the FC representation does not reflect conceptual structure of rules as well as OC does.

\subsubsection{Comparison between row and column structures.}

To examine whether predecessor-rule influence and successor-rule sensitivity follow similar organizational patterns, the row and column dendrograms were compared using tanglegrams together with quantitative similarity measures.

\begin{figure}[htbp]
    \centering
    \includegraphics[width=0.92\linewidth]{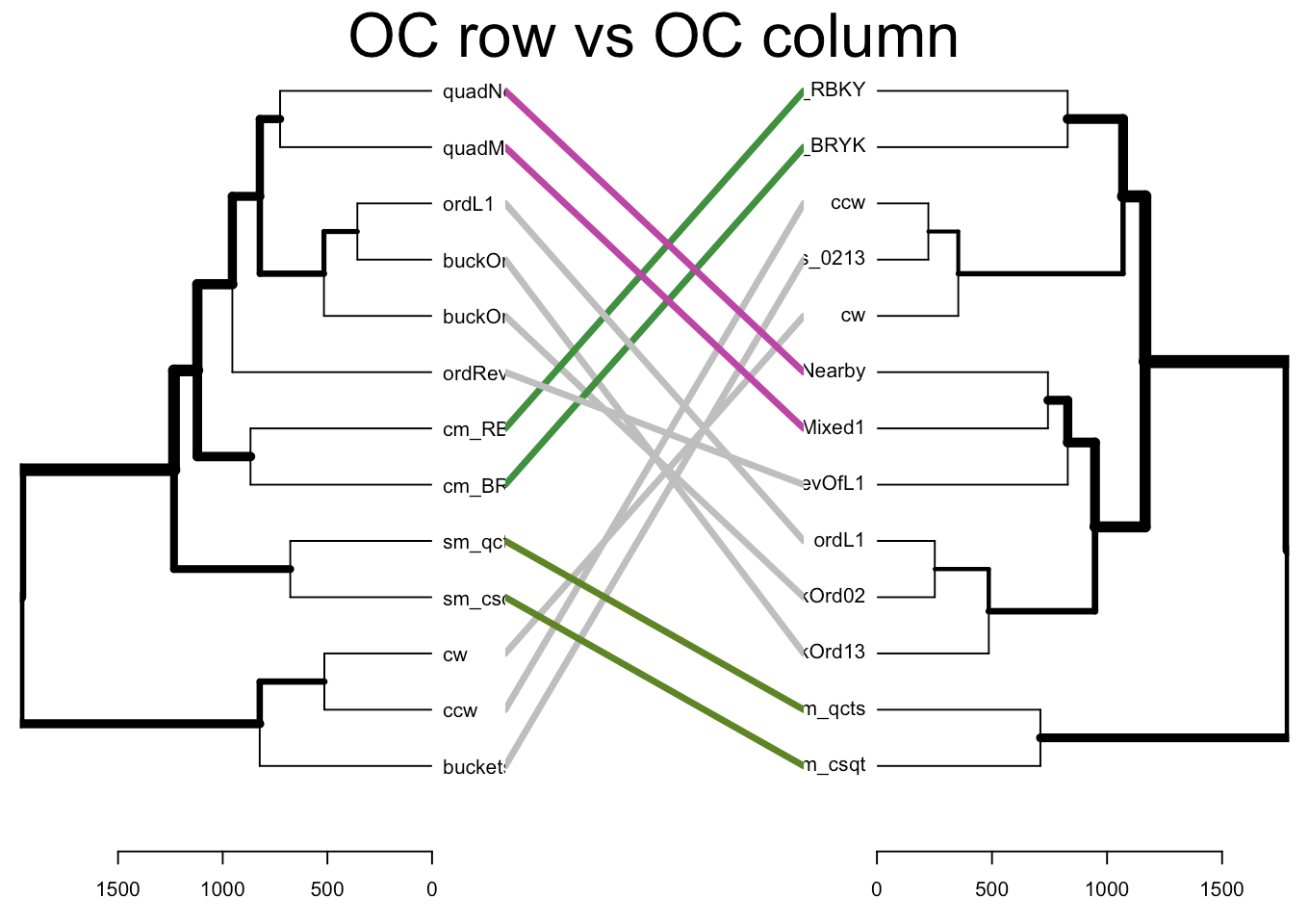}
    \caption{Tanglegram comparing OC row clustering (successor-rule behavior) and OC column clustering (predecessor-rule behavior).}
    \label{fig:oc_row_vs_col_tanglegram}

\end{figure}

The OC row-vs-column comparison (Fig.~\ref{fig:oc_row_vs_col_tanglegram}) shows comparatively strong agreement between the two clustering structures. Several major rule families remain grouped together across both trees, particularly the feature-based rules (\texttt{cm\_RBKY}, \texttt{cm\_BRYK}), the spatial rules (\texttt{quadNearby}, \texttt{quadMixed1}), and the ordering-related rules. The corresponding similarity measures are:

\[
\text{Adjusted Rand Index} = 0.633,
\qquad
\text{Baker's gamma} = 0.542.
\]

These values indicate that, in OC, rules that exhibit similar transfer effects as predecessors also tend to display similar sensitivity as successor rules. In other words, the OC representation develops a relatively coherent and self-consistent transfer-learning geometry.

In contrast, the FC row-vs-column comparison exhibits substantially weaker agreement. The corresponding tanglegram contains noticeably more crossing structure and weaker alignment between clusters, indicating less consistency between predecessor and successor organization. Quantitatively, the similarity values are:

\[
\text{Adjusted Rand Index} = 0.168,
\qquad
\text{Baker's gamma} = 0.039.
\]

This suggests that FC predecessor influence and successor sensitivity are organized according to less stable and less unified transfer relationships.

Cross-representation comparisons further highlight the difference between FC and OC. The FC-vs-OC row comparison shows moderate structural similarity:

\[
\text{Baker's gamma} = 0.445,
\]

indicating that both representations capture some common high-level relationships regarding how rules are influenced by predecessor experience.

%However, 
Unsuprisingly, the FC-vs-OC column comparison shows very weak agreement:

\[
\text{Adjusted Rand Index} = -0.035,
\qquad
\text{Baker's gamma} = -0.063.
\]

This indicates that FC and OC organize predecessor-rule influence in substantially different ways, despite some overlap in successor-rule organization.

The MDS visualizations support these observations. In the OC embeddings, several rule families form relatively compact and well-separated groups. In particular, the feature-mapping rules and spatial rules remain consistently separated from ordering-based rules across both row and column analyses. The OC predecessor-space MDS also places \texttt{quadNearby} and \texttt{quadMixed1} in close proximity, reflecting their similar transfer influence on successor learning.

The FC embeddings, by contrast, are less clearly separated and exhibit greater distortion in cluster organization. Some local relationships remain visible, but the overall geometry is less regular and more strongly affected by individual rules. In particular, \texttt{cw} frequently appears isolated or inconsistently positioned across analyses, consistent with the unstable transfer behavior observed earlier in the simple transfer experiments.
%\pk{IT IS NICE TO SEE, NOW THAT WE ARE AT THE END OF THE PROJECT, THAT WE STARTED OUT WORKING WITH A REPRESENTATION THAT MADE IT HARD TO SEE CONCEPTS.!! }

Overall, these analyses suggest that the OC representation develops a more coherent and structured transfer-learning geometry than FC. Structurally related rules repeatedly cluster together across dendrograms, MDS embeddings, and tanglegram comparisons, and the resulting organization remains comparatively stable between predecessor and successor views. FC transfer behavior, while still containing some local structure, appears more diffuse and internally inconsistent. These findings align with the broader transfer-learning results, suggesting that OC forms stronger relational abstractions that support more systematic transfer across rules.

% \subsubsection{Distance Construction}
% % Explain how rule distances were computed.

% \subsubsection{MDS Visualization}
% % Discuss rule-space geometry.

% \subsubsection{Interpretation of Rule Clusters}
% % Explain what the clusters suggest about representation and rule similarity.

% =========================================================
\section{Human Learning Analysis via Pseudo-Bot Modeling}

\subsection{Motivation}

In addition to studying reinforcement learning agents in GOHR, we also investigated patterns of human learning behavior in two-player settings involving varying levels of pseudo-bot assistance. The goal of this analysis was to determine whether sequences of player actions contain statistically identifiable behavioral signatures that distinguish purely human play from bot-assisted play.

Rather than modeling full board states or explicit strategies, the analysis focuses on the temporal structure of player outcomes during gameplay. This allows us to examine whether differences in consistency, recovery from errors, and progression dynamics can be used to characterize human versus assisted behavior.

% ============================================================
\subsection{Dataset and Problem Setup}
% ============================================================

The analysis was conducted using detailed gameplay transcript files collected from GOHR experiments ( see~\cite{feldman_gallos_wang_menkov_kantor_2026}). Each transcript contains move-level information with columns of the form:

\begin{center}
\small
\texttt{\#playerId, trialListId, seriesNo, ruleId, episodeNo, episodeId,}\\
\texttt{moveNo, timestamp, mover, reactionTime, objectType, objectId,}\\
\texttt{y, x, bucketId, by, bx, code, objectCnt}
\end{center}

The data consists of two-player games in which one side may receive varying levels of pseudo-bot assistance. The objective is to classify whether a sequence of gameplay outcomes is more consistent with 
% human behavior or bot-assisted behavior. 
%\christo{
bot-assisted or unassisted human behaviour. 
% \jf{Confusing terminology because both are human. How about: bot-assisted vs. unassisted human behavior}\pk{DONE}

For preprocessing, moves were grouped by
\[
(\texttt{playerId}, \texttt{ruleId}, \texttt{mover}),
\]
and sorted chronologically using
\[
(\texttt{episodeNo}, \texttt{episodeId}, \texttt{moveNo}).
\]

Each resulting sequence corresponds to the history of one player side interacting with a particular rule.

% ============================================================
\subsection{Behavioral Modeling Approach}
% ============================================================

\subsubsection{Outcome Sequence Representation}

The raw gameplay stream was converted into a filtered sequence of move outcomes based on changes in the number of remaining objects. 
% \pk{IT'S INTERESTING THAT YOU DID THIS, SINCE THE code DATA ALREADY CONTAINS INFORMATION ABOUT WHETHER THE MOVE WAS CORRECT OR NOT.}

Let $n_t$ denote the object count after move $t$, and $n_{t-1}$ the object count before the move.

% For two-player games, the move outcome $o_t$ was defined as follows:
%\christo{
For two-player games, accepted and denied outcomes can be inferred from changes in the number of remaining objects and turn transitions. Specifically:

\begin{itemize}
    \item If
    \[
    n_t < n_{t-1},
    \]
    the move is considered successful:
    \[
    o_t = A
    \]
    where $A$ denotes an accepted or correct move.

    \item If
    \[
    n_t = n_{t-1},
    \]
    and the turn passes to the opposing player, the move is considered unsuccessful:
    \[
    o_t = D
    \]
    where $D$ denotes a denied or incorrect move.

    \item If
    \[
    n_t = n_{t-1},
    \]
    and the same player continues, the move is treated as an intermediate state and excluded from the sequence. 
    % \jf{These definitions follow from the rules, so it seems a little weird to give them as definitions.}\pk{I THINK IT IS OK TO LEAVE A LITTLE WEIRDNESS} 
    
    \item If
    \[
    n_t > n_{t-1},
    \]
    the event is treated as a boundary or reset condition and excluded.
\end{itemize}

This preprocessing converts the raw interaction stream into a filtered outcome sequence:
\[
o_1, o_2, \dots, o_K,
\quad
o_t \in \{A,D\}.
\]

For each timestep, two contextual variables were additionally computed.

\paragraph{Previous Outcome.}
The previous valid outcome is represented by $o_{t-1}$. For the first move in a sequence, a special start token $S$ is used.

\paragraph{Success-Rate Context.}
The cumulative success rate prior to timestep $t$ is defined as:
\[
r_{t-1}
=
\frac{
\sum_{i=1}^{t-1} \mathbf{1}(o_i = A)
}{
t-1
}.
\]

The cumulative
% continuous
%\pk{Is this the same as ``cumulative''? If so, why not use the same word? Are the bin percentages?}\christo{corrected}
success-rate value is discretized into bins:
\[
s_{t-1} \in
\{
0\_25,\,
25\_50,\,
50\_75,\,
75\_100
\}.
\]

The first timestep uses a dedicated \texttt{start} bin.

% ============================================================
\subsubsection{Markov / Backoff Model}
% ============================================================

Separate probabilistic models were trained for the two behavioral classes:
\[
c \in \{\texttt{unassisted}, \texttt{assisted}\}. 
\]

% \jf{See comment above. Perhaps: \{ assisted, unassisted \} } \pk{DONE}

The primary quantity estimated by the model is
\[
P(o_t \mid o_{t-1}, s_{t-1}, c),
\]
which represents the probability of the next outcome conditioned on the previous outcome, the cumulative success-rate bin, and the behavioral class.

From the training data, we computed
\[
N_c(o_{t-1}, s_{t-1}, o_t),
\]
the number of times outcome $o_t$ follows context $(o_{t-1}, s_{t-1})$ within class $c$.

Similarly,
\[
N_c(o_{t-1}, s_{t-1})
\]
counts the total number of occurrences of the context itself.

The full conditional probability was estimated using Laplace smoothing:
\[
P_{\text{full}}
=
\frac{
N_c(o_{t-1}, s_{t-1}, o_t) + \alpha
}{
N_c(o_{t-1}, s_{t-1}) + 2\alpha
},
\]
with
\[
\alpha = 1.
\]

To reduce sparsity, two lower-order backoff probabilities were additionally estimated.

\paragraph{Previous-outcome model.}
\[
P_{\text{prev}}
=
P(o_t \mid o_{t-1}, c).
\]

\paragraph{Global class model.}
\[
P_{\text{global}}
=
P(o_t \mid c).
\]

Both lower-order probabilities were estimated using the same Laplace smoothing procedure (\(\alpha = 1\)) as the full model.

The final probability estimate was computed as a weighted interpolation: 
%\pk{WAS THERE SOME EXPERIMENTATION TO SEE IF THESE WERE GOOD WEIGHTS?}\christo{YES, SOME EXPERIMENTS WERE CONDUCTED EVEN THOUGH ITS NOT EXHAUSTIVE. ADDED THE LINE BELOW}
\[
P(o_t \mid o_{t-1}, s_{t-1}, c)
=
0.7P_{\text{full}}
+
0.2P_{\text{prev}}
+
0.1P_{\text{global}}.
\]

%\christo{
The interpolation weights were selected after limited empirical experimentation with alternative weight combinations. 
% The classifier performance was relatively robust to moderate changes in the weights, and the final values were chosen because they consistently yielded strong performance while emphasizing the highest-order context. %} 
The final values place most of the probability mass on the full contextual model while retaining smaller contributions from the lower-order backoff models.
% \pk{WHAT DOES HIGHEST ORDER CONTEXT MEAN?}
This combination produced strong classification performance while providing robustness when exact contexts occurred infrequently in the training data.
% This interpolation provides robustness when exact contexts occur infrequently in the training data.

% ============================================================
\subsubsection{Sequence Classification}
% ============================================================

Given the first $K$ valid outcomes of a test sequence
\[
o_1, o_2, \dots, o_K,
\]
the model computes the log-likelihood under each behavioral class:
\[
\log L_c
=
\sum_{t=1}^{K}
\log
P(o_t \mid o_{t-1}, s_{t-1}, c).
\]

A classification score is then defined as
\[
\text{score}
=
\log L_{\text{assisted}}
-
\log L_{\text{unassisted}}.
\]

The final prediction is:
\[
\hat{y}
=
\begin{cases}
\texttt{assisted} & \text{if score} > 0 \\
\texttt{unassisted} & \text{otherwise}.
\end{cases}
\]

Thus, the model evaluates whether the temporal progression of outcomes is statistically more consistent with %\christo{
unassisted or bot-assisted human behavior.

% ============================================================
\subsection{Evaluation Protocol}
% ============================================================

The classifier was evaluated separately for multiple pseudo-bot assistance levels:
\[
\texttt{ph4},
\texttt{ph8},
\texttt{ph12},
\texttt{ph16}.
\]

For each setting, the model was trained on outcome sequences extracted from gameplay transcripts and evaluated using sequence-level classification metrics.

Performance was measured using accuracy, precision, recall, F1-score, and specificity.

Both test-only and combined train+test evaluations were examined to analyze the stability of the learned behavioral distributions.

% ============================================================
\subsection{Results}
% ============================================================

\subsubsection{Bot-Assisted Side Classification}

% The Markov/backoff classifier successfully distinguished human gameplay from pseudo-bot-assisted gameplay across all evaluated assistance levels.
%\christo{
The Markov/backoff classifier achieved above-chance discrimination between unassisted and pseudo-bot-assisted gameplay across all evaluated assistance levels, although performance varied substantially across conditions. %}

Table~\ref{tab:pseudo_bot_results} summarizes the classification performance across different pseudo-bot configurations. %\pk{
The code ``phX'' refers to the pseudo-half-life of the bot. For ph4, the chance that the suggestion is wrong decreses by 50\% every 4 suggestions.  The strongest performance was obtained for the \texttt{ph4} 
% \color{red}
(fastest-learning) 
% \color{black} 
setting, which achieved an accuracy of $93.5\%$ and an F1-score of $0.964$ under the test-only evaluation setting.

% \color{red} 
Lower 
% \color{black} 
%\jf{Fixed word}
levels of assistance remained distinguishable, although performance varied depending on the behavioral overlap between human and assisted play. Across most configurations, F1-scores ranged approximately between $0.55$ and $0.86$, demonstrating that temporal outcome patterns contain meaningful information for identifying assisted gameplay.

\begin{table}[htb]
\centering
\caption{Classification performance of the Markov/backoff model for distinguishing human and pseudo-bot-assisted gameplay.}
\label{tab:pseudo_bot_results}
\begin{tabular}{lcccccc}
\toprule
\textbf{Setting} & \textbf{TN} & \textbf{FP} & \textbf{FN} & \textbf{TP} & \textbf{Accuracy} & \textbf{F1} \\
\midrule
\texttt{ph4\_Test}  & 2  & 1  & 1  & 27 & 0.935 & 0.964 \\
\texttt{ph8\_Test}  & 6  & 5  & 6  & 34 & 0.784 & 0.861 \\
\texttt{ph12\_Test} & 14 & 8  & 14 & 18 & 0.593 & 0.621 \\
\texttt{ph16\_Test} & 18 & 7  & 13 & 20 & 0.655 & 0.667 \\
\midrule
\texttt{ph4\_Test+Train}  & 9  & 8  & 3  & 30 & 0.780 & 0.845 \\
\texttt{ph8\_Test+Train}  & 21 & 12 & 6  & 38 & 0.766 & 0.809 \\
\texttt{ph12\_Test+Train} & 24 & 12 & 11 & 27 & 0.689 & 0.701 \\
\texttt{ph16\_Test+Train} & 22 & 12 & 17 & 18 & 0.580 & 0.554 \\
\bottomrule
\end{tabular}
\end{table}

% ============================================================
\subsubsection{Effect of Number of Moves}
% ============================================================

% The model operates on the first $K$ valid outcomes of each sequence. Increasing the number of observed moves generally improves classification confidence by providing more evidence for the underlying behavioral distribution. \jf{Is there any statistical corroboration of empirical claims like this (changing X improves Y)? In Psychology one usually doesn't draw conclusions like this without some sort of confirmatory statistical test.}

Because the classifier operates on the first \(K\) valid outcomes of each sequence, its predictions are based on a finite amount of behavioral evidence. Longer sequences provide additional observations of outcome transitions and success-rate dynamics, which may improve the stability of the likelihood estimates. However, the relationship between sequence length and classification performance was not evaluated systematically in the present study.

Early portions of gameplay often contain exploratory behavior shared between humans and assisted agents, while longer sequences expose differences in consistency and adaptation patterns.

% ============================================================
\subsubsection{Rule-Level Differences}
% ============================================================

% Performance varied across rules and assistance levels. Rules with more complex spatial or relational dependencies tended to produce noisier behavioral sequences, reducing separability between the two classes.

%\christo{
Qualitatively, for some rules it appeared more difficult to distinguish assistance from non-assistance, than for others. One possible explanation is that rules with more complex spatial or relational dependencies produce more variable behavioral sequences, thereby reducing separability between assisted and unassisted play. However, this hypothesis was not tested directly in the present analysis.

At the same time, simpler rules often produced more stable outcome dynamics, making bot-assisted behavior easier to distinguish from purely human play.

% ============================================================
\subsection{Discussion}
% ============================================================

The results suggest that temporal patterns of successes and failures contain sufficient statistical structure, in this dataset, to distinguish between assisted and unassisted gameplay behavior.
% human gameplay from pseudo-bot-assisted gameplay.

Importantly, the model does not rely on explicit board-state representations or handcrafted strategic features. Instead, it captures behavioral differences through short-term outcome transitions and cumulative performance context.

The interpolation-based Markov/backoff formulation provides a simple but effective mechanism for handling sparse behavioral contexts while remaining interpretable.

Several limitations remain. The current model uses only binary outcome sequences and ignores richer information such as reaction times, object identities, spatial board configurations, and longer-term strategic dependencies. In addition, the discretization of success rates introduces information loss and may limit sensitivity to subtle behavioral changes. 
% \jf{Yes, I also thought that! Is it possible to do this analysis without discretizing?} \pk{I THINK WE ARE LEAVING THIS FOR THE FUTURE?}

Nevertheless, the analysis demonstrates that relatively simple probabilistic sequence models can recover meaningful distinctions between unassisted  and assisted gameplay behavior in GOHR.

% \section{Human Learning Analysis via Pseudo-Bot Modeling}

% \subsection{Motivation}
% % Explain why human-vs-bot-assisted learning is analyzed.

% \subsection{Dataset and Problem Setup}
% % Describe human-human games, pseudo-bot assistance, moves, outcomes, and labels.

% \subsection{Behavioral Modeling Approach}
% \subsubsection{Outcome Sequence Representation}
% % Accepted/Denied/Immovable or success/failure sequences.

% \subsubsection{Markov/Backoff Model}
% % P(outcome_t | previous outcome, success-rate bin, class)

% \subsubsection{Feature-Based or Hybrid Classifier}
% % If included, describe logistic regression or hybrid Markov + classifier.

% \subsection{Evaluation Protocol}
% % Explain train/test split, first K moves, metrics.

% \subsection{Results}
% \subsubsection{Bot-Assisted Side Classification}
% \subsubsection{Effect of Number of Moves}
% \subsubsection{Rule-Level Differences}

% \subsection{Discussion}
% % Explain what the model suggests about human learning, assistance effects, and limitations.

% =========================================================

% =========================================================
% =========================================================
\section{Overall Discussion and Conclusion}

\subsection{Key Findings}

This work investigated hidden-rule learning in the Game of Hidden Rules (GOHR) through reinforcement learning, transfer learning, transfer-geometry analysis, and pseudo-bot-assisted human learning analysis. Across these experiments, several consistent patterns emerged regarding the role of representation, rule structure, and structural transfer.

One of the clearest findings is that state representation strongly influences hidden-rule learning behavior. Across independent-rule learning, transfer learning, clustering analyses, and transfer-geometry experiments, the Object-Centric (OC) representation consistently produced more stable and interpretable behavior than the Feature-Centric (FC) representation. Systems using the  OC generally learned rules more efficiently, exhibited stronger positive transfer within related rule families, and formed clearer geometric organization in transfer space. In contrast, FC transfer behavior was often more diffuse and sensitive to individual training checkpoints. The \texttt{cw} rule, in particular, exhibited unstable FC transfer behavior despite belonging conceptually to the ordering-rule family, suggesting that FC representations may rely more heavily on incidental spatial training dynamics rather than stable relational abstractions.

The experiments further demonstrated that learning difficulty depends strongly on the structural properties of the underlying rule. Feature-mapping rules were generally easier to learn, while ordering-based, spatial, and compound rules produced substantially greater learning difficulty. The rule-property analysis showed that increasing relational abstraction and compositional complexity leads to slower convergence and higher variability. Compound rules combining multiple relational constraints were consistently among the most difficult tasks for both representations.

Transfer-learning experiments revealed that successful transfer depends primarily on structural alignment between predecessor and successor rules. Positive transfer was strongest when predecessor rules shared meaningful relational components with the target rule. In the compound-transfer experiments, full-component pretraining consistently accelerated learning, while unrelated or partially related curricula often produced weaker transfer or interference. The simple-to-simple transfer experiments showed similar trends, particularly within ordering-based and feature-based rule families. These findings suggest that effective curricula for hidden-rule learning should preserve structural continuity between tasks rather than permitting  arbitrary sequencing.

The transfer-geometry analyses provided additional insight into how the two representations organize relational knowledge. Hierarchical clustering, multidimensional scaling (MDS), and tanglegram comparisons showed that OC develops a comparatively coherent transfer-learning geometry. Structurally related rules repeatedly formed stable clusters across row clustering, column clustering, and MDS embeddings. Moreover, predecessor-rule influence and successor-rule sensitivity exhibited relatively strong agreement in OC, indicating that OC develops internally consistent transfer structures. FC exhibited substantially weaker alignment between predecessor and successor organization, suggesting that its transfer relationships are less stable and less systematically structured.

The pseudo-bot-assisted human-learning analysis demonstrated that temporal patterns of successes and failures contain sufficient statistical structure to distinguish assisted and non-assisted gameplay. Even without explicit board-state information, the Markov/backoff sequence model was able to capture behavioral differences using short-term outcome transitions and cumulative success-rate context. Several observations from the reinforcement-learning experiments parallel the human-learning results. In both settings, ordering-based and spatial rules produced more complex learning dynamics than direct feature-mapping rules, and structurally related tasks exhibited stronger transfer relationships. Although exploratory, these experiments suggest that GOHR provides a useful unified framework for studying both machine and human hidden-rule learning behavior.

Overall, the results suggest that explicit object-level representations better support relational abstraction, transferable learning, and coherent transfer organization in hidden-rule environments. More broadly, the experiments indicate that hidden-rule learning is governed not only by task difficulty, but also by the structural relationships between rules and the representations used to encode them.

\subsection{Limitations}\label{subsec:Limitations}

Several limitations should be considered when interpreting the results of this work.

First, many experiments were conducted using a relatively small number of independent runs due to computational constraints. Although median-based aggregation and ratio-based transfer analyses reduced some variability, additional runs would improve statistical robustness and confidence. 
In addition, our exploration of the data included so many comparisons (literally, hundreds) that even with the most sophisticated adjustments for multiple comparisons, we cannot assert that any of our conclusions have been established ``with such and such degree of confidence/'' All of this work is exploratory, to help define a path toward principled exploration of the ``geometry of rules and concepts.'' [See Section~\ref{subsec:Future Work} ]

Second, the rule suite used in this work represents only a subset of the possible hidden-rule space in GOHR. Additional rule families, larger compositional structures, and more diverse relational dependencies may reveal transfer behaviors not 
exposed   in the current experiments.

Third, for the MDS analysis, the  data in the transfer matrices are missing many  predecessor-successor combinations, requiring imputation during clustering and MDS analyses. Although complementary row-mean and column-mean imputations were used to preserve needed  statistical structures, the resulting inferred  geometry may still be influenced by incomplete transfer coverage. In companion studies we have worked with complete transfer data, supporting assessment of the effects of this imputation.  

%\pk{Christo: Might we include the .Rmd code for the clustering and MDS, in an appendix to show how the imputation was done. Or perhaps describe it in a few equations or words? }

Fourth, the reinforcement-learning agents 
%\pk{
that we have studied rely 
%\sout{relied} } 
on relatively simple exploration strategies and memory mechanisms. More advanced exploration methods or longer-horizon memory architectures may improve learning efficiency for difficult relational tasks. 
%\pk{IT IS POSSIBLE THAT YOUR REFACTORING MAY SHOW BIG DIFFERENCES IN THE M* AND OTHER MEASURES. THAT COULD BE VERY GOOD -- IF IT IS FASTER LEARNING. BUT THEN WE HAVE TO FIGURE OUT HOW TO PRESENT IT. CAN IT BE A WHOLE SEPARATE TECHNICAL REPORT? }

We did experiment with using a  modular Gymnasium-compatible GOHR environment with the goal of making it easy to let others use the GOHR as a tool for future studies on relational reasoning, abstraction learning, curriculum learning, transfer learning, and human-agent comparison in hidden-rule environments. Our specific refactoring did not retain enough information about the history of boards and moves, and did not yield performance comparable to our legacy system, See Appendix~\ref{appendix:gymnasium_env} 

%Finally, the human-learning analysis used pseudo-bot-assisted behavioral modeling rather than direct cognitive modeling of human reasoning processes. \jf{JF suggests: 
Finally, the human learning analysis focuses on classification of human learning into bot-assisted vs unassisted performance, rather than on direct modeling of human reasoning processes. The current model captures short-term outcome dynamics but does not incorporate richer behavioral information such as reaction times, spatial reasoning strategies, or long-term planning behavior.

\subsection{Future Work}\label{subsec:Future Work}

Several directions remain for future research.

One important direction is expanding the GOHR rule suite to include more complex relational, temporal, and compositional rules. A larger and more diverse rule space would allow deeper investigation of abstraction, curriculum learning, and transfer geometry.

Future work could also explore stronger reinforcement-learning approaches, including improved exploration strategies, memory-augmented architectures, offline reinforcement learning, or model-based methods. These approaches may improve learning efficiency for difficult relational rules and reduce transfer instability.

The transfer-learning framework can be extended further through adaptive curricula, automated curriculum generation, and transfer-aware task sequencing. Such methods may help identify optimal training orders for accelerating hidden-rule learning.

Another promising direction is improving statistical robustness through larger experimental sweeps and broader predecessor-successor coverage. More complete transfer matrices would provide stronger clustering structure and more reliable transfer geometries.

The human-learning analysis could also be extended substantially. Future work may incorporate richer behavioral features, direct human strategy modeling, cognitive comparisons, or human-in-the-loop learning experiments. Comparing learned reinforcement-learning representations against human behavioral abstractions may provide additional insight into how relational rules are internally represented across biological and artificial systems.

% \pk{I WOULD DELETE THIS HERE AND USE THE SUGGESTED TEXT AT THE END OF LIMITATIONS:Finally, the modular Gymnasium-compatible GOHR environment developed in this work provides a foundation for future studies on relational reasoning, abstraction learning, curriculum learning, transfer learning, and human-agent comparison in hidden-rule environments.}

\section{Acknowledgments}\label{sec:Acknowledgments}

The systems used here to study the GOHR were developed at the University of Wisconsin. Support for their research was provided by the University of Wisconsin-Madison Office of the Vice
Chancellor for Research and Graduate Education with funding from the Wisconsin Alumni Research
Foundation, and by the National Science Foundation under Grant No. 2041428.  Additional participants in that development are acknowledged in~\cite{pulick_comparing_2024}. The specific work reported here  was supported  in part by the Defense Advanced Research Projects
Agency (HR00112420363); the content of this report   does not necessarily reflect the position or the policy of the
Government; and no official endorsement should be inferred. Distribution Statement: Approved for
public release; distribution is unlimited.  Any opinions,
findings, and conclusions or recommendations expressed in this material are those of the author(s)
and do not necessarily reflect the views of the sponsors.    

% =========================================================
% \bibliographystyle{plain}
% \bibliography{references}
\printbibliography

@techreport{bier_can_2019,
	title = {Can {We} {Distinguish} {Machine} {Learning} from {Human} {Learning}?},
	url = {http://arxiv.org/abs/1910.03466},
	author = {Bier, Vicki and Kantor, Paul B. and Lupyan, Gary and Zhu, Xiaojin},
	year = {2019},
	note = {\_eprint: 1910.03466},
    institution = {arXiv},
}

@techreport{pulick_game_2022,
	title = {The {Game} of {Hidden} {Rules}: {A} {New} {Kind} of {Benchmark} {Challenge} for {Machine} {Learning}},
	shorttitle = {The {Game} of {Hidden} {Rules}},
	url = {http://arxiv.org/abs/2207.10218},
	number = {arXiv:2207.10218},
	urldate = {2022-11-20},
	institution = {arXiv},
	author = {Pulick, Eric and Bharti, Shubham and Chen, Yiding and Menkov, Vladimir and Mintz, Yonatan and Kantor, Paul and Bier, Vicki M.},
	month = jul,
	year = {2022},
	doi = {10.48550/arXiv.2207.10218},
	note = {arXiv:2207.10218 [cs]
type: article},
}

@article{pulick_comparing_2024,
	title = {Comparing {Reinforcement} {Learning} and {Human} {Learning} with the {Game} of {Hidden} {Rules}},
	issn = {2169-3536},
	doi = {10.1109/ACCESS.2024.3395249},
	journal = {IEEE Access},
	author = {Pulick, Eric and Menkov, Vladimir and Mintz, Yonatan D. and Kantor, Paul B. and Bier, Vicki M.},
	year = {2024},
	pages = {65362--65372},
}

@misc{bier_gohr_nodate,
	title = {{GOHR} {Home} {Page}},
	url = {https://rulegame.wisc.edu/},
	urldate = {2024-05-14},
	journal = {Game of Hidden Rules},
	author = {Bier, Vicki and Kantor, Paul and Menkov, Vladimir},
	year = {2024},
}

@inproceedings{mnih2016asynchronous,
  title={Asynchronous methods for deep reinforcement learning},
  author={Mnih, Volodymyr and Badia, Adria Puigdomenech and Mirza, Mehdi and Graves, Alex and Lillicrap, Timothy and Harley, Tim and Silver, David and Kavukcuoglu, Koray},
  booktitle={International conference on machine learning},
  pages={1928--1937},
  year={2016},
  organization={PmLR}
}

@article{mathew2025toward,
  title={Toward a Metrology for Artificial Intelligence: Hidden-Rule Environments and Reinforcement Learning},
  author={Mathew, Christo and Wang, Wentian and Feldman, Jacob and Gallos, Lazaros K and Kantor, Paul B and Menkov, Vladimir and Wang, Hao},
  journal={arXiv preprint arXiv:2509.06213},
  year={2025}
}

@misc{feldman_gallos_wang_menkov_kantor_2026,
 title={Benefits of co-learning with an AI agent},
 url={osf.io/preprints/psyarxiv/bx5q4_v2},
 publisher={PsyArXiv},
 author={Feldman, Jacob and Gallos, Lazaros and Wang, Hao and Menkov, Vladimir and Kantor, Paul B.}, year={2026},
 month=may
 }

@article{vaswani2017attention,
  title={Attention is all you need},
  author={Vaswani, Ashish and Shazeer, Noam and Parmar, Niki and Uszkoreit, Jakob and Jones, Llion and Gomez, Aidan N and Kaiser, {\L}ukasz and Polosukhin, Illia},
  journal={Advances in neural information processing systems},
  volume={30},
  year={2017}
}

@article{towers2026gymnasium,
  title={Gymnasium: A standard interface for reinforcement learning environments},
  author={Towers, Mark and Kwiatkowski, Ariel and Balis, John and De Cola, Gianluca and Deleu, Tristan and Goul{\~a}o, Manuel and Andreas, Kallinteris and Krimmel, Markus and Kg, Arjun and Perez-Vicente, Rodrigo and others},
  journal={Advances in Neural Information Processing Systems},
  volume={38},
  year={2026}
}

% =========================================================

\appendix

\titleformat{\section}[display]
  {\normalfont\LARGE\bfseries}
  {Appendix \thesection}
  {0.5em}
  {}
  
\section{Gymnasium-Compatible GOHR Environment}
% \section*{Appendix A: Gymnasium-Compatible GOHR Environment}
% \addcontentsline{toc}{section}{Appendix A: Gymnasium-Compatible GOHR Environment}
\label{appendix:gymnasium_env}

% To improve modularity and compatibility with modern reinforcement learning frameworks, a Gymnasium-compatible implementation of GOHR was developed. The objective of this refactor was not to modify the underlying game mechanics or hidden-rule definitions, but rather to provide a cleaner interface for experimentation and future development.

% It is important to note that all experiments reported in this work used the legacy observation pipeline described in Section~\ref{sec:gohr_environment}. The Gymnasium-compatible environment is included here as a software contribution that supports future research and development.

This appendix describes an experimental refactoring of GOHR into a Gymnasium-compatible environment~\cite{towers2026gymnasium}. The primary motivation was to improve software modularity, support modern reinforcement-learning libraries, and separate environment dynamics from state-representation construction.

The refactoring also provided an opportunity to investigate whether GOHR could be represented using a more conventional reinforcement-learning interface, in which observations contain only the current environment state and temporal history is modeled by the learning algorithm rather than by the environment itself.

However, preliminary experiments indicated that this simplified observation design did not reproduce the learning behavior observed with the legacy GOHR implementation, particularly for rules that depend on information from previous states and actions. As a result, all experiments reported in the main body of this thesis use the legacy history-augmented observation pipeline. The Gymnasium implementation is therefore presented as both a software contribution and a case study illustrating the importance of the legacy observation design for GOHR.

\subsection{Design Goals}

The refactored environment was designed with the following objectives:

\begin{itemize}
    \item Provide a standard Gymnasium-compatible interface.
    % \item Decouple environment dynamics from state representation construction.
    \item Decouple environment dynamics from state-representation construction in order to evaluate alternative observation designs.
    \item Support both Feature-Centric (FC) and Object-Centric (OC) representations through a unified API.
    \item Enable integration with modern reinforcement learning libraries such as Stable-Baselines3.
    \item Improve maintainability, extensibility, and debugging capabilities.
\end{itemize}

\subsection{Gymnasium Interface}

The environment follows the standard Gymnasium API:

\begin{itemize}
    \item \texttt{reset(seed, options)} $\rightarrow$ $(\text{observation}, \text{info})$
    \item \texttt{step(action)} $\rightarrow$ $(\text{observation}, \text{reward}, \text{terminated}, \text{truncated}, \text{info})$
\end{itemize}

Unlike the legacy implementation, the default observation corresponds only to the current environment state. Temporal history is not automatically included in the observation and may instead be modeled explicitly by the agent or through optional wrappers.

\subsection{Support for Multiple Representations}

The environment supports both state representations used throughout GOHR research:

\begin{enumerate}
    \item \textbf{Feature-Centric (FC)} representation based on spatial feature maps.
    \item \textbf{Object-Centric (OC)} representation based on explicit object attributes.
\end{enumerate}

Both representations are accessible through the same environment interface, allowing experiments to be conducted without modifying the underlying game logic.

\subsection{Legacy Compatibility Wrappers}

To preserve compatibility with previous GOHR experiments, optional wrappers are provided that reproduce the legacy observation pipeline.

These wrappers can:

\begin{itemize}
    \item maintain a history of previous successful states,
    \item include previously executed actions,
    \item construct history-augmented observations matching those used in earlier studies.
\end{itemize}

As a result, the same environment can operate in either a standard Gymnasium mode or a legacy-compatible mode.

\subsection{Observations from the Refactoring Experiment}

A central difference between the Gymnasium implementation and the legacy GOHR environment is the treatment of temporal history. The default Gymnasium design exposes only the current board state, whereas the legacy environment incorporates previous successful states and actions directly into the observation.

\begin{figure}[htbp]
    \centering

    \begin{subfigure}{0.48\linewidth}
        \centering
        \includegraphics[width=\linewidth]{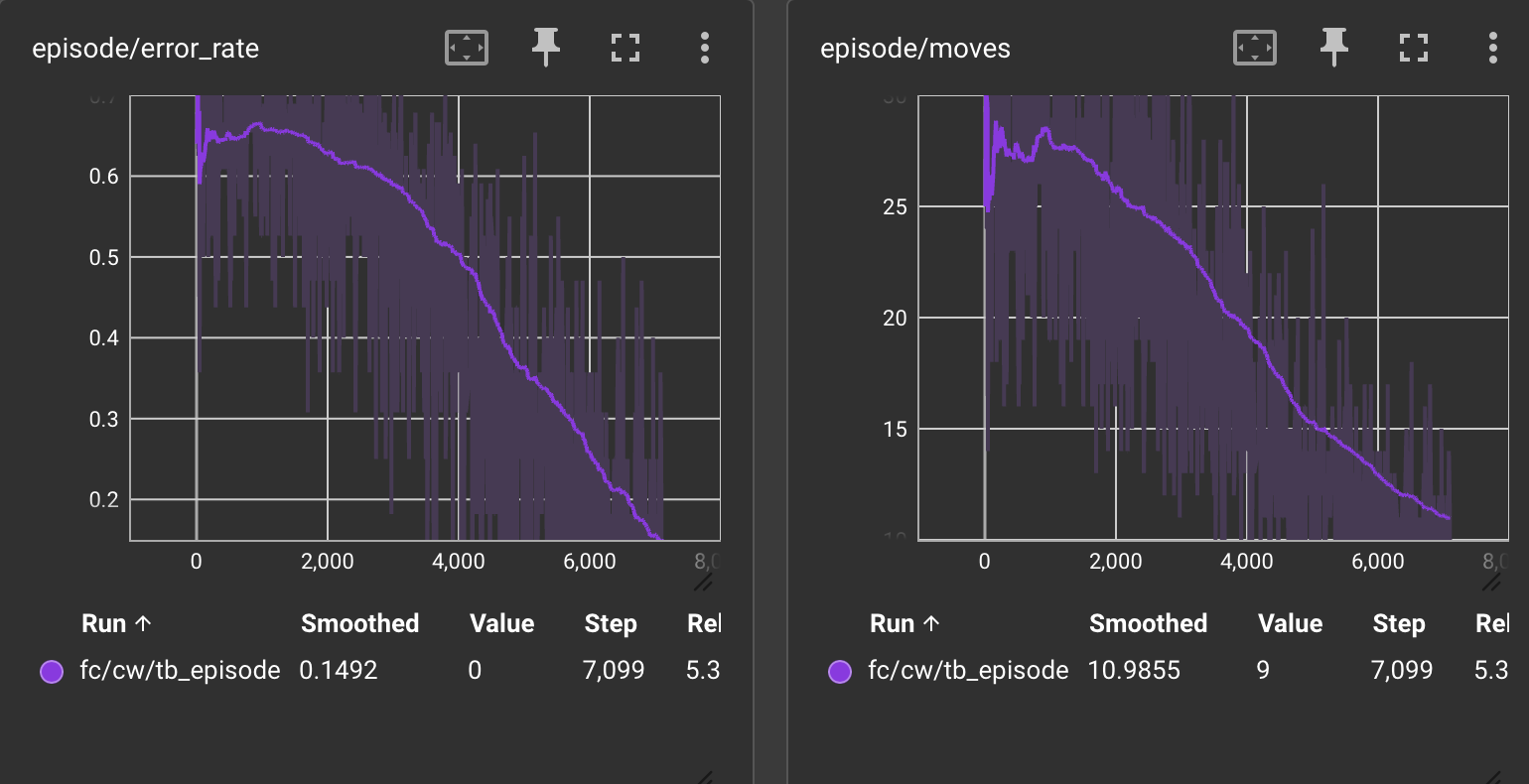}
        \caption{FC, legacy=True}
    \end{subfigure}
    \hfill
    \begin{subfigure}{0.48\linewidth}
        \centering
        \includegraphics[width=\linewidth]{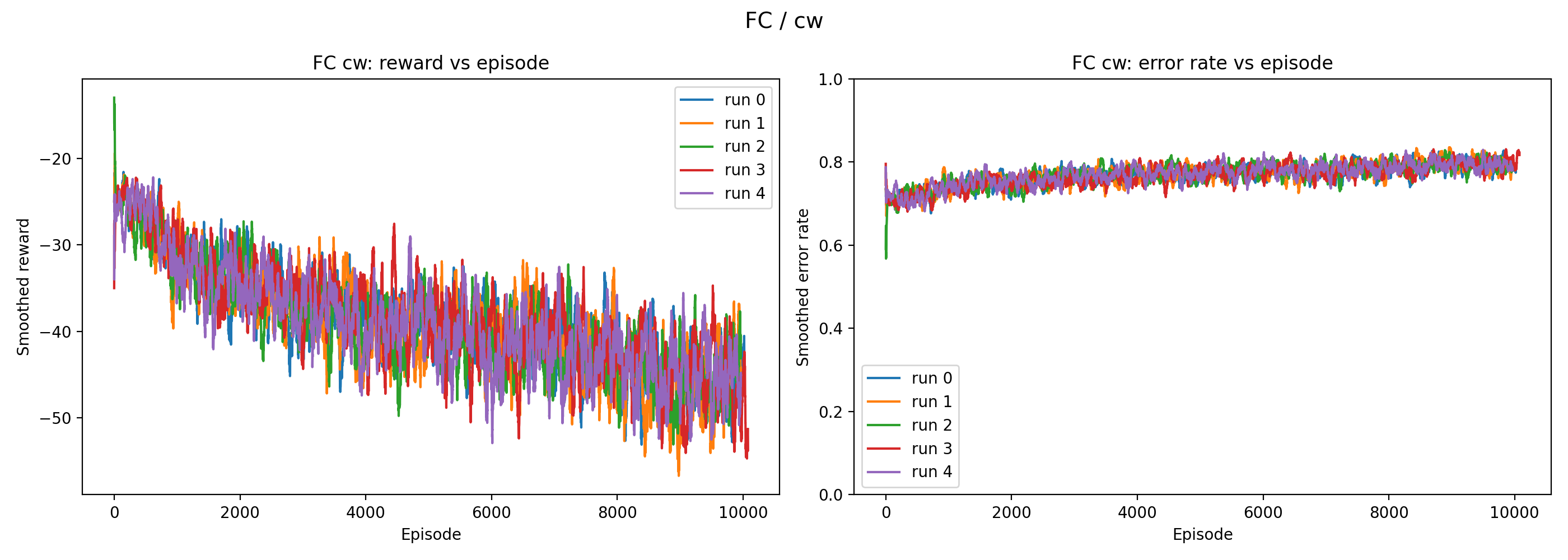}
        \caption{FC, legacy=False}
    \end{subfigure}

    \vspace{0.8em}

    \begin{subfigure}{0.48\linewidth}
        \centering
        \includegraphics[width=\linewidth]{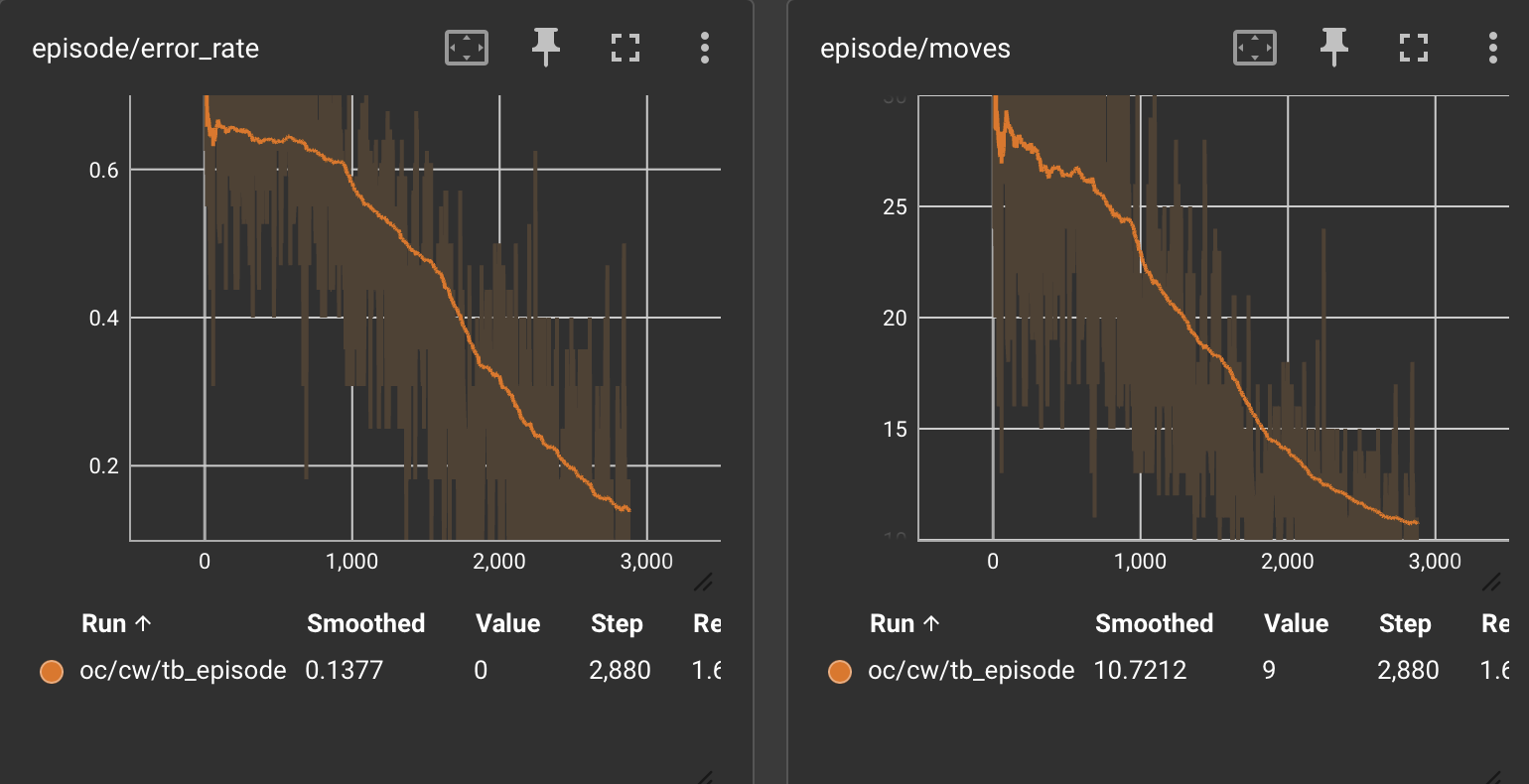}
        \caption{OC, legacy=True}
    \end{subfigure}
    \hfill
    \begin{subfigure}{0.48\linewidth}
        \centering
        \includegraphics[width=\linewidth]{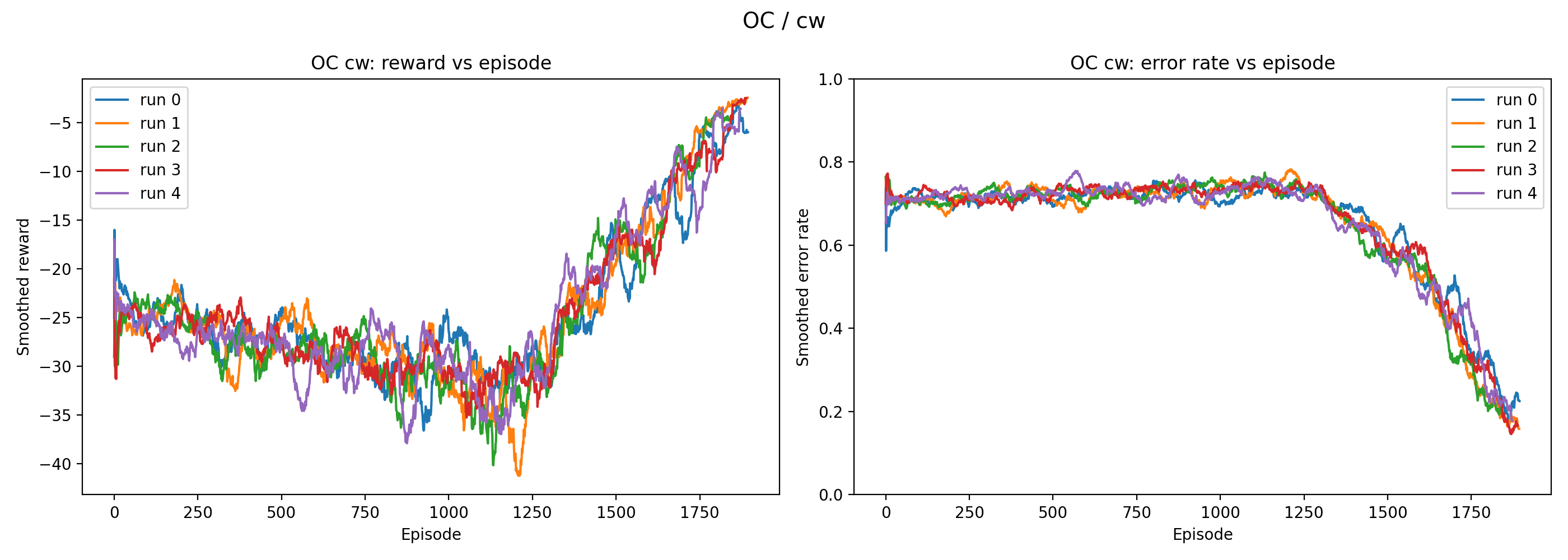}
        \caption{OC, legacy=False}
    \end{subfigure}

    \caption{
    Comparison of learning behavior for the clockwise rule \texttt{cw} under the legacy history-augmented observation pipeline and the simplified Gymnasium current-state observation design. Dark-background TensorBoard plots correspond to \texttt{legacy=True}, while white-background plots correspond to \texttt{legacy=False}.
    }
    \label{fig:gym_legacy_comparison}
\end{figure}

Fig.~\ref{fig:gym_legacy_comparison} illustrates the effect of replacing the legacy history-augmented observations with the simplified Gymnasium observation design. For the clockwise ordering rule (\texttt{cw}), both FC and OC learn successfully under the legacy observation pipeline. However, when only the current state is provided, FC fails to learn the rule and OC exhibits markedly different learning dynamics.

Preliminary experiments using the default Gymnasium observation design produced substantially different learning behavior from that observed with the legacy implementation. In particular, several history-dependent rules, including clockwise and counter-clockwise ordering rules, exhibited substantially degraded learning performance when history information was removed from the observation. Similar behavior was observed for several other history-dependent rules.

These results suggest that the legacy observation pipeline contains information that is important for solving certain GOHR rule families. Although the Gymnasium refactoring successfully reproduced the game mechanics and rule definitions, the simplified observation design did not provide an adequate replacement for the legacy representation.

These observations should not be interpreted as evidence that Gymnasium-based implementations are unsuitable for GOHR. Rather, they suggest that the history information embedded in the legacy observation pipeline plays an important role in the learning process. Future work may investigate alternative mechanisms for providing this information, such as recurrent architectures, explicit memory modules, or history wrappers that preserve compatibility with standard reinforcement-learning interfaces.

To preserve compatibility with previous GOHR studies, legacy-compatible wrappers were therefore developed to reconstruct the original history-augmented observations within the Gymnasium framework.

\subsection{Rendering and Visualization}

The refactored environment also introduces a rendering interface for visualization and debugging.

\begin{figure}[htbp]
    \centering
    \includegraphics[width=0.9\linewidth]{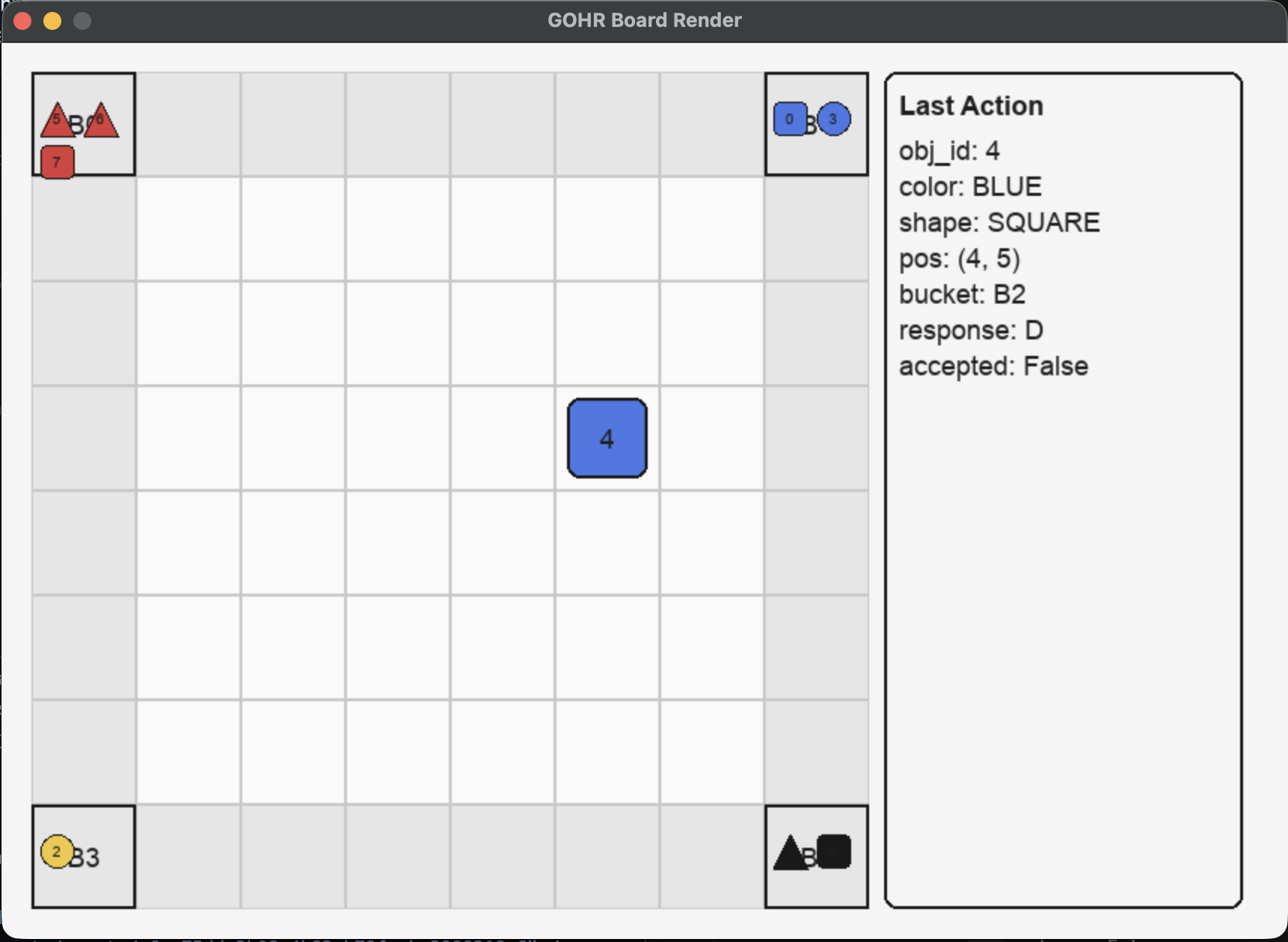}
    \caption{
    Example visualization generated by the Gymnasium-compatible GOHR environment. The display shows the current board state, bucket contents, and feedback associated with the most recent action.
    }
    \label{fig:gym_render}
\end{figure}

The rendering interface supports interactive inspection of gameplay and can be accessed through:

\begin{center}
\texttt{env.render(mode="human")}
\end{center}

This functionality simplifies debugging and provides a convenient way to visualize agent behavior during training and evaluation.

\subsection{Availability}

The source code for the Gymnasium-compatible GOHR environment is publicly available on GitHub.\footnote{\url{https://github.com/christo357/GOHR_gymnasium}}
% \pk{This is not a real url  is it?}

The implementation provides a foundation for future studies involving reinforcement learning, transfer learning, curriculum learning, and human-agent comparisons within the GOHR framework.

\FloatBarrier

\section{Rule Descriptions}

\label{appendix:rules}

% \tagpdfsetup{table/header-rows=1}
\begin{longtable}{@{} l p{0.65\textwidth} @{}}
\caption{List of rules and their descriptions.}
\label{tab:rules-desc} \\

\toprule
\textbf{Rule} & \textbf{Description} \\
\midrule
\endfirsthead

\toprule
\textbf{Rule} & \textbf{Description} \\
\midrule
\endhead

\midrule
\multicolumn{2}{r}{\small Continued on next page} \\
\endfoot

\bottomrule
\endlastfoot

\texttt{allOfColOrd\_BRKY} & All pieces of one color are removed before others, in order: Blue \textrightarrow Red \textrightarrow Black \textrightarrow Yellow. \\
\texttt{allOfShaOrd\_qcts} & All pieces of one shape are removed before others, in order: Square \textrightarrow Circle \textrightarrow Triangle \textrightarrow Star. \\
\texttt{ccw} & Start with any bucket and fill buckets in counterclockwise order. \\
\texttt{cm\_RBKY} & Assign colors to specific buckets in any order: B0=Red, B1=Blue, B2=Black, B3=Yellow. \\
\texttt{col1Ord\_BRKY} & Assign colors to any bucket in order. Skip missing colors: Blue \textrightarrow Red \textrightarrow Black \textrightarrow Yellow; loop. \\
\texttt{col1Ord\_KRBY} & Assign colors to any bucket in order. Skip missing colors: Black \textrightarrow Red \textrightarrow Blue \textrightarrow Yellow; loop. \\
\texttt{col1OrdBuck\_BRKY0213} & Assign colors to specific buckets in order: B0=Blue, B2=Red, B1=Black, B3=Yellow; loop. \\
\texttt{colOrdL1\_BRKY} & Remove colors in order; if a color is missing, take the next object of that color in reading order. Blue \textrightarrow Red \textrightarrow Black \textrightarrow Yellow; loop. \\
\texttt{cw} & Start with any bucket and fill buckets in clockwise order. \\
\texttt{ordL1} & Remove pieces in reading order and assign pieces to any bucket. \\
\texttt{ordL1\_Nearby} & Remove pieces in reading order and assign each to the nearest bucket. \\
\texttt{ordRevOfL1} & Remove pieces in reverse reading order and assign pieces to any bucket. \\
\texttt{ordRevOfL1\_Remotest} & Remove pieces in reverse reading order and assign each to the farthest bucket. \\
\texttt{quadMixed1} & Assign pieces in each quadrant to specific buckets: B0= Q3, B1= Q0, B2= Q2, B3= Q1. \\
\texttt{quadNearby} & Assign pieces to the nearest bucket (by quadrant) in any order. \\
\texttt{sha1Ord\_qcts} & Assign shapes to any bucket in order. Skip missing shapes: Square \textrightarrow Circle \textrightarrow Triangle \textrightarrow Star; loop. \\
\texttt{sha1OrdBuck\_qcts0213} & Assign shapes to specific buckets in order: B0=Square, B2=Circle, B1=Triangle, B3=Star; loop. \\
\texttt{shaOrdL1\_qcts} & Remove shapes in order; if a shape is missing, take the next object of that shape in reading order. Square \textrightarrow Circle \textrightarrow Triangle \textrightarrow Star; loop. \\
\texttt{sm\_csqt} & Assign shapes to specific buckets in any order: B0=Circle, B1=Star, B2=Square, B3=Triangle. \\
\texttt{cm\_RBKY\_cw\_0123} & Assign colors to specific buckets in order: B0=Red \textrightarrow B1=Blue \textrightarrow B2=Black \textrightarrow B3=Yellow; loop. \\
\texttt{cw\_0123} & Start with bucket 0 and fill buckets clockwise: 0 \textrightarrow 1 \textrightarrow 2 \textrightarrow 3; loop. \\
\texttt{cm\_ordL1} & Colors are removed in reading order and assigned to specific buckets: B0=Blue, B1=Red, B2=Black, B3=Yellow. \\
\texttt{cw\_qn2} & Start with bucket 0 and fill buckets clockwise using pieces from the nearest quadrant to that bucket: B0= Q0, B1= Q1, B2= Q2, B3= Q3. \\

\end{longtable}
% \bottomrule
% \end{longtblr}

\leavevmode\textbf{Note.}
(1) Bucket and quadrant indices range from 0 to 3. Bucket 0 and quadrant 0 correspond to the top-left of the board; 1 top-right; 2 bottom-right; 3 bottom-left. 
(2) Reading order proceeds left-to-right, top-to-bottom. 
(3)  B0–B3 denote buckets 0–3, and Q0–Q3 denote quadrants 0–3.

\FloatBarrier

\section{Rules Used in Rule-Based Experiments}

\begin{table}[htbp]
\centering
\caption{Rules used in Rule-Based experiments.}
\label{tab:rule_based_exp}
% \tagpdfsetup{table/header-rows=1}

\begin{tabular}{@{} r l @{\hspace{1.5cm}} r l @{}}
\toprule
\textbf{S.No} & \textbf{Rules} & \textbf{S.No} & \textbf{Rules} \\
\midrule
1  & allOfColOrd\_BRKY       & 10 & ordL1\_Nearby      \\
2  & allOfShaOrd\_qcts       & 11 & ordRevOfL1         \\
3  & ccw                     & 12 & ordRevOfL1\_Remotest\\
4  & cm\_RBKY                & 13 & quadMixed1         \\
5  & col1Ord\_BRKY           & 14 & quadNearby         \\
6  & col1OrdBuck\_BRKY0213   & 15 & sha1Ord\_qcts      \\
7  & colOrdL1\_BRKY          & 16 & sha1OrdBuck\_qcts0213 \\
8  & cw                      & 17 & shaOrdL1\_qcts     \\
9  & ordL1                   & 18 & sm\_csqt           \\
\bottomrule
\end{tabular}
\end{table}

\FloatBarrier

\clearpage
\section{Rule-Property Difficulty Rankings}

\begin{table}[ht]
\centering
\caption{Rule properties arranged in increasing order of difficulty (measured by $M^\star$) for the FC model. The window size for $M^\star$ is 15. Highlighted rows indicate properties that are combined with other properties.}
\label{tab:aspects_FC}

% \tagpdfsetup{table/header-rows=1}

\normalsize
\begin{tabular}{@{} l l S[table-format=5.0, mode=text] @{}}
\toprule
\textbf{Property} & \textbf{Rule} & \textbf{$M^\star$} \\
\midrule
Quadrant-to-bucket mapping & \texttt{quadNearby} & 421 \\
\rowcolor{HighlightA}Feature ordering & \texttt{sha1OrdBuck\_qcts0213} & 426 \\
Quadrant-to-bucket mapping & \texttt{quadMixed1} & 468 \\
\rowcolor{HighlightC}Feature ordering & \texttt{shaOrdL1\_qcts} & 708 \\
\rowcolor{HighlightB}Proximity & \texttt{ordL1\_Nearby} & 782 \\
\rowcolor{HighlightB}Proximity & \texttt{ordRevOfL1\_Remotest} & 795 \\
\rowcolor{HighlightA}Feature ordering & \texttt{col1OrdBuck\_BRKY0213} & 988 \\
Reading order & \texttt{ordL1} & 1458 \\
\rowcolor{HighlightB}Reading order & \texttt{ordRevOfL1\_Remotest} & 1628 \\
\rowcolor{HighlightB}Reading order & \texttt{ordL1\_Nearby} & 1716 \\
Reading order & \texttt{ordRevOfL1} & 1809 \\
\rowcolor{HighlightC}Feature ordering & \texttt{colOrdL1\_BRKY} & 2943 \\
Feature-to-bucket mapping & \texttt{sm\_csqt} & 10577 \\
Feature-to-bucket mapping & \texttt{cm\_RBKY} & 12278 \\
All pieces of feature & \texttt{allOfShaOrd\_qcts} & 15020 \\
All pieces of feature & \texttt{allOfColOrd\_BRKY} & 15563 \\
\rowcolor{HighlightA}Feature-to-bucket mapping & \texttt{col1OrdBuck\_BRKY0213} & 22709 \\
\rowcolor{HighlightA}Feature-to-bucket mapping & \texttt{sha1OrdBuck\_qcts0213} & 23040 \\
Bucket order correct & \texttt{ccw} & 56327 \\
Bucket order correct & \texttt{cw} & 63377 \\
Feature ordering & \texttt{col1Ord\_BRKY} & 80000 \\
Feature ordering & \texttt{sha1Ord\_qcts} & 124342 \\
\rowcolor{HighlightC}Conditional & \texttt{colOrdL1\_BRKY} & 170992 \\
\rowcolor{HighlightC}Conditional & \texttt{shaOrdL1\_qcts} & 229772 \\
\bottomrule
\end{tabular}
\end{table}

\begin{table}[ht]
\centering
\caption{Rule properties arranged in increasing order of difficulty (measured by $M^\star$) for the OC model. The window size for $M^\star$ is 15. Highlighted rows indicate properties that are combined with other properties.}
\label{tab:aspects_OC}

% \tagpdfsetup{table/header-rows=1}

\normalsize
\begin{tabular}{@{} l l S[table-format=5.0, mode=text] @{}}
\toprule
\textbf{Property} &
\textbf{Rule} &
\textbf{$M^\star$} \\
\midrule
\rowcolor{HighlightA}Feature ordering & \texttt{sha1OrdBuck\_qcts0213} & 156 \\
\rowcolor{HighlightA}Feature ordering & \texttt{col1OrdBuck\_BRKY0213} & 526 \\
All pieces of feature & \texttt{allOfShaOrd\_qcts} & 561 \\
All pieces of feature & \texttt{allOfColOrd\_BRKY} & 611 \\
Feature-to-bucket mapping & \texttt{cm\_RBKY} & 662 \\
Quadrant-to-bucket mapping & \texttt{quadNearby} & 696 \\
Feature-to-bucket mapping & \texttt{sm\_csqt} & 733 \\
Quadrant-to-bucket mapping & \texttt{quadMixed1} & 915 \\
Bucket order correct & \texttt{ccw} & 1128 \\
\rowcolor{HighlightA}Feature-to-bucket mapping & \texttt{sha1OrdBuck\_qcts0213} & 1176 \\
\rowcolor{HighlightA}Feature-to-bucket mapping & \texttt{col1OrdBuck\_BRKY0213} & 1212 \\
Bucket order correct & \texttt{cw} & 1305 \\
Reading order & \texttt{ordRevOfL1} & 1456 \\
Reading order & \texttt{ordL1} & 1648 \\
\rowcolor{HighlightB}Reading order & \texttt{ordRevOfL1\_Remotest} & 2358 \\
\rowcolor{HighlightB}Proximity & \texttt{ordRevOfL1\_Remotest} & 3139 \\
\rowcolor{HighlightC}Feature ordering & \texttt{shaOrdL1\_qcts} & 3336 \\
\rowcolor{HighlightB}Proximity & \texttt{ordL1\_Nearby} & 3939 \\
\rowcolor{HighlightB}Reading order & \texttt{ordL1\_Nearby} & 4061 \\
\rowcolor{HighlightC}Feature ordering & \texttt{colOrdL1\_BRKY} & 8905 \\
Feature ordering & \texttt{sha1Ord\_qcts} & 21197 \\
Feature ordering & \texttt{col1Ord\_BRKY} & 25980 \\
\rowcolor{HighlightC}Conditional & \texttt{shaOrdL1\_qcts} & 34115 \\
\rowcolor{HighlightC}Conditional & \texttt{colOrdL1\_BRKY} & 37964 \\
\bottomrule
\end{tabular}
\end{table}

\FloatBarrier

\clearpage
\section{Overall Rule Difficulty for FC and OC Models}
\begin{table}[htb]
\centering
% \caption{Rules ordered from easiest (top) to hardest (bottom) for the FC transformer model. Ordering is based on majority ranking across three metrics. \pk{WHAT DOES THIS MEAN, EXACTLY?} Values shown ($E^\star_{\mathrm{mean}}$, $E^\star_{\mathrm{max}}$, $M^\star$) are medians across five runs.}

\caption{Rules are grouped from easier (top) to harder (bottom) categories for the FC transformer model, based on their overall performance across $M^\star$, $E^\star_{\mathrm{mean}}$ and $E^\star_{\mathrm{max}}$. Rules within a category are not intended to represent a strict ordering. Reported values are medians across five runs.
}
\label{tab:FC_difficulty}
% \tagpdfsetup{table/header-rows=1}

\begin{tabular}{@{} l r r r @{}}
\toprule
\textbf{Rule} &
\textbf{$M^\star$} & \textbf{$E^\star_{mean}$} & \textbf{$E^\star_{max}$} \\
\midrule

\textbf{Highly learnable} & & & \\
\texttt{quadNearby}                &   273 &    14 &    24 \\
\texttt{quadMixed1}                &   325 &    15 &    27 \\
\addlinespace[4pt]

\textbf{Moderately learnable} & & & \\
\texttt{ordL1\_Nearby}             &  1189 &    62 &    73 \\
\texttt{ordL1}                     &  1071 &    62 &    87 \\
\texttt{ordRevOfL1\_Remotest}      &  1294 &    64 &    79 \\
\texttt{ordRevOfL1}                &   984 &    75 &    88 \\
\addlinespace[4pt]

\textbf{Challenging} & & & \\
\texttt{sm\_csqt}                  &  9413 &   403 &   509 \\
\texttt{cm\_RBKY}                  &  9527 &   426 &   440 \\
\texttt{allOfColOrd\_BRKY}         &  3976 &   846 &   934 \\
\texttt{allOfShaOrd\_qcts}         &  2115 &   890 &  1000 \\
\addlinespace[4pt]

\textbf{Difficult} & & & \\
\texttt{col1OrdBuck\_BRKY0213}     & 28112 &  1865 &  2161 \\
\texttt{sha1OrdBuck\_qcts0213}     & 30993 &  2017 &  2321 \\
\texttt{ccw}                       & 28265 &  3269 &  3372 \\
\texttt{cw}                        & 32638 &  3592 &  3869 \\
\addlinespace[4pt]

\textbf{Very difficult} & & & \\
\texttt{sha1Ord\_qcts}             &  2377 & 14145 & 13632 \\
\texttt{col1Ord\_BRKY}             &  3382 & 17353 & 17572 \\
\texttt{colOrdL1\_BRKY}            & 58034 & 30268 & 28992 \\
\texttt{shaOrdL1\_qcts}            & 35089 & 33004 & 32998 \\
\bottomrule
\end{tabular}
\end{table}

\begin{table}[htb]
\centering
\caption{Rules are grouped from easier (top) to harder (bottom) categories for the OC transformer model, based on their overall performance across $M^\star$, $E^\star_{\mathrm{mean}}$ and $E^\star_{\mathrm{max}}$. Rules within a category are not intended to represent a strict ordering. Reported values are medians across five runs.}
\label{tab:OC_difficulty}
% \tagpdfsetup{table/header-rows=1}

\begin{tabular}{@{} l r r r @{}}
\toprule
\textbf{Rule} &
% \textbf{M*} &
% \textbf{E* mean} &
% \textbf{E* max} \\
\textbf{$M^\star$} & \textbf{$E^\star_{mean}$} & \textbf{$E^\star_{max}$} \\
\midrule

\textbf{Learnable} & & & \\
\texttt{cm\_RBKY}              &  565 &   17 &   28 \\
\texttt{sm\_csqt}              &  683 &   20 &   26 \\
\texttt{allOfShaOrd\_qcts}     &  548 &   30 &   44 \\
\texttt{quadNearby}            &  667 &   33 &   42 \\
\texttt{allOfColOrd\_BRKY}     &  476 &   37 &   47 \\
\texttt{quadMixed1}            &  829 &   36 &   50 \\
\texttt{ccw}                   &  612 &   60 &   85 \\
\texttt{cw}                    &  738 &   53 &   68 \\
\texttt{ordL1}                 &  689 &   74 &   99 \\
\texttt{ordRevOfL1}            &  807 &   87 &  122 \\
\addlinespace[4pt]

\textbf{Challenging / conditional} & & & \\
\texttt{ordRevOfL1\_Remotest}  & 2129 &  136 &  233 \\
\texttt{ordL1\_Nearby}         & 2478 &  149 &  183 \\
\texttt{col1OrdBuck\_BRKY0213} & 1816 &  554 &  568 \\
\texttt{sha1OrdBuck\_qcts0213} & 2729 &  395 &  556 \\
\addlinespace[4pt]

\textbf{Difficult} & & & \\
\texttt{sha1Ord\_qcts}         &  4530 & 3294 & 4113 \\
\texttt{col1Ord\_BRKY}         &  5058 & 2114 & 2891 \\
\texttt{shaOrdL1\_qcts}        & 13057 & 3205 & 3262 \\
\texttt{colOrdL1\_BRKY}        & 13110 & 3232 & 3675 \\
\bottomrule
\end{tabular}

\end{table}

% tables / descriptions
\FloatBarrier

\section{Transfer Plots}
\label{appendix:transfer_plots}

\begin{figure}[htbp]
    \centering
    \includegraphics[width=0.92\linewidth]{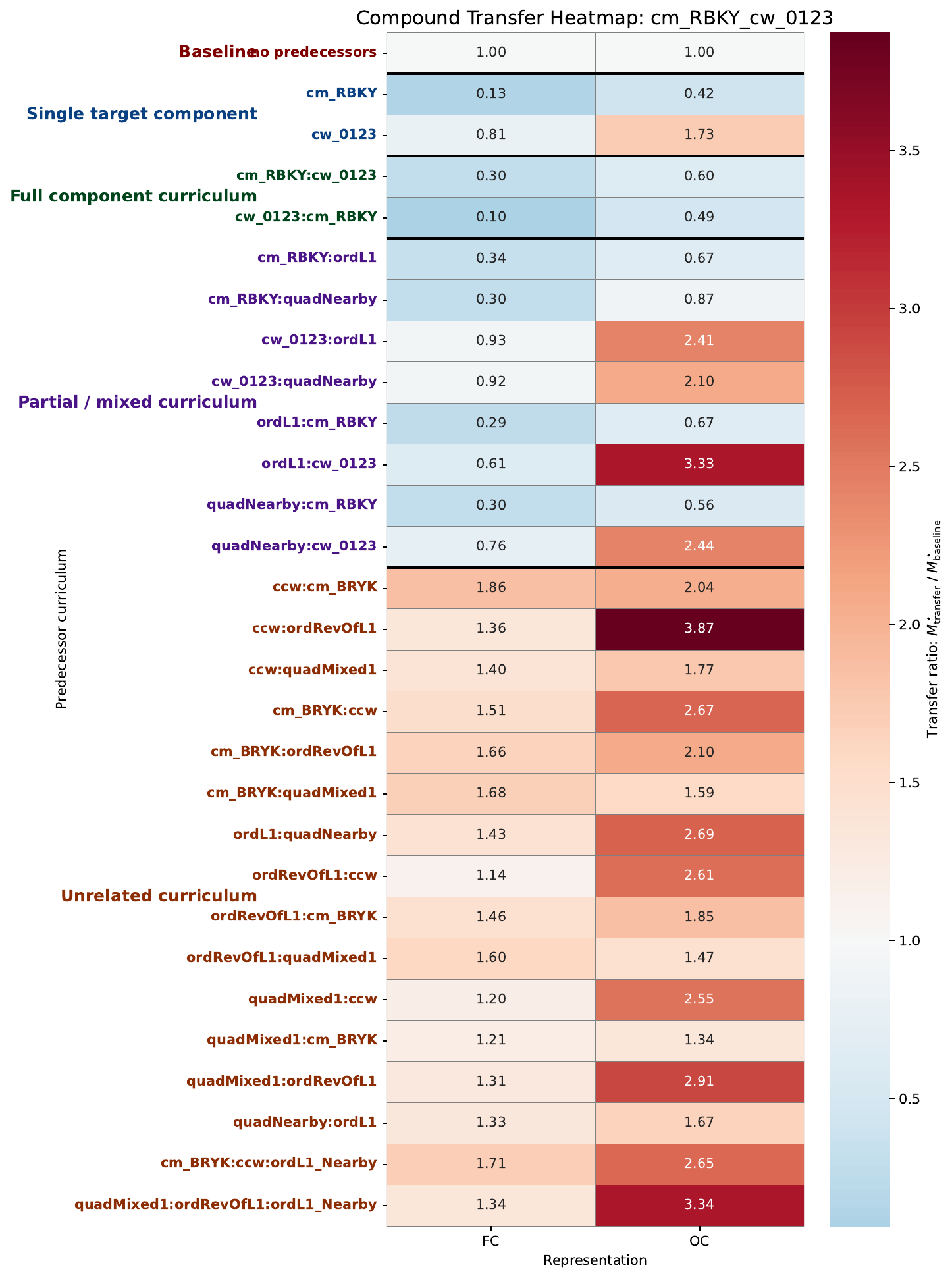}
    \caption{
    Transfer heatmap for the compound rule \texttt{cm\_RBKY\_cw\_0123}. 
    Each cell reports the normalized transfer ratio
    $M^\star_{\mathrm{transfer}} / M^\star_{\mathrm{baseline}}$.
    Values below $1$ indicate positive transfer, while values above $1$ indicate negative transfer.
    }
    \label{fig:compound_cm_RBKY_cw_0123_heatmap}
\end{figure}

\FloatBarrier

% figures

% ============================================================
\section{Clustering and MDS Results}
\label{appendix:clustering}
% ============================================================
%\pk{CHRISTO: CAN YOU FIGURE OUT A WAY TO INCLUDE THE RELEVANT BLOCK OF F-CODE FROM THE .Rmd FILE, FOR EACH OF THESE DIAGRAMS?}
% \pk{THANKS!}
%\christo{
The main R code blocks used to generate these visualizations are included below. The FC and OC figures were produced using the same analysis pipeline, with the input file changed from \texttt{FC\_diff\_matrix.csv} to \texttt{OC\_diff\_matrix.csv}.
This appendix provides additional hierarchical clustering, multidimensional scaling (MDS), and tanglegram visualizations discussed in Section~\ref{sec:transfer_geometry}. These figures complement the main-text analyses by showing the complete set of FC and OC transfer-geometry results.

% ------------------------------------------------------------
\subsection{Hierarchical Clustering Results}
% ------------------------------------------------------------

\begin{lstlisting}[caption={R code for row and column hierarchical clustering.}, label={lst:cluster_code}]
mat <- as.matrix(read.csv("FC_diff_matrix.csv",
                          row.names = 1,
                          check.names = FALSE))

rownames(mat) <- paste0(LETTERS[seq_len(nrow(mat))], ":", rownames(mat))
colnames(mat) <- paste0(LETTERS[seq_len(ncol(mat))], ":", colnames(mat))

# Column-mean imputation for row clustering
mat_row <- mat
for (j in 1:ncol(mat_row)) {
  col_mean <- mean(mat_row[, j], na.rm = TRUE)
  mat_row[is.na(mat_row[, j]), j] <- col_mean
}

row_dist <- dist(mat_row)
row_hclust <- hclust(row_dist)
plot(row_hclust, main = "Row Clustering")

# Row-mean imputation for column clustering
mat_col <- mat
for (i in 1:nrow(mat_col)) {
  row_mean <- mean(mat_col[i, ], na.rm = TRUE)
  mat_col[i, is.na(mat_col[i, ])] <- row_mean
}

col_dist <- dist(t(mat_col))
col_hclust <- hclust(col_dist)
plot(col_hclust, main = "Column Clustering")
\end{lstlisting}

\begin{figure}[htbp]
    \centering
    \includegraphics[width=0.82\linewidth]{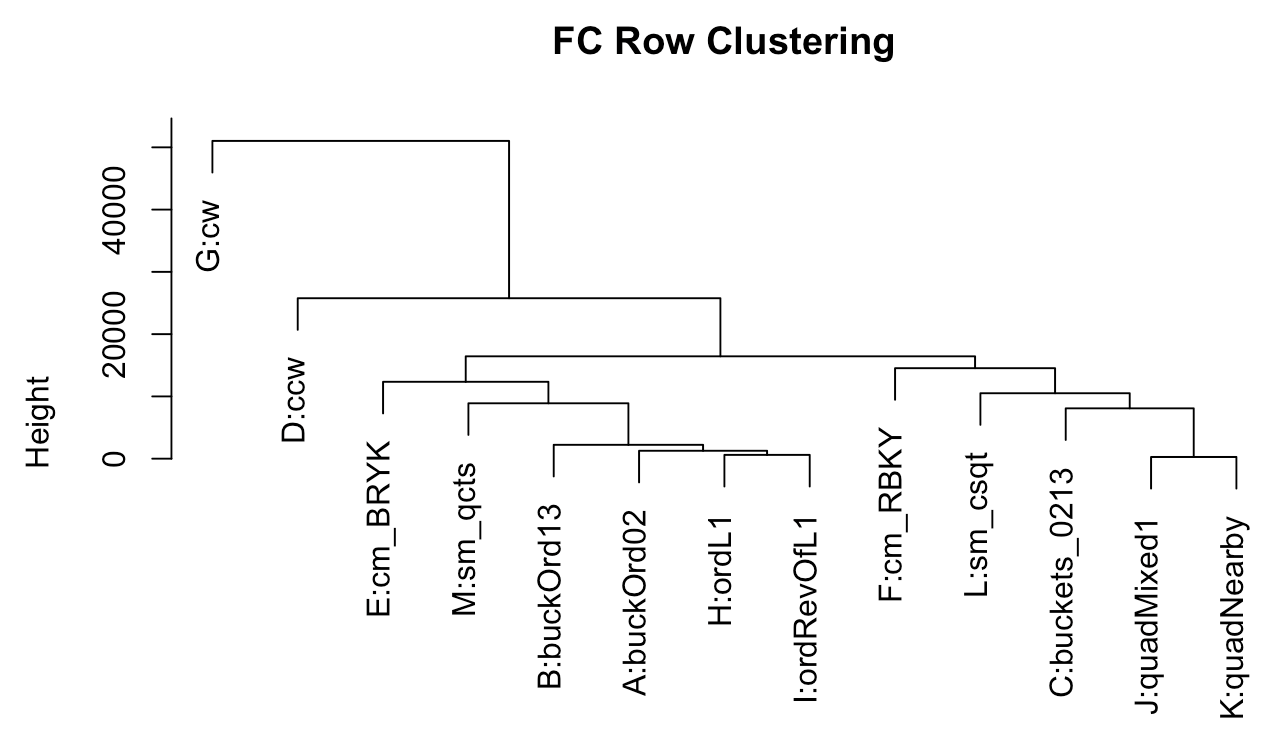}
    \caption{
    Hierarchical clustering of FC successor-rule transfer behavior. Rules that cluster together exhibit similar sensitivity to predecessor-rule experience.
    }
    \label{fig:appendix_fc_row_cluster}
\end{figure}

\begin{figure}[htbp]
    \centering
    \includegraphics[width=0.82\linewidth]{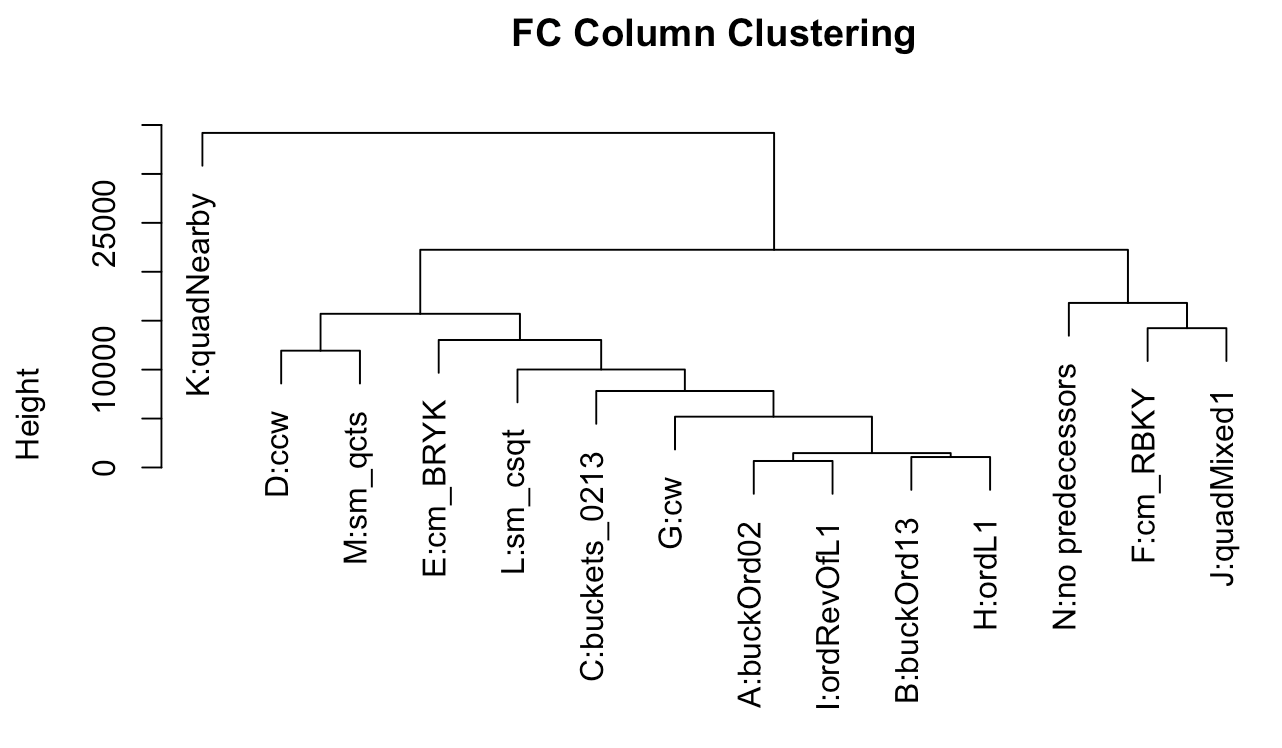}
    \caption{
    Hierarchical clustering of FC predecessor-rule influence. Rules that cluster together produce similar transfer effects on successor-rule learning.
    }
    \label{fig:appendix_fc_column_cluster}
\end{figure}

\begin{figure}[htbp]
    \centering
    \includegraphics[width=0.82\linewidth]{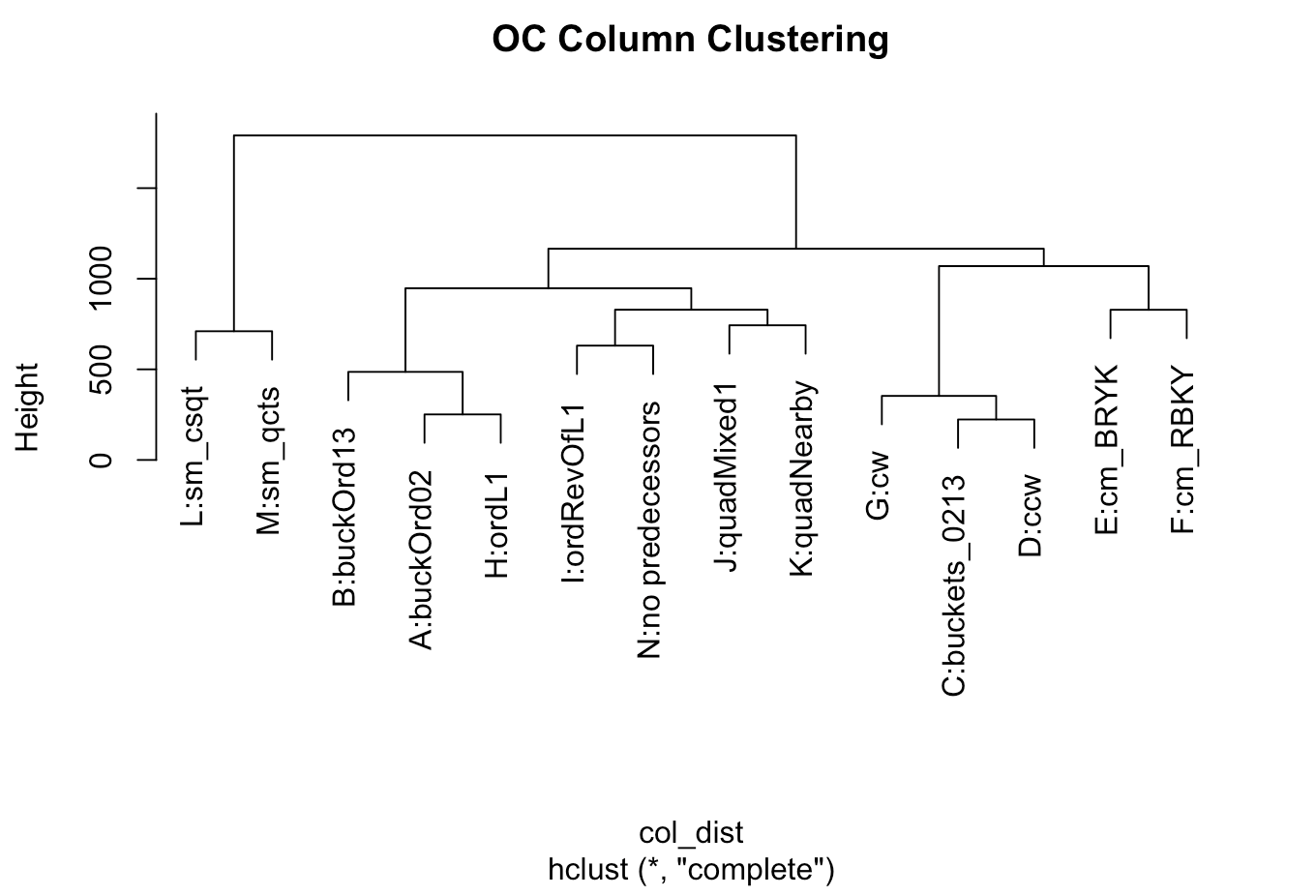}
    \caption{
    Hierarchical clustering of OC predecessor-rule influence. The resulting clusters exhibit stronger separation between ordering-based, feature-based, and spatial rule families than the corresponding FC clustering.
    }
    \label{fig:appendix_oc_column_cluster}
\end{figure}

\clearpage

% ------------------------------------------------------------
\subsection{Multidimensional Scaling Results}
% ------------------------------------------------------------

\begin{lstlisting}[caption={R code for 3-dimensional MDS visualization.}, label={lst:mds_code}]
library(smacof)
library(plotly)

display_rotatable_MDS <- function(mat, plot_title) {
  dist_mat <- dist(mat)

  mds_result <- smacof::smacofSym(dist_mat, ndim = 3)

  mds_df <- data.frame(
    mds_result$conf,
    label = sub(":.*", "", rownames(mat)),
    rule = sub("^[^:]*:", "", rownames(mat))
  )

  plot_ly(
    data = mds_df,
    x = ~D1,
    y = ~D2,
    z = ~D3,
    type = "scatter3d",
    mode = "text",
    text = ~label,
    hovertext = ~rule,
    hoverinfo = "text"
  ) %>%
    layout(
      title = plot_title,
      scene = list(
        xaxis = list(title = "Dimension 1"),
        yaxis = list(title = "Dimension 2"),
        zaxis = list(title = "Dimension 3")
      )
    )
}

display_rotatable_MDS(mat_row, "Row MDS")
display_rotatable_MDS(t(mat_col), "Column MDS")
\end{lstlisting}

\begin{figure}[htbp]
    \centering
    \includegraphics[width=0.88\linewidth]{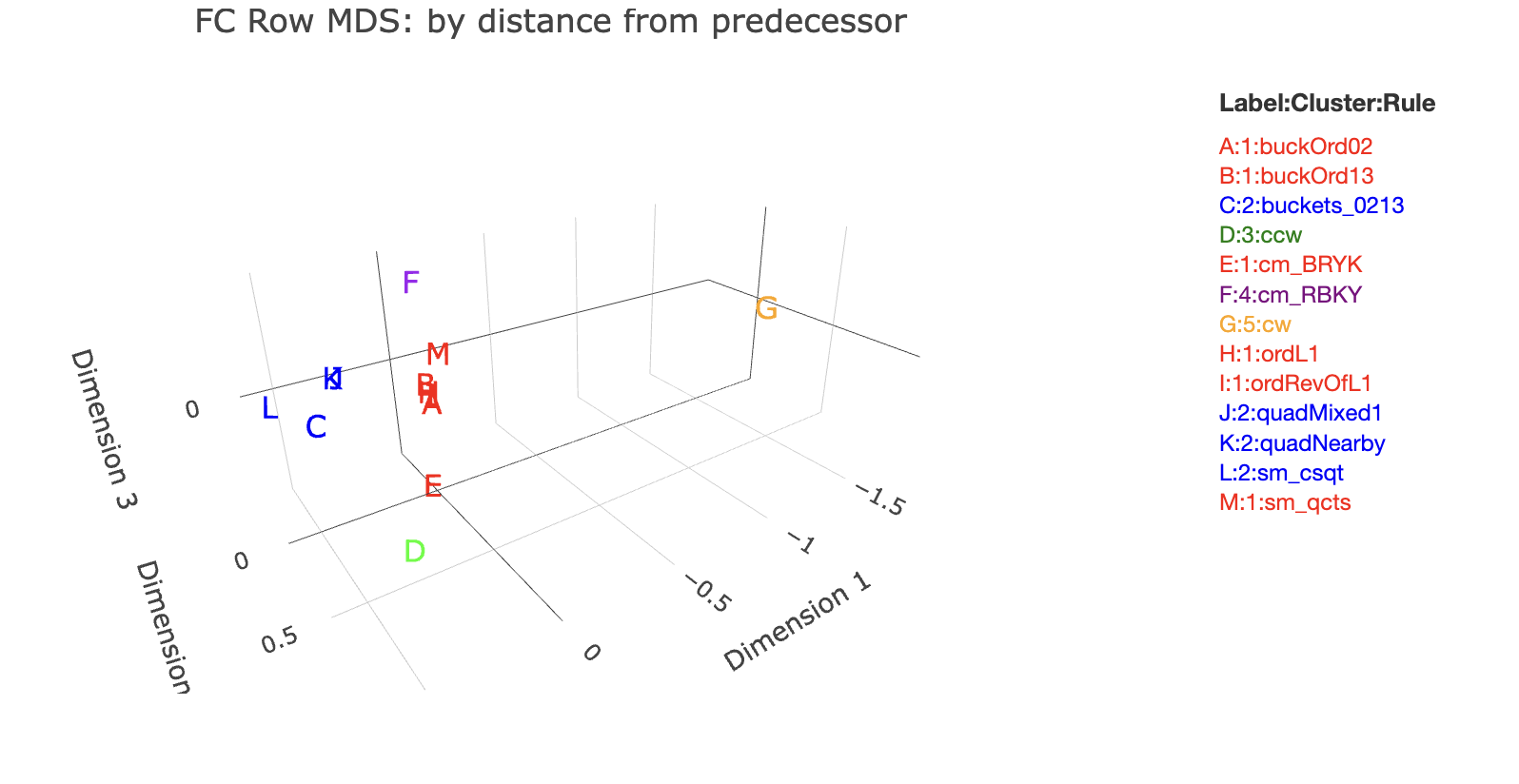}
    \caption{
    3-dimensional MDS embedding of FC successor-rule transfer behavior. Distances reflect similarity in how rules respond to predecessor experience.
    }
    \label{fig:appendix_mds_fc_row}
\end{figure}

\begin{figure}[htbp]
    \centering
    \includegraphics[width=0.88\linewidth]{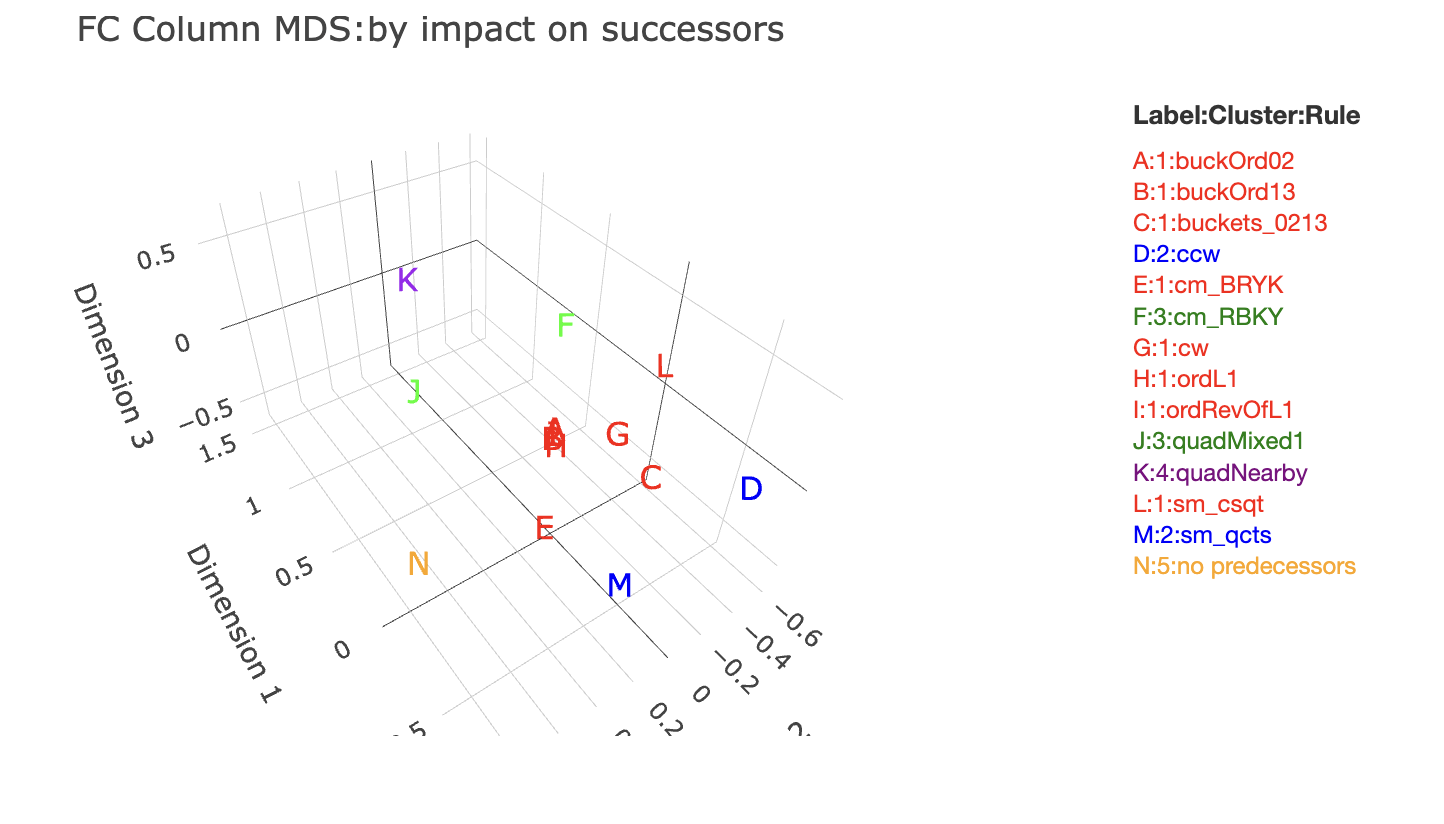}
    \caption{
    3-dimensional MDS embedding of FC predecessor-rule influence. Distances reflect similarity in how predecessor rules affect successor learning.
    }
    \label{fig:appendix_mds_fc_column}
\end{figure}

\begin{figure}[htbp]
    \centering
    \includegraphics[width=0.88\linewidth]{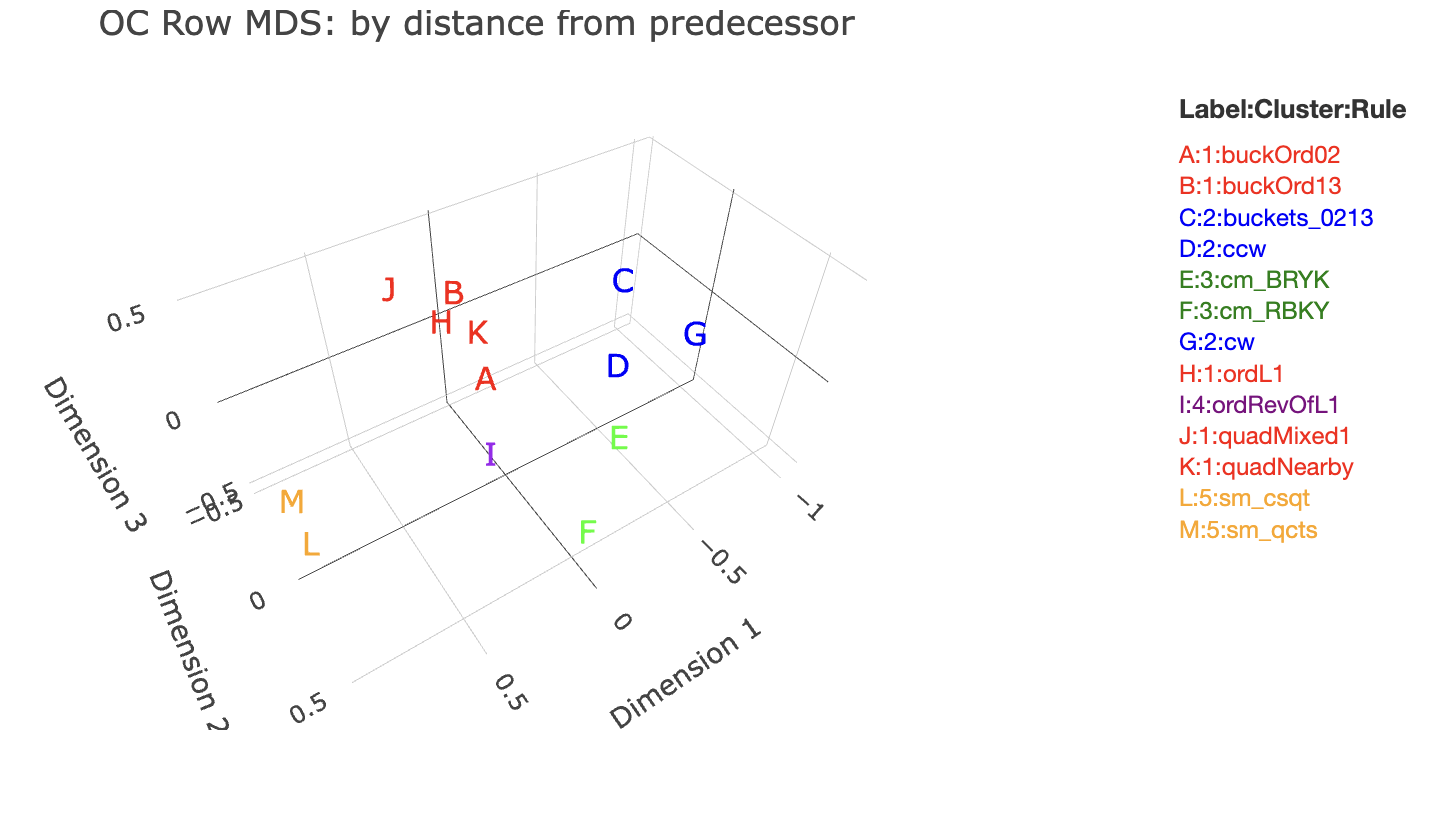}
    \caption{
    3-dimensional MDS embedding of OC successor-rule transfer behavior. Structurally related rules form more compact and separable clusters than in the FC embedding.
    }
    \label{fig:appendix_mds_oc_row}
\end{figure}

\clearpage

% ------------------------------------------------------------
\subsection{Tanglegram Comparisons}
% ------------------------------------------------------------

\begin{lstlisting}[caption={R code for comparing clustering structures using tanglegrams.}, label={lst:tanglegram_code}]
library(dendextend)

strip_prefix <- function(x) {
  sub("^[A-Z]:", "", x)
}

compare_dendrograms <- function(h1, h2, labels1, labels2, name1, name2) {
  common_rules <- intersect(labels1, labels2)

  d1 <- as.dendrogram(h1)
  d2 <- as.dendrogram(h2)

  labels(d1) <- labels1
  labels(d2) <- labels2

  d1 <- prune(d1, setdiff(labels(d1), common_rules))
  d2 <- prune(d2, setdiff(labels(d2), common_rules))

  dlist <- dendlist(d1, d2)

  cat("Cophenetic correlation:", cor_cophenetic(d1, d2), "\n")
  cat("Baker's gamma:", cor_bakers_gamma(d1, d2), "\n")
  cat("Entanglement:", entanglement(dlist), "\n")

  tanglegram(
    d1,
    d2,
    main = paste(name1, "vs", name2),
    common_subtrees_color_lines = TRUE,
    highlight_distinct_edges = FALSE
  )
}
\end{lstlisting}

\begin{figure}[htbp]
    \centering
    \includegraphics[width=\linewidth]{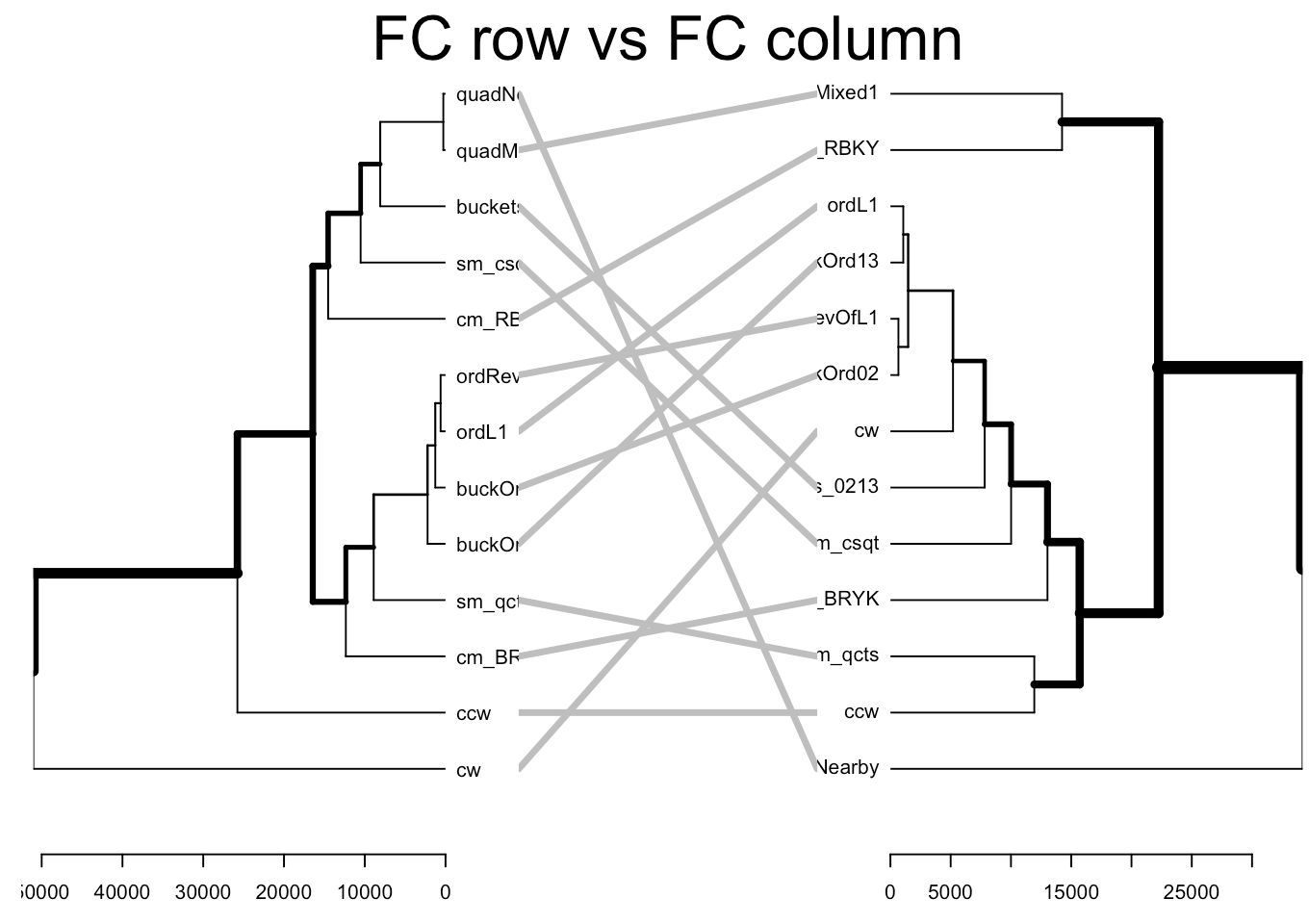}
    \caption{
    Tanglegram comparing FC row clustering and FC column clustering. The large number of crossing connections reflects weaker agreement between predecessor-rule influence and successor-rule sensitivity under the FC representation.
    }
    \label{fig:appendix_fc_row_vs_fc_column}
\end{figure}

\begin{figure}[htbp]
    \centering
    \includegraphics[width=\linewidth]{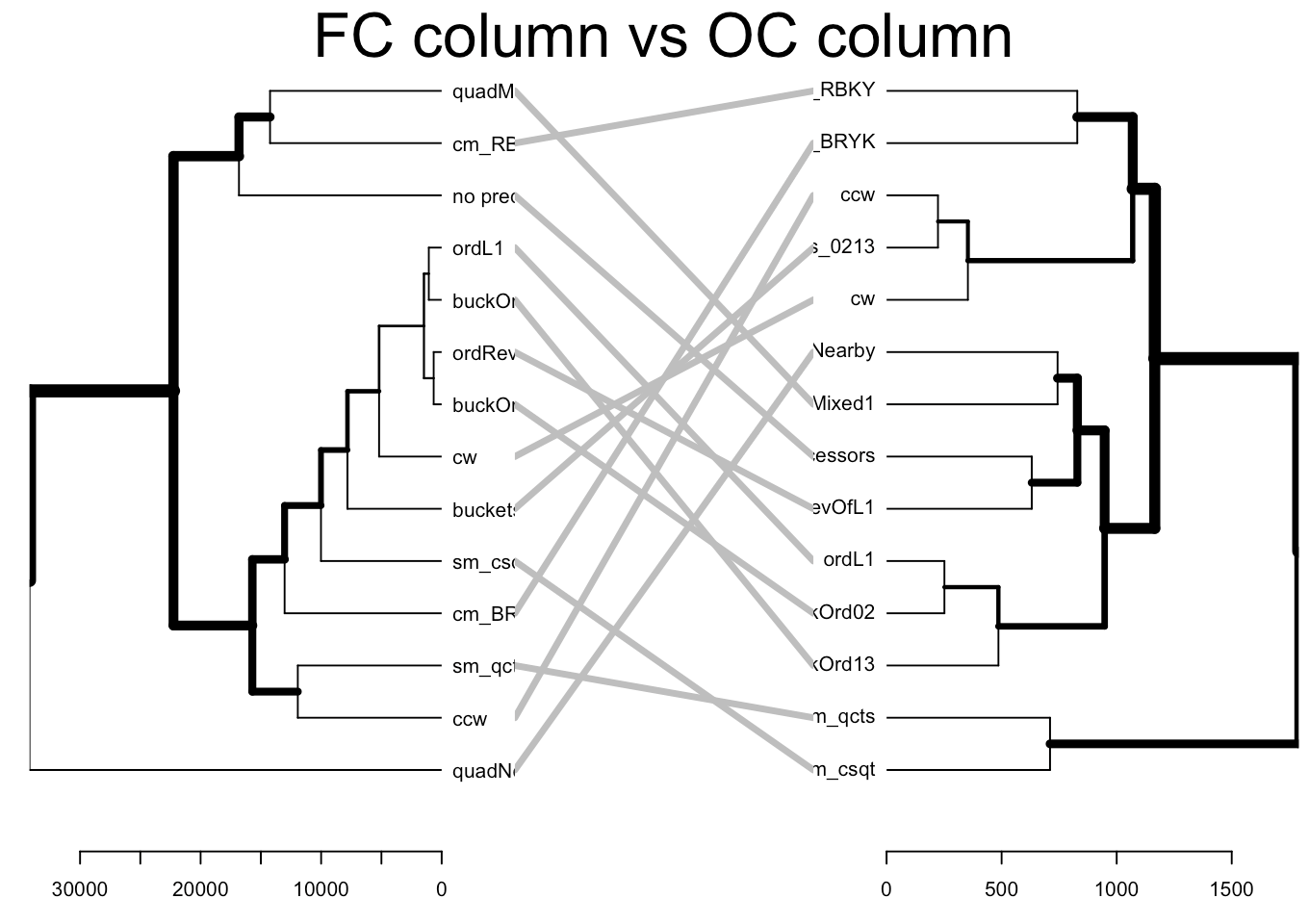}
    \caption{
    Tanglegram comparing FC and OC predecessor-rule clustering structures. The relatively weak alignment indicates that FC and OC organize predecessor-rule influence differently.
    }
    \label{fig:appendix_fc_column_vs_oc_column}
\end{figure}

\begin{figure}[htbp]
    \centering
    \includegraphics[width=\linewidth]{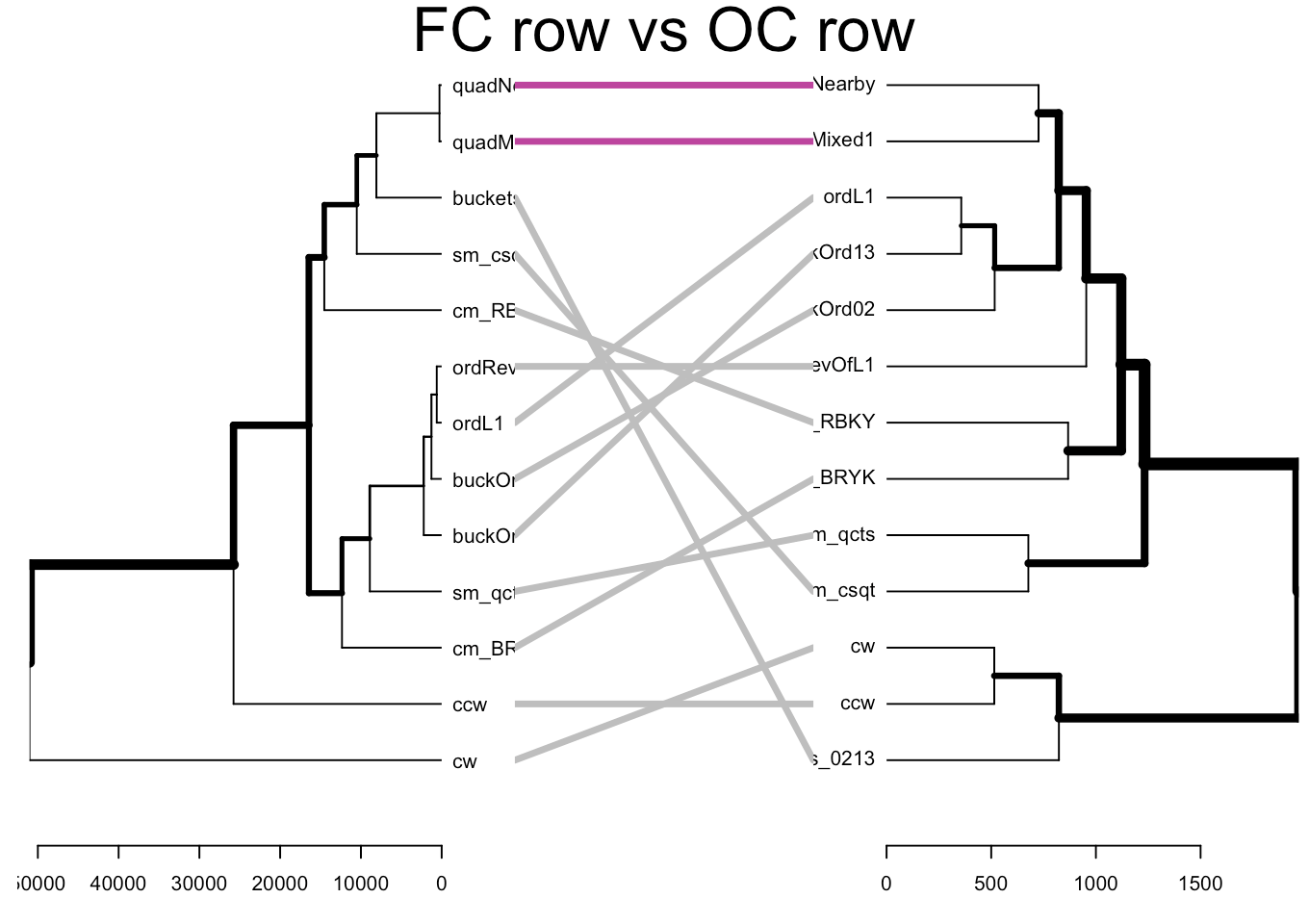}
    \caption{
    Tanglegram comparing FC and OC successor-rule clustering structures. Some high-level relationships remain preserved across representations, particularly among spatial and ordering-based rules.
    }
    \label{fig:appendix_fc_row_vs_oc_row}
\end{figure}

\FloatBarrier

\section{Supplementary Materials}
\label{app:supplementary_materials}

Supplementary materials associated with this report are available at:

\begin{center}
\url{http://action.rutgers.edu/REPORTS/}
\end{center}

These materials include additional interactive visualizations and analysis outputs supporting the results reported in this paper.

\end{document}